\documentclass[sigconf, authorversion,nonacm]{acmart}

\usepackage{booktabs} %
\usepackage{overpic}

\usepackage{numprint}
\usepackage[ruled]{algorithm2e} %
\usepackage{float}
\usepackage{caption}
\usepackage{subcaption}
\usepackage{placeins}

\SetAlFnt{\small}
\SetAlCapFnt{\small}
\SetAlCapNameFnt{\small}
\SetAlCapHSkip{0pt}

\usepackage{amsmath}
\usepackage{comment}
\usepackage{multirow,bigdelim}
\usepackage{lipsum}
\usepackage{array}
\usepackage{wrapfig}

\usepackage{makecell}
\usepackage{blindtext}
\usepackage{xcolor}
\usepackage{soul}
\usepackage{cleveref}
\usepackage{amsfonts}
\usepackage{subcaption}

\makeatletter
\newcommand{\settitle}{\@maketitle}
\makeatother

\newcolumntype{C}[1]{>{\centering\let\newline\\\arraybackslash\hspace{0pt}}m{#1}}

\newif\ifdraft
\drafttrue

\ifdraft
\definecolor{darkpink}{rgb}{0.561, 0.282, 0.427}
\definecolor{darkturquoise}{rgb}{0., 0.81, 0.822}

\newcommand{\dcc}[1]{{\color{red}[\textbf{DC:} #1]}}
\newcommand{\rgc}[1]{{\color{purple}[\textbf{RG:} #1]}}
\newcommand{\opc}[1]{{\color{blue}[\textbf{OP:} #1]}}
\newcommand{\dgc}[1]{{\color{teal}[\textbf{DG:} #1]}}

\newcommand{\gd}[1]{{\color{darkpink}#1}}

\newcommand{\drop}[1]{}

\else
\newcommand{\dcc}[1]{}
\newcommand{\rgc}[1]{}
\newcommand{\opc}[1]{}
\newcommand{\dgc}[1]{}
\newcommand{\gd}[1]{}
\newcommand{\abc}[1]{}

\fi

\newcommand{\munet}[1]{\mu\left(#1\right)}
\newcommand{\xhat}[1]{\hat{x}_{#1}}
\newcommand{\chat}{\hat{c}}

\makeatletter
\DeclareRobustCommand\onedot{\futurelet\@let@token\@onedot}
\def\@onedot{\ifx\@let@token.\else.\null\fi\xspace}

\def\eg{\emph{e.g}\onedot}

\def\ie{\emph{i.e}\onedot}

\makeatother

\usepackage[bottom]{footmisc}
\usepackage{arydshln}

\makeatletter
\def\blfootnote{\xdef\@thefnmark{}\@footnotetext}
\makeatother

\begin{document}
\title{TurboEdit: Text-Based Image Editing Using Few-Step Diffusion Models}

\author{Gilad Deutch}
\affiliation{%
 \institution{Tel-Aviv University}
  \country{} 
 }

\author{Rinon Gal}
\affiliation{%
 \institution{NVIDIA, Tel-Aviv University}
  \country{} 
 }

\author{Daniel Garibi}
\affiliation{%
 \institution{Tel-Aviv University}
  \country{} 
 }

\author{Or Patashnik}
\affiliation{%
 \institution{Tel-Aviv University}
  \country{} 
 }

\author{Daniel Cohen-Or}
\affiliation{%
 \institution{Tel-Aviv University}
  \country{} 
 } 

\begin{abstract}
    
Diffusion models have opened the path to a wide range of text-based image editing frameworks. However, these typically build on the multi-step nature of the diffusion backwards process, and adapting them to distilled, fast-sampling methods has proven surprisingly challenging. Here, we focus on a popular line of text-based editing frameworks - the ``edit-friendly'' DDPM-noise inversion approach. We analyze its application to fast sampling methods and categorize its failures into two classes: the appearance of visual artifacts, and insufficient editing strength. We trace the artifacts to mismatched noise statistics between inverted noises and the expected noise schedule, and suggest a shifted noise schedule which corrects for this offset. To increase editing strength, we propose a pseudo-guidance approach that efficiently increases the magnitude of edits without introducing new artifacts. All in all, our method enables text-based image editing with as few as three diffusion steps, while providing novel insights into the mechanisms behind popular text-based editing approaches. \\
Project page: \url{https://turboedit-paper.github.io/}

\end{abstract}

\begin{teaserfigure}
    \centering
    \includegraphics[width=0.99\textwidth]{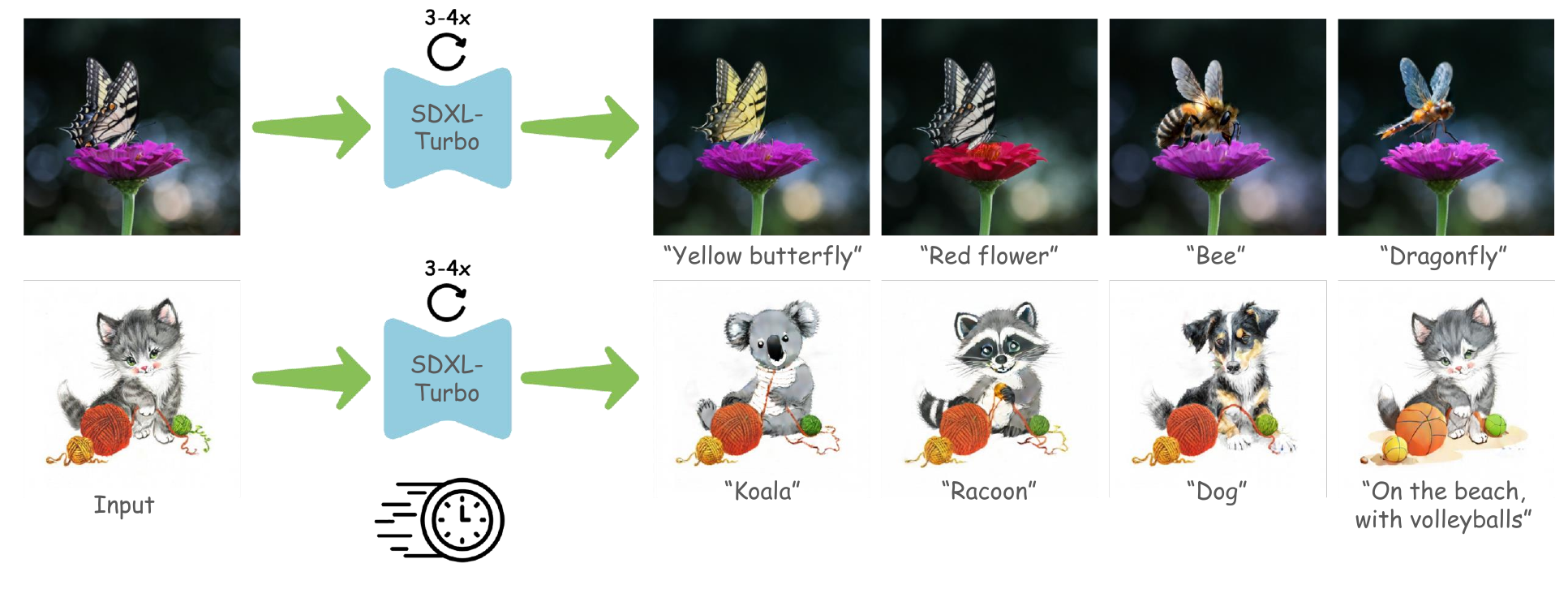}\vspace{-5pt}
    \caption{Our method enables text-based editing of real images with as few as 3 diffusion steps (0.321 seconds on an A5000 GPU).}\label{fig:teaser}
\end{teaserfigure}

\maketitle
\section{Introduction}

The unprecedented expressive power of large-scale text-to-image diffusion models~\cite{ramesh2022hierarchical,rombach2021highresolutionLDM,nichol2021glide} has contributed to a rise of text-based editing frameworks. Through these frameworks, users are empowered to modify existing, real images using natural language instructions. However, these methods commonly rely on the multi-step nature of the diffusion process. In such many-step scenarios, any deviation from train-time statistics can be lessened by injecting smaller changes across many steps, or by allowing the final steps of the diffusion process to draw the result back to the prior distribution, thereby correcting for any artifacts that may arise along the way.

Recently, model distillation methods~\cite{salimans2022progressive,song2023consistency,sauer2023adversarial} have enabled the creation of ``fast'' diffusion models, from which novel images can be sampled in few ($1$-$8$) steps. Ideally, we would like to use such models as a backbone for image editing, thereby accelerating the editing process. However, absent sufficient steps, existing text-based editing approaches tend to create noticeable artifacts, or show poor performance.

Here, we propose to address these limitations by analyzing the behaviour of one such family of editing methods --- the DDPM noise-inversion framework~\cite{HubermanSpiegelglas2023,tsaban2023ledits,wu2023cyclediffusion}.
There, inversion takes the form of a series of per-timestep noise maps. These are pre-calculated, such that using them in place of the random noise samples in the DDPM backwards process, will lead to a re-construction of the original image under a given prompt (see \cref{sec:ddpm_inversion} for more details). Importantly, editing through this approach is simple, requiring only a change of the text describing the image, while still using the same pre-calculated noise samples during the DDPM backwards process. However, attempting to apply this approach to image editing with fast-sampling methods (e.g., SDXL Turbo~\cite{sauer2023adversarial}) leads to the creation of severe visual artifacts, and significantly diminished adherence to the novel prompts (\ie insufficient editing strength). 

To tackle the visual artifacts, we analyze the statistics of the inverted noise maps, and observe that they behave more closely to the noise expected from earlier (more noisy) diffusion steps. We hypothesize that in the case of few-step models, there is no time to correct for the noise-distribution shift induced by these corrections, and hence artifacts arise. To overcome them, we propose a shifted denoising schedule, where the denoising sampler is instructed to remove noise as if it was observing an earlier, more noisy step. Additionally, in typical noise-inversion schemes, there is little impact to the noise used at the last denoising step, and indeed it is often simply discarded. In the case of few-step models, the last step's impact is non-trivial, and we find it helpful to inject noise also in this step, while explicitly normalizing it towards the correct statistics.

To overcome the insufficient editing strength, we first rephrase the noise-inversion approach, demonstrate that it bears similarities to Delta-Denoising methods~\cite{hertz2023delta}, and that under certain conditions the DDPM-inversion and Delta-Denoising approaches are exactly equivalent. This re-phrasing not only provides additional insight as to why DDPM noise-inversion approaches are successful, but it reveals that the inversion process itself can be skipped and replaced with an evaluation of a single correction term at each denoising step. Hence, it can be combined into the same batch as the backwards denoising steps themselves, further decreasing editing times.
Finally, we identify specific terms in the edit-friendly denoising process that are responsible for the impact of the prompt, and strengthen them in a similar approach to classifier-free guidance (CFG, \cite{ho2021classifier}). We show that under commonly occurring assumptions, this pseudo-guidance approach is equivalent to re-introducing CFG into the fast-sampling method, but using fewer network evaluation steps.

By combining these components, we enable real-image editing with as few as $3$ diffusion steps, achieving $
\times5$-$\times500$ speedup compared with existing editing methods, while preserving and even improving their output quality. Our code will be made public.

\section{Related Work}

\paragraph{\textbf{Fast Diffusion Sampling.}}
Early diffusion models~\cite{ho2020denoising,sohl2015deep} were notoriously slow to sample from, requiring hundreds of expensive neural evaluations (and several minutes) to produce a single image. To overcome this hurdle, a range of advanced sampling methods were proposed~\cite{song2020denoising,lu2022dpmsolver,lu2023dpmsolverpp,liu2022pseudo,karras2022elucidating}. These are typically grounded in ordinary differential equation solvers, and can successfully reduce the sampling process to several dozen steps without modifying the underlying denoising model.

To further reduce sampling times, a recent line of work proposes to employ a distillation process~\cite{salimans2022progressive}, where a pre-trained diffusion model is fine-tuned using objectives that promote sampling in few ($1$-$8$) steps. These can range from the use of adversarial~\cite{goodfellow2014generative} losses~\cite{sauer2023adversarial,lin2024sdxllightning} to distribution matching~\cite{yin2024onestep} and consistency objectives~\cite{song2023consistency,luo2023lcmlora,luo2023latent,kim2024consistency}. 

The emergence of these fast-sampling methods offers an opportunity to speed up existing workflows. However, \textit{existing} control and editing methods often struggle in this few-step regime. Hence, several approaches were proposed to enable additional controls~\cite{xiao2023ccm,parmar2024one} or improve personalization~\cite{gal2024lcmlookahead,guo2024pulid} using few-step models. Our work seeks to bring another crucial component into this fast-sampling realm - the text-based image editing workflow.

\paragraph{\textbf{Text-based image editing.}}
The unprecedented semantic control offered by large scale text-to-image diffusion models~\cite{rombach2021highresolutionLDM, podell2024sdxl,ramesh2022hierarchical} has inspired a large volume of work that leverages them as a prior for image editing.
Early works proposed simple approaches, such as adding noise to an image and removing it conditioned on a novel prompt~\cite{meng2022sdedit}. However, such methods often lead to significant changes in the image shape and layout. Hence, more advanced methods proposed to use in-painting and other localization approaches~\cite{avrahami2023blendedlatent,nichol2021glide,patashnik2023localizing,brack2023ledits++} to modify a single image region.

Others works align attention maps~\cite{hertz2022prompt,mokady2022null} or other internal feature representations~\cite{parmar2023zeroshot,tumanyan2022plug} to better preserve the original image structure. The outputs of such methods can also be distilled into another diffusion model, trained to modify a conditioning image based on an instruction prompt~\cite{brooks2022instructpix2pix}.
In another approach, the diffusion model itself is fine-tuned on the source image~\cite{kawar2022imagic,valevski2023unitune} to better align it to the original content. However, such approaches are often costly in both time and compute.
Finally, a recent self-attention sharing approach~\cite{cao2023masactrl} has proven effective in maintaining image content across prompts~\cite{tewel2024trainingfree}, and in transferring styles~\cite{hertz2023style} or appearances~\cite{alaluf2023cross} across images.

Common to all these approaches is that they operate in a many-step regime, \ie they typically require dozens of diffusion steps, and struggle when applied to fast sampling models~\cite{parmar2024one}. We aim to enable text-based editing in the fast sampling regime.

\paragraph{\textbf{Diffusion inversion.}}
Applying editing techniques to a real image commonly requires one to first find some latent representation of the image, which can be fed into the model in order to reconstruct the image. This latent can then be perturbed, directly or through modifications of the generative pass, to affect a change in the image.

Initial inversion efforts focused on GANs~\cite{goodfellow2014generative}, opting for either direct optimization~\cite{zhu2016generative,abdal2019image2stylegan,abdal2020image2stylegan++,zhu2020improved} or encoder-based approaches~\cite{richardson2020encoding, alaluf2021restyle, tov2021designing, alaluf2021hyperstyle, dinh2022hyperinverter, parmar2022spatially}. With the rise of diffusion-based editing, several works sought to invert real images into the diffusion space, often by determining some initial noise that will be cleaned into a specific image. Initial inversion approaches overwrote the low-frequency content of a generated image with the low-pass filtered content from a source image~\cite{choi2021ilvr}, enabling scribble-based modifications or texture changes. Others relied on inverting the deterministic DDIM~\cite{song2020denoising} process~\cite{dhariwal2021diffusionBeatsGAN}, but these commonly require many steps to be accurate, and struggle to modify the image through inference-time prompt changes~\cite{mokady2022null}. To overcome this issue, several works intervene in the CFG process and replace the null text condition with a learned embedding that represents the original image~\cite{mokady2022null, miyake2023negative,han2023improving}. However, these commonly require lengthy optimization. As an alternative to optimization, DDIM-based inversion can also be improved by leveraging a fixed-point iterative scheme~\cite{Pan_2023_ICCV, meiri2023fixed,garibi2024renoise}, 
but such solutions require dozens of inversion steps even when attached to a fast-sampling backwards process~\cite{luo2023lcmlora}.

Rather than focusing on predicting an initial noise that will re-create the image with deterministic sampling, an alternative approach is to consider the DDPM generative process, and invert the image into the intermediate noise maps that are added to a generated image~\cite{HubermanSpiegelglas2023,wu2023cyclediffusion}. This approach serves as our backbone for few-step editing, and we expand on it in greater detail in \cref{sec:ddpm_inversion}.

Finally, some methods invert sets of images into the text conditioning space of the model ~\cite{gal2022image,gal2023designing,alaluf2023neural,dong2022dreamartist,voynov2023p+,zhang2023prospect}. However, these are typically used for personalization, where the goal is not to preserve the structure of an image, but to learn a global representation that allows to re-create a concept in novel scenes.
\section{Preliminaries}\label{sec:ddpm_inversion}

We begin with a high level overview of the DDPM noise-inversion approach of \citet{HubermanSpiegelglas2023}. There, the goal is to find a meaningful latent representation that can be used to reconstruct an image, such that this latent can be manipulated in more intuitive ways than with existing approaches like DDIM inversion. To this end, the authors propose to use the DDPM-noise space (\ie the noise maps added to the image at each step of the DDPM denoising process), and term it an ``edit-friendly'' noise space.

More concretely, recall the DDPM denoising equation:
\begin{equation}
\label{eq:diffusion}
    x_{t-1} = \mu_t(x_t, c) + \sigma_t z_t,\quad t=T,\ldots,1,
\end{equation}
where $z_t\sim\mathcal{N}(0,\boldsymbol{I})$, $c$ is a conditioning prompt, $\mu_t(x_t, c)$ is derived from the denoising network's output through:
\begin{equation}
\label{eq:mu_prediction}
\mu_t(x_t, c) = \frac{1}{\sqrt{\alpha_t}}\left(x_t - \frac{1-\alpha_t}{\sqrt{1-\bar\alpha_t}} \epsilon_\theta(x_t, t, c)\right),
\end{equation}
where $\epsilon_\theta(x_t, t, c)$ is the noise prediction and $\alpha_t,\sigma_t$ are derived from the noising schedule.

In the noise-inversion process, given an image $x_0$, one first calculates its noisy representations through the standard forward diffusion process equation, using a different independently-sampled noise for each time-step:
\begin{equation}
\label{eq:xt_from_x0_iid}
x_t = \sqrt{\bar\alpha_t} x_0 + \sqrt{1-\bar\alpha_t}\, \tilde{\epsilon}_t,\quad 1,\ldots,T,
\end{equation}
where $\tilde{\epsilon}_t\sim\mathcal{N}(0,\boldsymbol{I})$ are \emph{statistically independent}. This independence is a core difference between the ``edit-friendly'' approach and prior noise inversion approaches (CycleDiffusion, ~\cite{wu2023cyclediffusion}), and the authors show that it is crucial for achieving pleasing results.

Given two such noisy images, $x_t$ and $x_{t-1}$, one then calculates the noise that would be needed for \cref{eq:diffusion} to clean $x_t$ into $x_{t-1}$:
\begin{equation}\label{eq:edit_friendly_corrections}
        \sigma_t z_t = x_{t-1} - \mu_t(x_t, c)
\end{equation}

Finally, to perform text-based editing, one can simply denoise an image from the pre-calculated $x_T$ under a novel prompt $\hat c$, while applying the inverted noises at each step in place of the typical DDPM noise samples, i.e.:
\begin{equation}\label{eq:edit_friendly_inference}
        \hat x_{t-1} = \mu_t(\hat x_t, \hat c) + \sigma_t z_t = \mu_t(\hat x_t, \hat c) + x_{t-1} - \mu_t(x_t, c) .
\end{equation}

Throughout the rest of the paper, we make use of this ``edit-friendly'' DDPM noise-inversion technique. We demonstrate how it can be adapted to work with a few-step SDXL-Turbo model, and provide additional insights on the reason behind its success.

\section{Method}

In the following section, we outline a series of modifications that can enable the DDPM noise-inversion editing flow to operate in the few-step diffusion domain, and specifically with SDXL-Turbo. We analyze the two primary issues with the baseline results - the appearance of visual artifacts, and insufficient editing strength, and suggest a fix for each. Throughout the process, we shed some additional light on the DDPM-inversion process itself. %

\subsection{Treating the visual artifacts}
\begin{figure}[ht]
    \centering
    \includegraphics[width=0.97\linewidth]{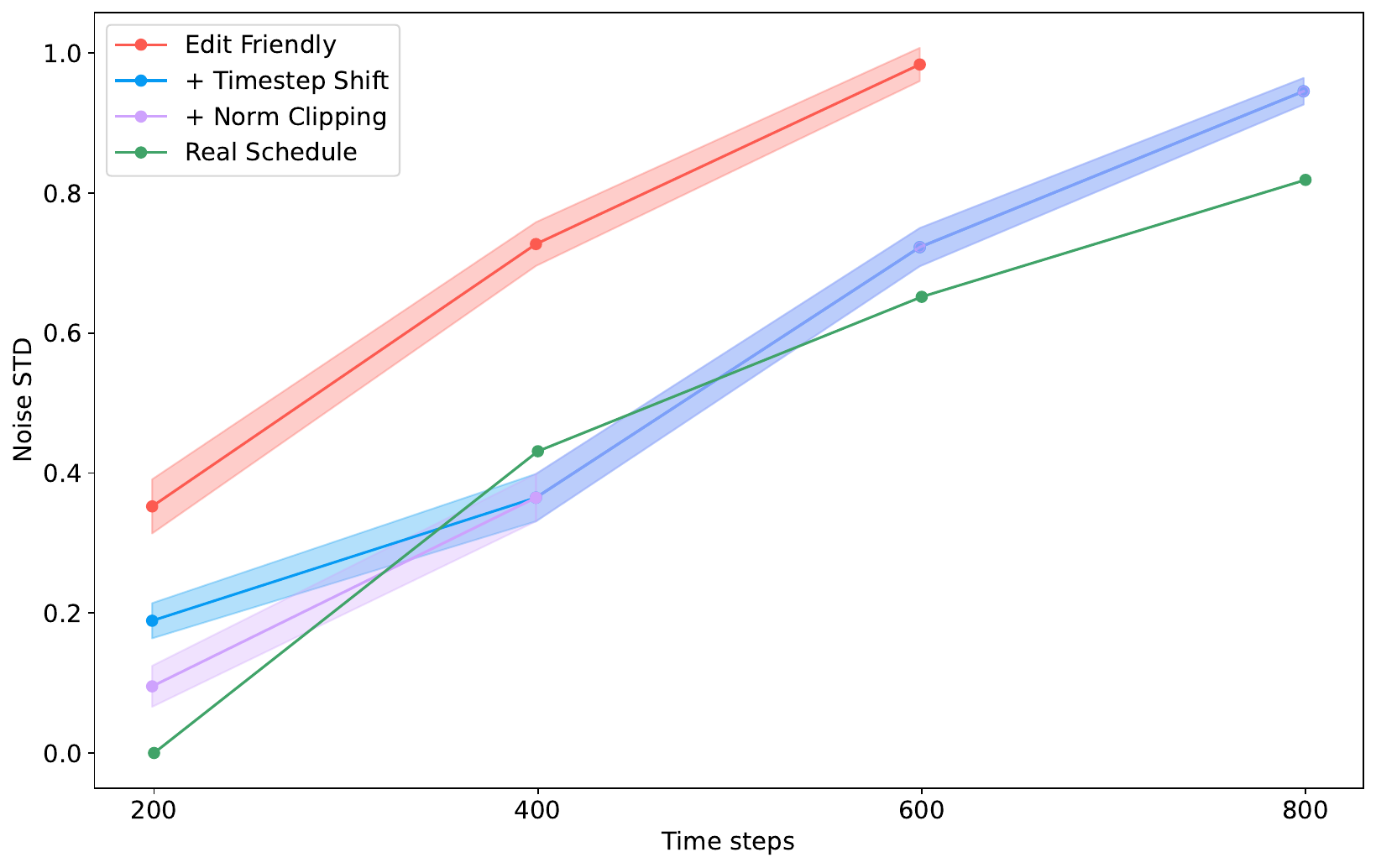}\vspace{-3pt}
    \caption{Comparison of the pixel-wise standard deviations of inverted noise maps, and the expected distribution. The scale of corrections predicted by standard edit-friendly DDPM inversion (red, \cref{eq:edit_friendly_corrections}) is consistently higher than the expected noise schedule (green). The higher values approximately align with a shift along the x-axis: \ie, edit-friendly noise scales align with earlier steps in the diffusion process. We thus propose a time-shifted inversion schedule, where the image is cleaned ``as-if'' it belonged to a time-point aligning with its noise scale, rather than the real step. In practice, shifting the schedule by a constant $200$ steps serves to provide good alignment (blue) and resolve most artifacts. To correct the statistics of the last step, we further apply norm-clipping to the predicted noise at that stage (purple). Shaded regions indicate the 68\% confidence interval.
    }\label{fig:z_t_std}\vspace{-3pt}
\end{figure}

As noted above, directly applying the DDPM noise-inversion approach to SDXL-Turbo leads to considerable artifacts in the generated images (see \cref{fig:ours_vs_few_steps,fig:ablation_study}). 
In their original work, \citet{HubermanSpiegelglas2023} observe that their inverted noises follow different dynamics than the standard Gaussian noise used in DDPM generation. Drawing on their observations, we hypothesize that the emergence of visual artifacts can be traced back to such deviations. To verify this hypothesis, we analyze the statistics of the noises derived from a noise-inversion process, and compare them with the scale of noises injected during a standard DDPM generation flow. Specifically, we apply standard edit-friendly inversion using an SDXL model with a few-step schedule, and compare the standard deviation of 
$x_{t-1} - \mu_{t}{(x_t)}$ \ie the DDPM-inversion corrections,
with $N(0, \sigma_t ^2)$, \ie the standard DDPM noise. Results are in \cref{fig:z_t_std}. 

An immediate observation is that the per-pixel standard deviations of the noise-inversion corrections are higher than those of the DDPM-noise schedule, throughout the entire process. Importantly, this contrasts with the dynamics observed in many-step scenarios~\cite{HubermanSpiegelglas2023}, where the statistics eventually converge towards the end of the generation process. A second observation, is that the noise statistics seem to deviate by an approximately constant margin. In other words, we notice that throughout much of the process, the inverted noises behave like a standard noise originating from a time-step shifted by roughly $200$ steps to the past (see also \cref{fig:offset_histogram} in the supplementary). All-in-all, this disagreement between the noise statistics and the timestep used to condition the model and the denoising step, leads to a train-test mismatch and results in the creation of artifacts.

Following these observations, we propose to alleviate the artifacts by re-aligning the denoising timestep schedule to the scale of the noises. Specifically, during the inversion process, for each timestep $t$ we first add noise to $x_0$ according to the standard DDPM noising schedule, and create $x_t$. However, we inject a larger timestep $t+\Delta$ into the denoising model and the scheduler's denoising step:
\begin{equation}
    \sigma_t \cdot z_t = x_{t-1} - \mu_{t +\Delta}{(x_t, c)},
\end{equation}
where the subscript $t +\Delta$ denotes the timestep input of both the denoising network and the scheduler. By applying this correction, we can once again compare the statistics of the inverted noises with standard noise schedule (\cref{fig:z_t_std}, blue) and see that they are in much better agreement across all timesteps.

Note that for the generative pass, we must also employ a similar shift to keep the two network prediction components synchronized: 
\begin{equation}
    \hat{x}_{t-1} = \mu_{t +\Delta}{(\hat{x}_t, \hat{c})} + \left(x_{t-1} - \mu_{t +\Delta}{(x_t, c)}\right) .
\end{equation}
This synchronization is crucial for our analysis in \cref{sec:pseudo_cfg}.

As an additional correction, we find it beneficial to clip the norm of the last noise-inversion correction (\cref{fig:z_t_std}, purple). The motivation here arises from the contrast with multi-step approaches, where the last correction is small and can be skipped. Here, the last step is large, and its associated correction still captures many of the details of the original image. Hence, we do not want to discard it. Instead, we simply clip it to avoid the introduction of novel artifacts.
This is reminiscent of other editing methods, e.g. \cite{hertz2022prompt} where the control is decreased towards the end of the generation process.

\subsection{Improving prompt alignment}\label{sec:pseudo_cfg}
Having treated the visual artifacts, we must now deal with the second limitation: insufficient editing strength. The issue here is that changing the prompt between the noise-inversion step and the generation pass, often leads to little or no change in the final image. 

To investigate this limitation, we once again consider the edit-friendly inference formula (\cref{eq:edit_friendly_inference}), which we can re-write as:
\begin{equation}\label{eq:ef_update}
    \hat{x}_{t-1} = x_{t-1} + \left(\mu_{t}{(\hat{x}_t, \hat{c})} - \mu_{t}{(x_t, c)}\right) .
\end{equation}
Under this framing, we can already see a hint to an underlying reason behind the noise-inversion approach's success. Specifically, the second term captures the difference between the model's prediction on the edited image, using the novel prompt, and the same prediction on the original image using the original prompt that describes it. This is analogous to the correction term found in Delta Denoising Score (DDS, \cite{hertz2023delta}) where the authors show that the performance of score distillation sampling methods (SDS, \cite{poole2023dreamfusion}) on image editing can be significantly improved if one computes a loss using a similar difference between the network's prediction on an edited image with the new prompt, and the prediction on the original image with the original prompt that describes it. The intuition provided for such a term in DDS (and later expanded on in NFSD~\cite{katzir2023noisefree}), is that this difference helps cancel out components of the denoising network's prediction that are unrelated to the prompt (e.g., the direction towards a cleaner image, or any general network bias). The existence of a similar term in the edit-friendly process implies that the same mechanism contributes to its improved performance in image editing tasks. Indeed, we will later show that the connection between EF and DDS is much deeper, with the two methods being functionally equivalent.

We can further expand this analysis of the edit-friendly process by adding and subtracting a cross-term:
\begin{equation}\label{eq:cross_terms}
    \hat{x}_{t-1} = x_{t-1} + \left(\overbrace{\mu_{t}{(\hat{x}_t, \hat{c})} - \mu_{t}{(\hat{x}_t, c)}}^{cross-prompt}  + \overbrace{\mu_{t}{(\hat{x}_t, c)} - \mu_{t}{(x_t, c)}}^{cross-trajectory}\right) .
\end{equation}
Now, we can identify two different directions of change to the original image. The first term is the difference between predictions for the generation trajectory under the novel and original prompts. Hence, one can regard this term as one that represents the direction that takes an image on the novel trajectory from the old prompt to the new one. The second term is the difference between predictions for the new trajectory and the old one, under the same prompt. Hence, it can be regarded as the direction that takes an image from the old trajectory, to the new one. Intuitively, applying both of them to $x_{t-1}$, itself a point on the original trajectory, will first shift it to the new trajectory and then pull it further in the direction of the difference between prompts.

\begin{figure}
\centering
\setlength{\tabcolsep}{3pt}
\begin{tabular}{c c c c c}

    & \small $w_p=1$ & \small $w_p=1.5$ & \small $w_p=2$ & \\
    &
    \includegraphics[width=0.1\textwidth,height=0.1\textwidth]{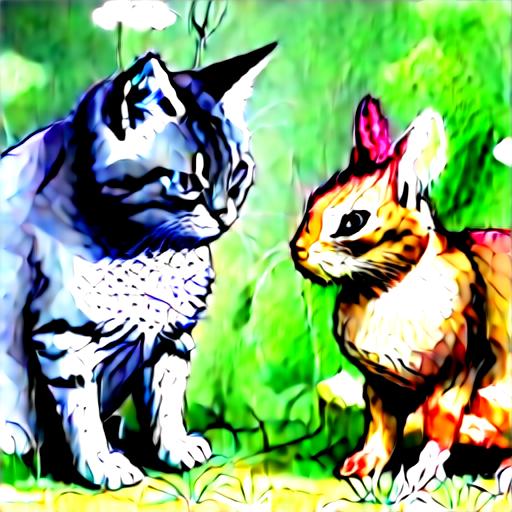} &
    \includegraphics[width=0.1\textwidth,height=0.1\textwidth]{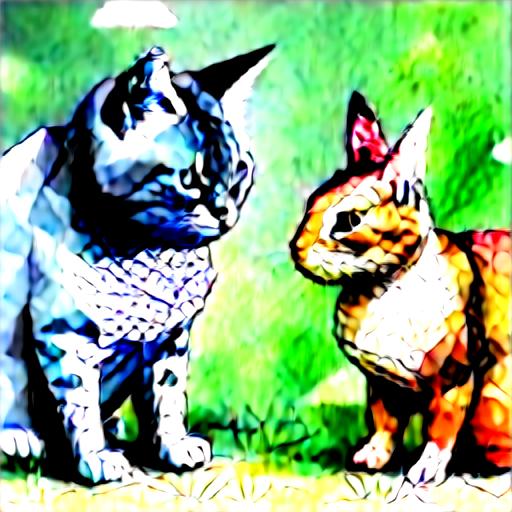} &
    \includegraphics[width=0.1\textwidth,height=0.1\textwidth]{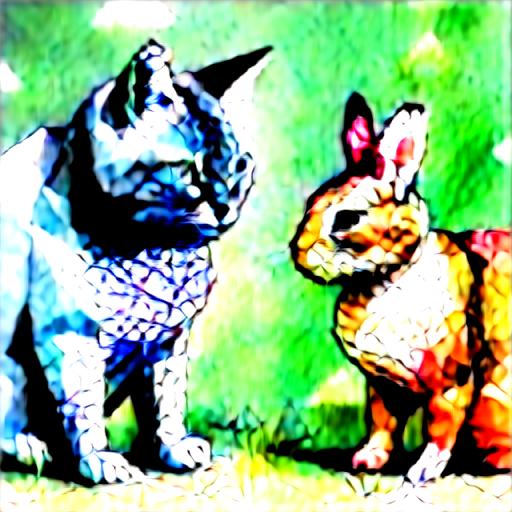} &

    \raisebox{0.04\textwidth}{\rotatebox[origin=t]{-90}{\scalebox{0.9}{\begin{tabular}{c@{}c@{}c@{}} $w_t=2$ \end{tabular}}}} \\ 

    &
    \includegraphics[width=0.1\textwidth,height=0.1\textwidth]{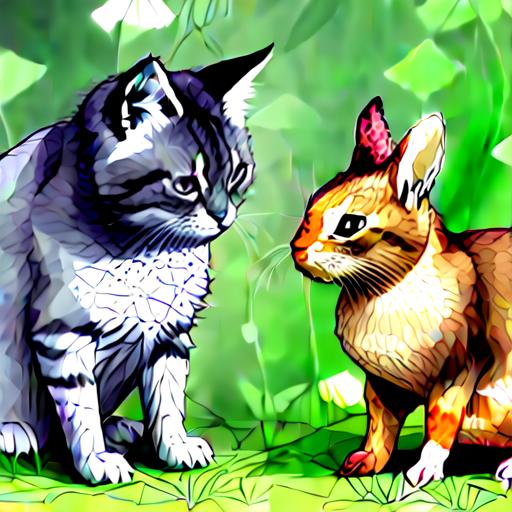} &
    \includegraphics[width=0.1\textwidth,height=0.1\textwidth]{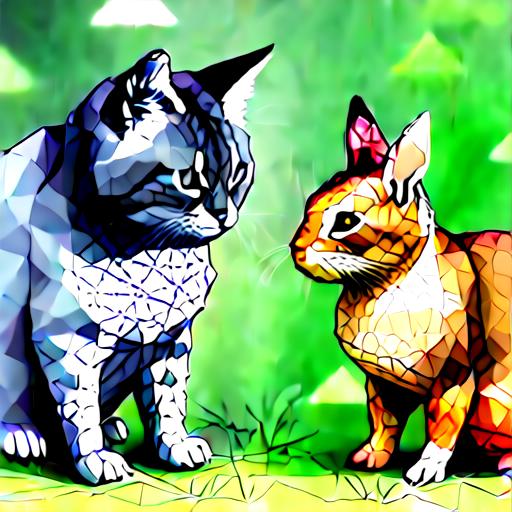} &
    \includegraphics[width=0.1\textwidth,height=0.1\textwidth]{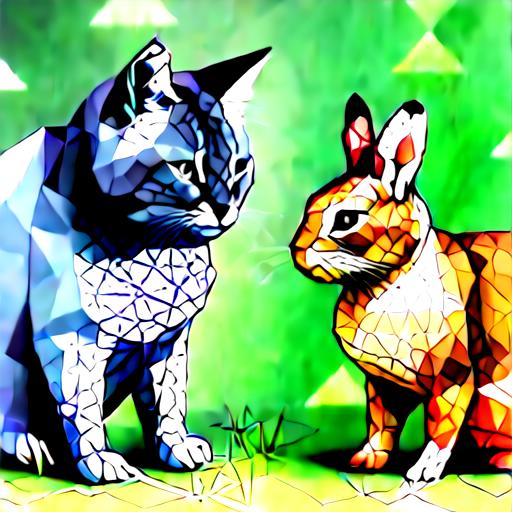} &

    \raisebox{0.04\textwidth}{\rotatebox[origin=t]{-90}{\scalebox{0.9}{\begin{tabular}{c@{}c@{}c@{}} $w_t=1.5$ \end{tabular}}}} \\ 

    \includegraphics[width=0.1\textwidth,height=0.1\textwidth]{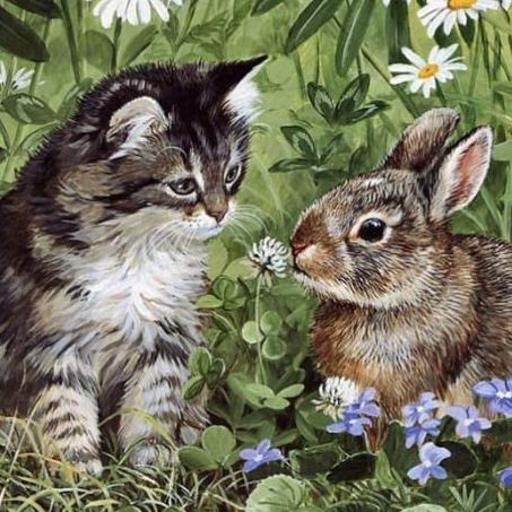} &

    \includegraphics[width=0.1\textwidth,height=0.1\textwidth]{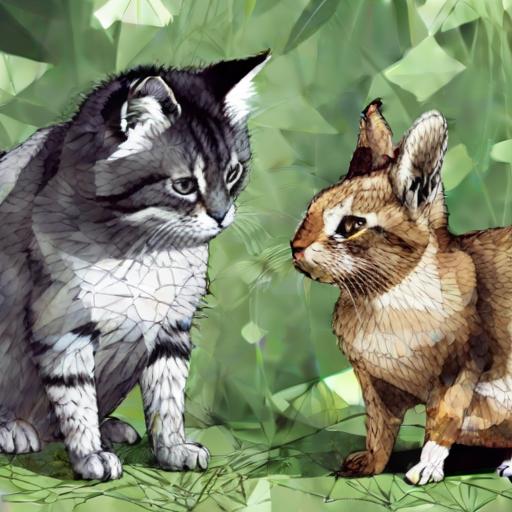} &
    \includegraphics[width=0.1\textwidth,height=0.1\textwidth]{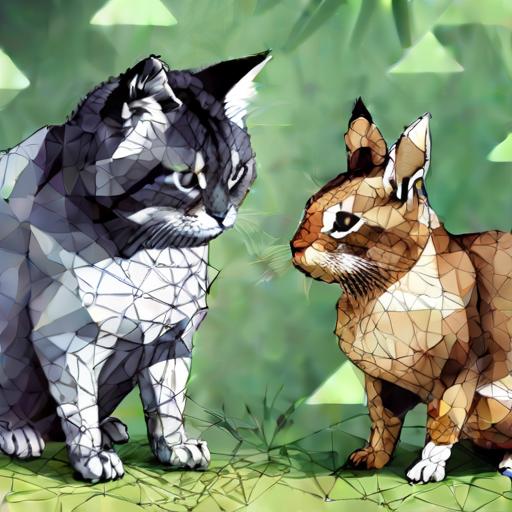} &
    \includegraphics[width=0.1\textwidth,height=0.1\textwidth]{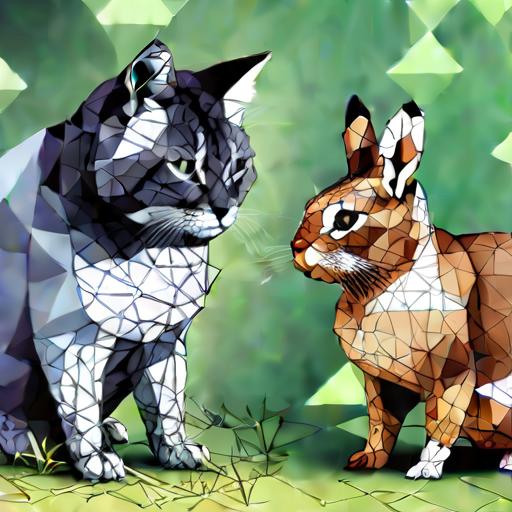} &

    \raisebox{0.04\textwidth}{\rotatebox[origin=t]{-90}{\scalebox{0.9}{\begin{tabular}{c@{}c@{}c@{}} $w_t=1$ \end{tabular}}}} \\
    \small Input Image & \multicolumn{3}{c}{\small ``a polygonal illustration of a cat and a bunny"} & \\

\end{tabular}
\caption{We show the effect of increasing the strength of the cross-prompt term ($w_p$) and cross-trajectory term ($w_t$) in the DDPM inversion.
While both terms can help increase the condition in the edited image, as we increase the cross-trajectory term we see artifacts and saturation.}\label{fig:scaling_terms}\vspace{-5pt}
\end{figure}

Recall that our goal was to strengthen the effect of the prompt. Hence, we can simply draw inspiration from CFG and extrapolate along this cross-prompt direction. Importantly, we do not want to increase the weight of the cross-trajectory term, because we do not want to overshoot the new trajectory. Indeed, in \cref{fig:scaling_terms} we investigate the behaviour of these two terms and demonstrate that scaling the cross-trajectory term leads to the creation of novel visual artifacts and increased saturation, while scaling the cross-prompt term leads to stronger editing effects. As such, we propose to convert the edit-friendly inference equation as follows:
\begin{equation}
    \hat{x}_{t-1} = x_{t-1} + \mu_{t}{(\hat{x}_t, c)} - \mu_{t}{(x_t, c)} + w \cdot \left(\mu_{t}{(\hat{x}_t, \hat{c})} - \mu_{t}{(\hat{x}_t, c)}\right) ,
    \label{eq:pseudo_guidance}
\end{equation}
where $w$ is the pseudo-guidance scale.

In the supplementary, we further analyze the connection between this guidance formula and a re-introduction of CFG into SDXL-Turbo. Specifically, we show that our approach is equivalent to applying CFG during both inversion and inference, so long as:
\begin{equation}\label{eq:cfg_equivalence}
\mu_{t}{(\hat{x}_t, c)} - \mu_{t}{(x_t, c)} \approx \mu_{t}{(\hat{x}_t, \phi)} - \mu_{t}{(x_t, \phi)},
\end{equation}
where $\phi$ indicates the null prompt. Experimentally, we find that this condition commonly holds for SDXL-Turbo predictions. Specifically, we find an average cosine similarity of $0.93$ between the two sides of \cref{eq:cfg_equivalence}, compared to an averaged cosine similarity of $-0.04$ between these two sides and the cross-prompt term of \cref{eq:cross_terms}.

Hence our approach roughly aligns with a re-introduction of CFG into SDXL-Turbo, but requires fewer neural function evaluations ($3$ instead of $4$), or equivalently a smaller batch size. In both cases, this allows us to gain further speed improvements over existing approaches. Moreover, this result sheds further light on the observations of \citet{HubermanSpiegelglas2023}, which report better performance when applying CFG during both the inversion and the generation process with multi-step models. If CFG is only applied to one pass and not the other, then CFG also increases the weight of the cross-trajectory term which may harm the results.

Finally, while the derivation of  \cref{eq:pseudo_guidance} was rooted in the ``edit-friendly'' DDPM-inversion process, we note that its evaluation does not require any pre-computation of a multi-step noise map. Indeed, since neither $x_{t-1}$ nor $\mu_{t}{(x_t, c)}$ depend on a multi-step denoising pass (they depend only on a closed-form projection of $x_0$ via \cref{eq:xt_from_x0_iid}), they can simply be calculated at inference-time, with $\mu_{t}{(x_t, c)}$ being predicted as part of the same batch. This approach is more closely related to CycleDiffusion's~\cite{wu2023cyclediffusion} denoising pass, and can effectively cut the number of editing steps by half.

\subsection{Connecting EF and DDS}\label{sec:ef_dds_duality}
In \cref{sec:pseudo_cfg} we observed edit-friendly inversion and DDS share a similar structure. In both cases, the image is being edited by shifting it along the vector between the diffusion predictions on the new trajectory with the new prompt, and the diffusion predictions on the original image with a prompt that describes it. In \cref{sec:ef_dds_equivalence} we demonstrate that this connection runs deeper, and that under an appropriate choice of timesteps and learning rates during the DDS optimization process, both methods become functionally equivalent, with the DDS corrections terms collapsing to \cref{eq:edit_friendly_inference}. Please see the supplementary for the proof, an empirical demonstration, and a deeper discussion of this equivalency.

\subsection{Implementation details}
We implemented our method on top of SDXL-Turbo~\cite{sauer2023adversarial}. Unless otherwise noted, all results use a pseudo-guidance scale $w=1.5$ and 4 denoising steps (starting at $t=599$ with $\Delta=200$). We clip the norm of the final step corrections to a maximum of 15.5. All experiments are performed on a single NVIDIA A5000 GPU.

\section{Results}

We structure our experimental verification in two parts. In the first part, we demonstrate our methods ability to edit real images in few steps. There, we conduct a series of evaluations, where we contrast our method against a range of baselines. These include both existing multi-step baselines, as well as several few-step alternatives.
We demonstrate that our approach can match, or even exceed the quality of current multi-step editing methods, while being significantly faster.
Then, we conduct an ablation study where we demonstrate the effect of individual components in our proposed solution.

\subsection{Evaluation}

\begin{figure*}
    \centering
    \setlength{\tabcolsep}{1.5pt}
    {\normalsize
    \begin{tabular}{c c c c c c c}
    
        {\small Original Image} & \multicolumn{6}{c}{\small Editing results} \\

        \vspace{-4pt} \includegraphics[width=0.125\textwidth,height=0.125\textwidth]{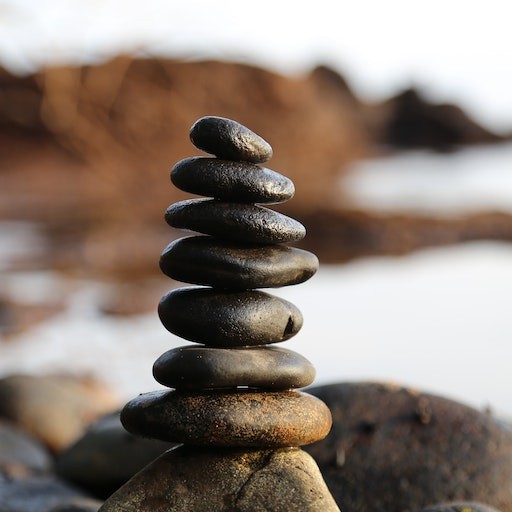} &
        \includegraphics[width=0.125\textwidth,height=0.125\textwidth]{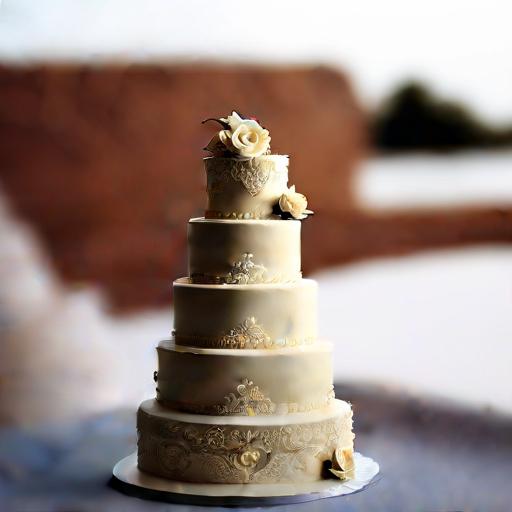} &
        \includegraphics[width=0.125\textwidth,height=0.125\textwidth]{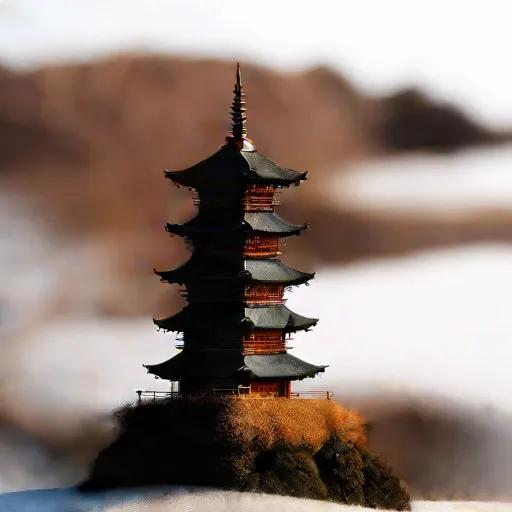} &
        \includegraphics[width=0.125\textwidth,height=0.125\textwidth]{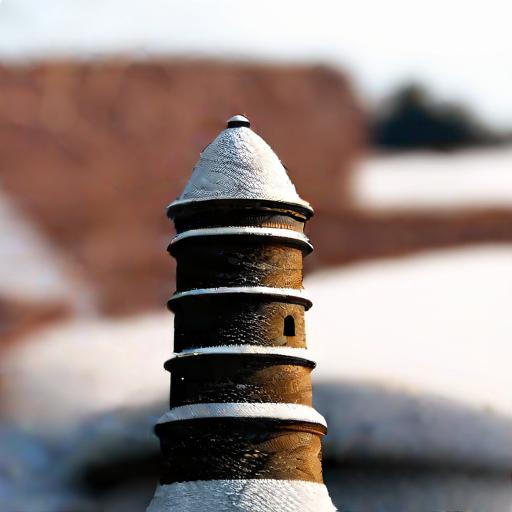} &
        \includegraphics[width=0.125\textwidth,height=0.125\textwidth]{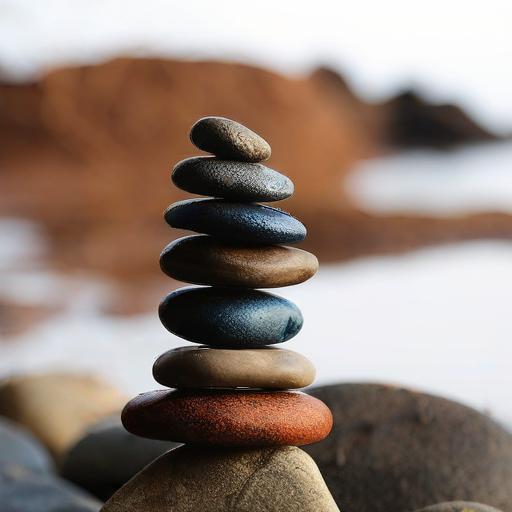} &
        \includegraphics[width=0.125\textwidth,height=0.125\textwidth]{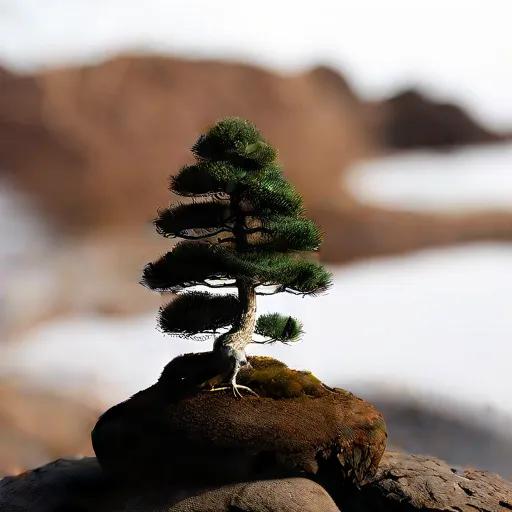} &
        \includegraphics[width=0.125\textwidth,height=0.125\textwidth]{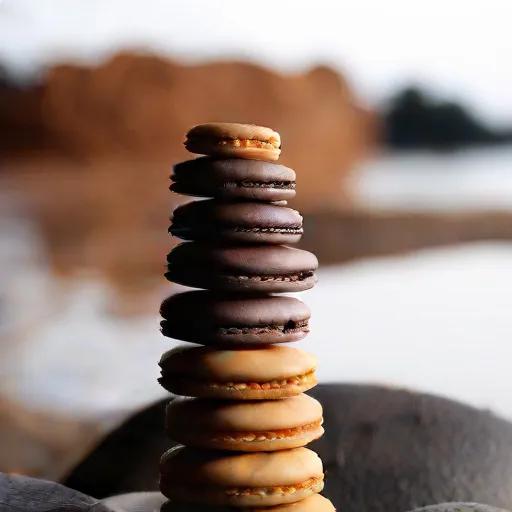} \\

        \vspace{2pt} & \footnotesize  ``White wedding cake'' & \footnotesize ``Pagoda'' & \footnotesize ``Snowy tower'' & \footnotesize ``Colorful'' & \footnotesize ``Bonsai'' & \footnotesize ``Macaroons'' \\
        
        \vspace{-4pt} \includegraphics[width=0.125\textwidth,height=0.125\textwidth]{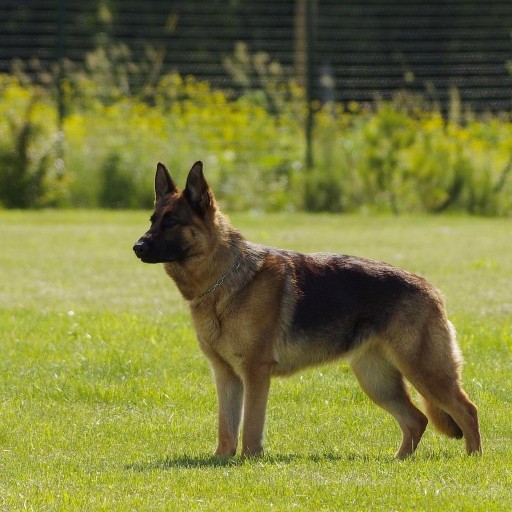} &
        \includegraphics[width=0.125\textwidth,height=0.125\textwidth]{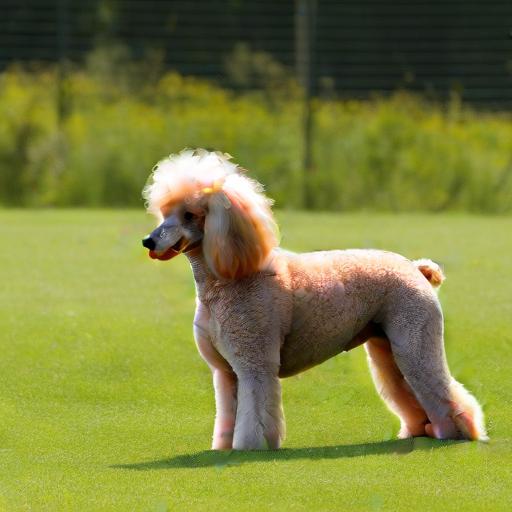} &
        \includegraphics[width=0.125\textwidth,height=0.125\textwidth]{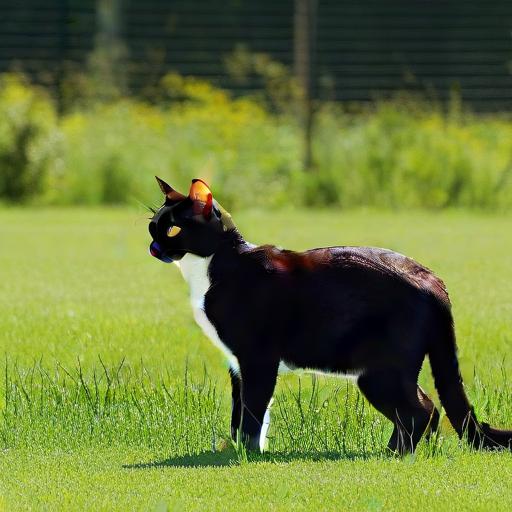} &
        \includegraphics[width=0.125\textwidth,height=0.125\textwidth]{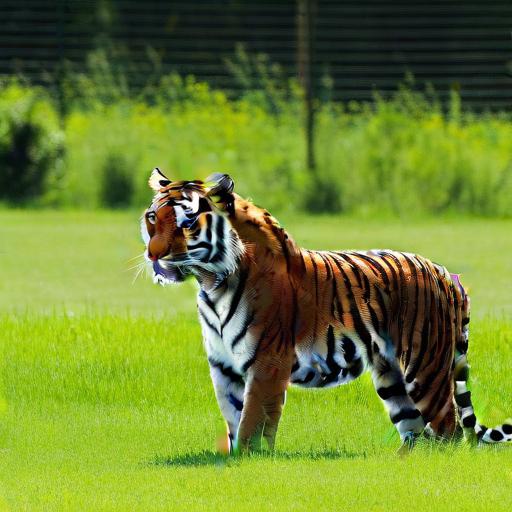} &
        \includegraphics[width=0.125\textwidth,height=0.125\textwidth]{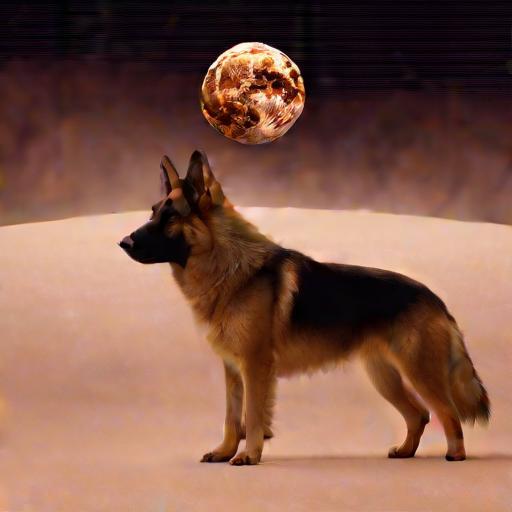} & \includegraphics[width=0.125\textwidth,height=0.125\textwidth]{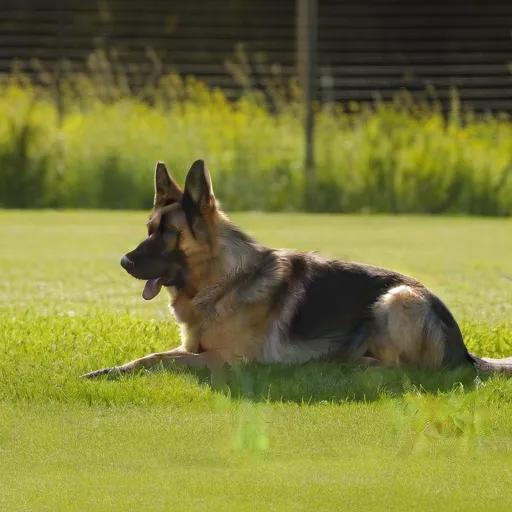}
        & \includegraphics[width=0.125\textwidth,height=0.125\textwidth]{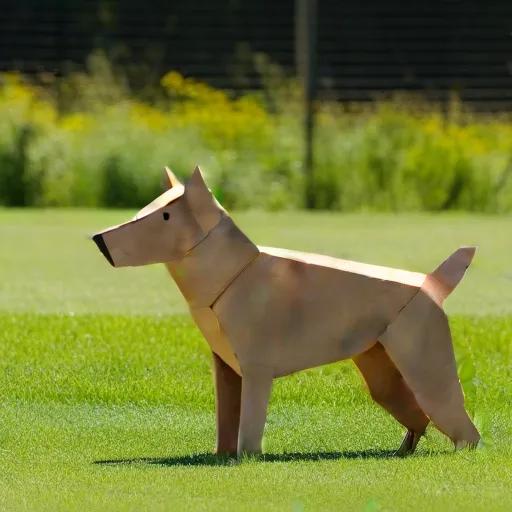}
        \\
        \vspace{2pt} & \footnotesize ``Poodle'' &  \footnotesize ``Cat'' & \footnotesize  ``Tiger'' & \footnotesize ``Full moon'' & \footnotesize ``Sitting'' & \footnotesize ``Cardboard'' \\

        \vspace{-4pt} \includegraphics[width=0.125\textwidth,height=0.125\textwidth]{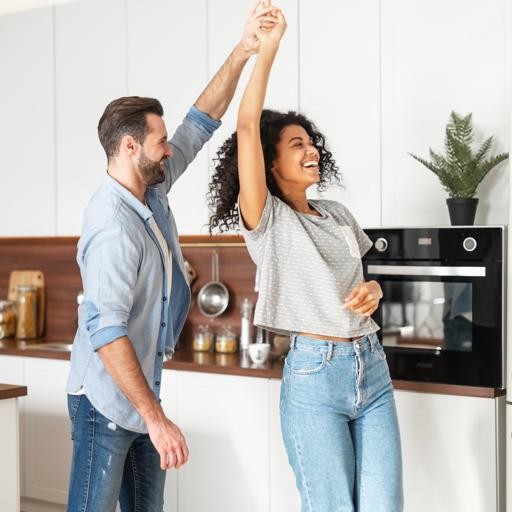} &
        \includegraphics[width=0.125\textwidth,height=0.125\textwidth]{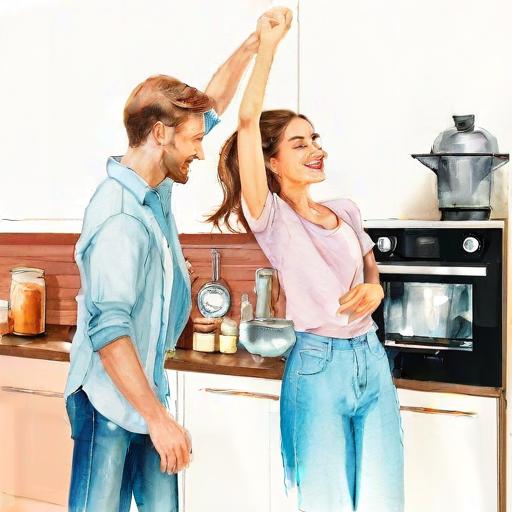} &
        \includegraphics[width=0.125\textwidth,height=0.125\textwidth]{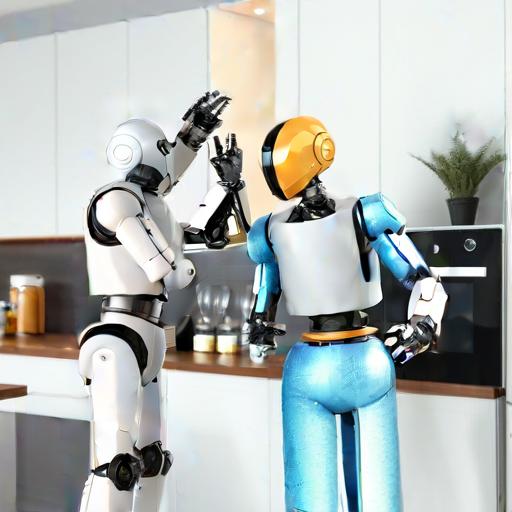} &
        \includegraphics[width=0.125\textwidth,height=0.125\textwidth]{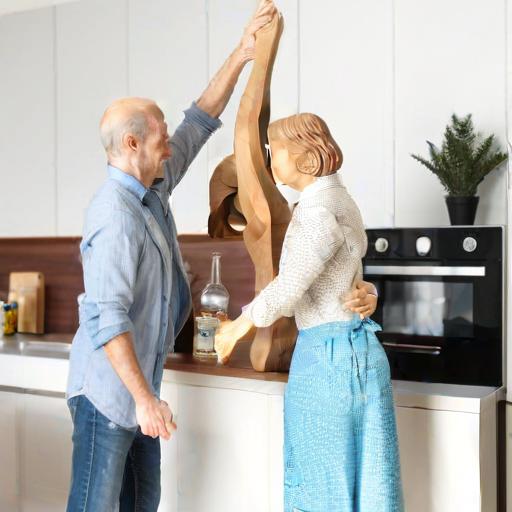} &
        \includegraphics[width=0.125\textwidth,height=0.125\textwidth]{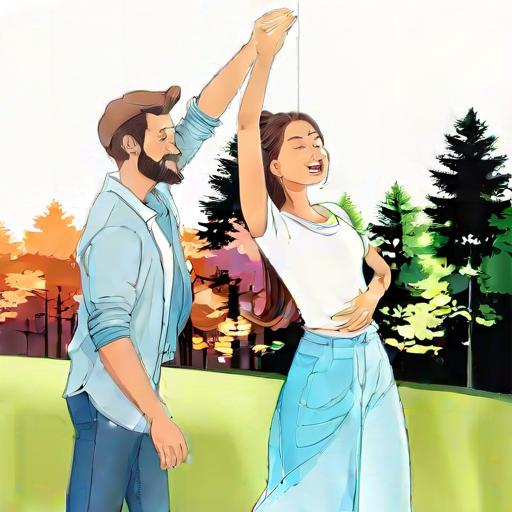} &
        \includegraphics[width=0.125\textwidth,height=0.125\textwidth]{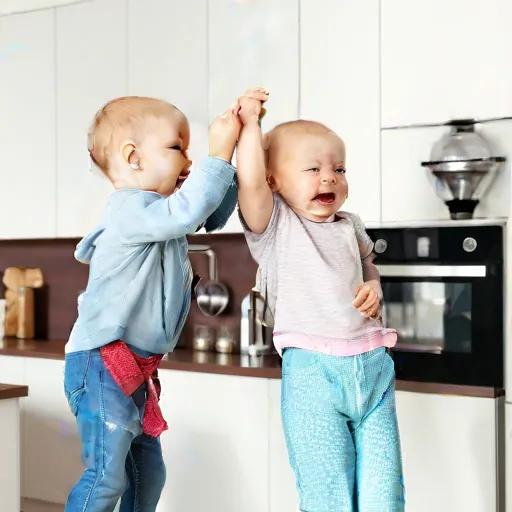} &
        \includegraphics[width=0.125\textwidth,height=0.125\textwidth]{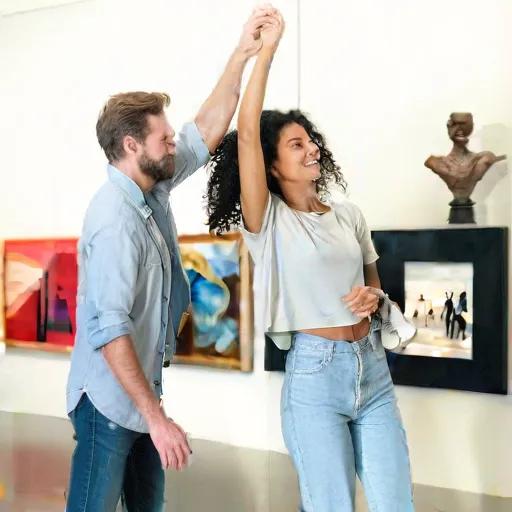} \\

        \vspace{2pt} & {\footnotesize ``Watercolor''} &  \footnotesize ``Robots'' & \footnotesize ``Wooden sculpture'' & {\footnotesize ``In the forest''} & {\footnotesize ``Huge babies''} & {\footnotesize ``Art gallery''} \\

        \vspace{-4pt} \includegraphics[width=0.125\textwidth,height=0.125\textwidth]{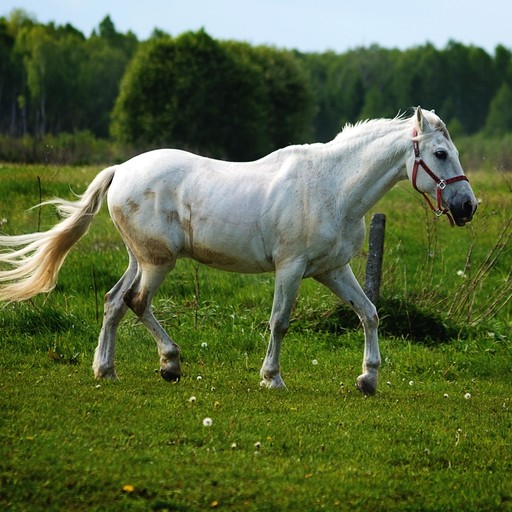} &
        \includegraphics[width=0.125\textwidth,height=0.125\textwidth]{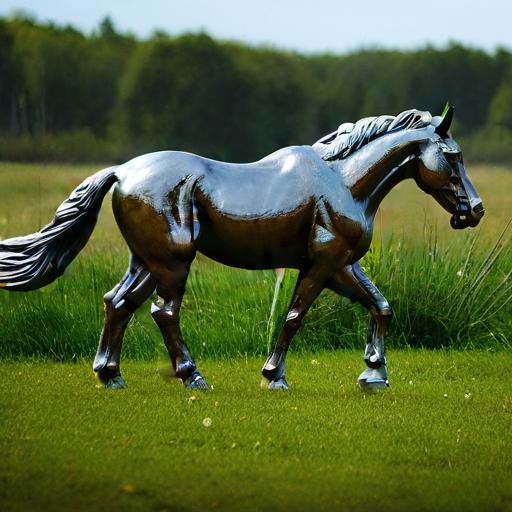} &
        \includegraphics[width=0.125\textwidth,height=0.125\textwidth]{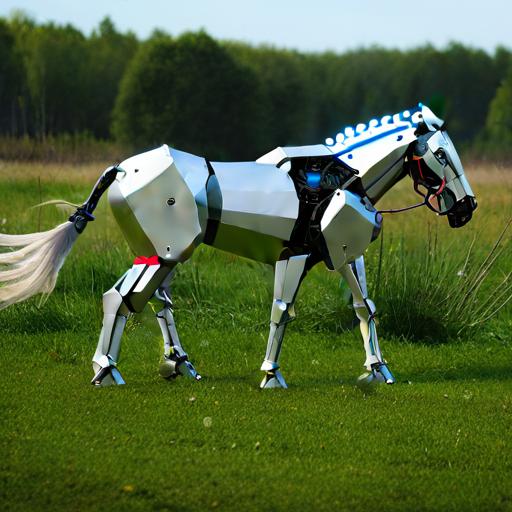} &
        \includegraphics[width=0.125\textwidth,height=0.125\textwidth]{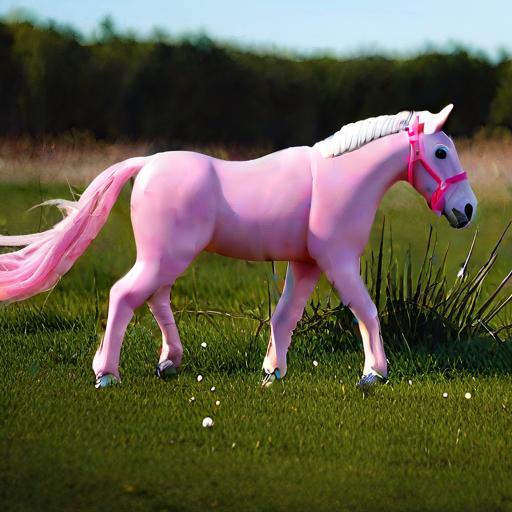} &
        \includegraphics[width=0.125\textwidth,height=0.125\textwidth]{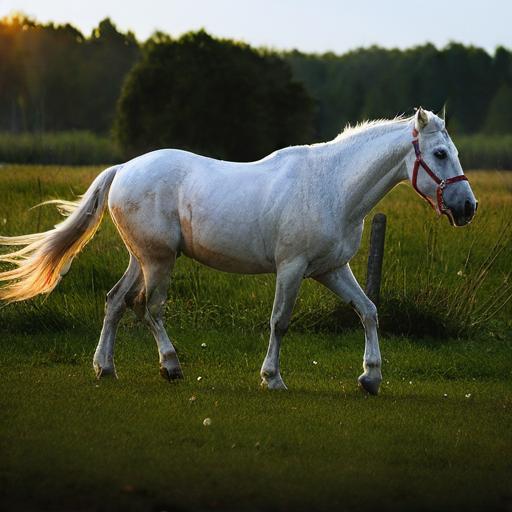} &
        \includegraphics[width=0.125\textwidth,height=0.125\textwidth]{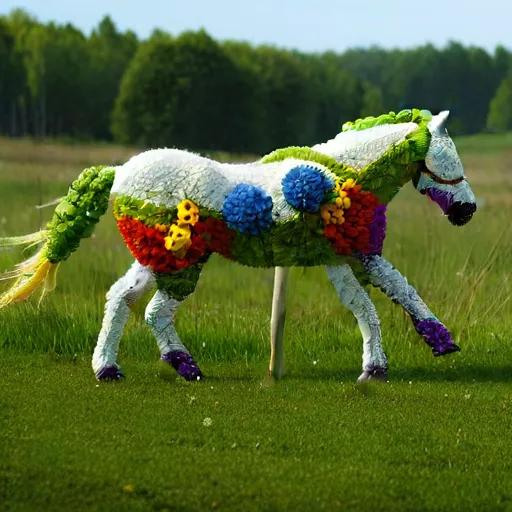} &
        \includegraphics[width=0.125\textwidth,height=0.125\textwidth]{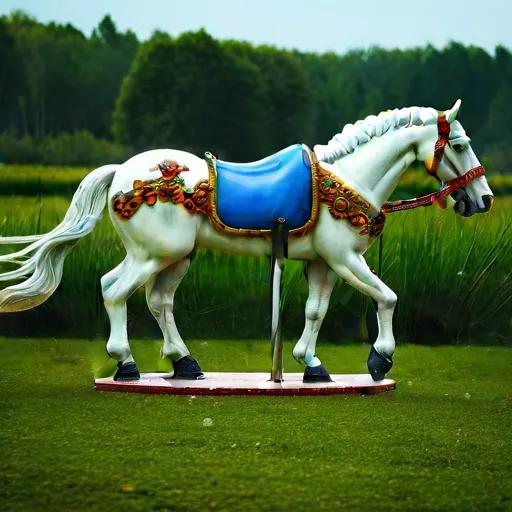} \\

        \vspace{2pt} & \footnotesize ``Bronze'' & \footnotesize ``Robot'' & \footnotesize ``Pink toy'' & \footnotesize ``At sunset'' & \footnotesize ``Made of flowers'' & \footnotesize ``Carousel horse'' \\

    \end{tabular}
    
    }
    \caption{Qualitative editing results of our method. All results use $4$ diffusion steps.}\label{fig:ours_only}
\end{figure*}

We being by evaluating our method, starting with a qualitative evaluation. In \cref{fig:ours_only} we show a range of editing results achieved with our method using $4$ diffusion steps. These include object-level modifications, style changes, or object replacement. Additional results are shown in \cref{fig:ours_only_extra}.

\begin{figure*}
    \centering
    \setlength{\abovecaptionskip}{6pt}
    \setlength{\tabcolsep}{1.5pt}
    {\normalsize
    \begin{tabular}{c c c c c c c c c c}
    
        {\small Input} & {\small Ours (4 steps)} & {\small Ours (3 steps)} & {\small PnP} & {\small Edit-Friendly} & 
        \begin{tabular}{@{}c@{}} \small Edit-Friendly \\ \small + P2P \end{tabular} & {\small ReNoise (1)} & {\small ReNoise (0.75)} &
        \begin{tabular}{@{}c@{}} \small Null-text \\ \small + P2P \end{tabular} & \\

        \includegraphics[width=0.1\textwidth,height=0.1\textwidth]{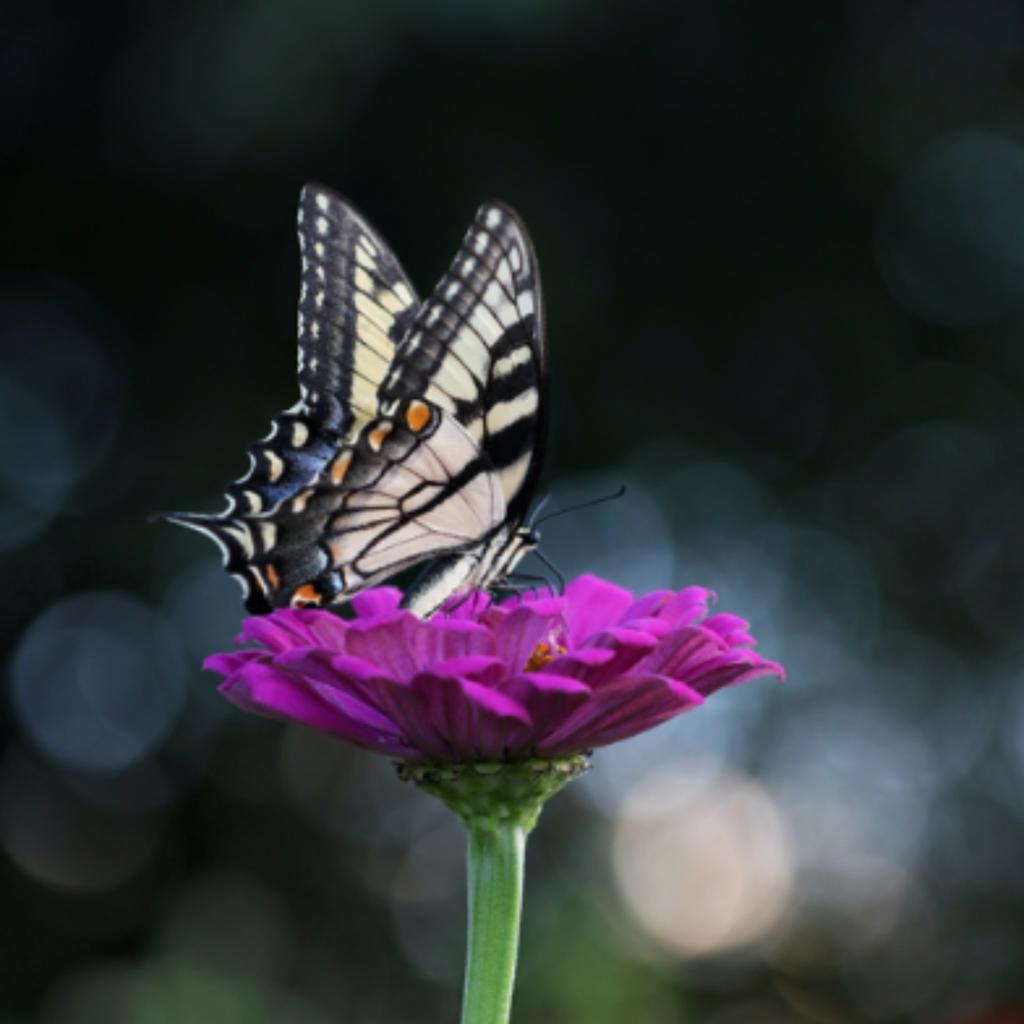} &
        \includegraphics[width=0.1\textwidth,height=0.1\textwidth]{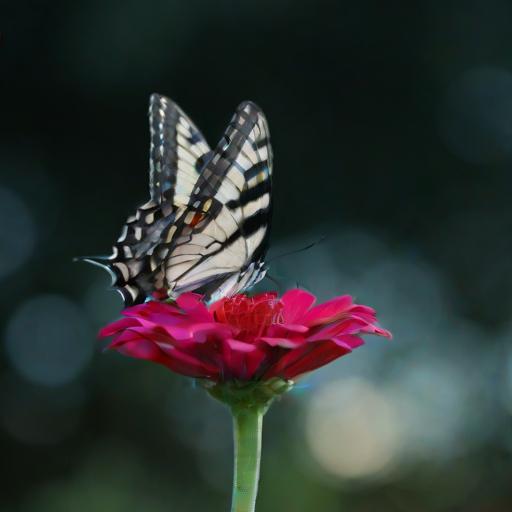} &
        \includegraphics[width=0.1\textwidth,height=0.1\textwidth]{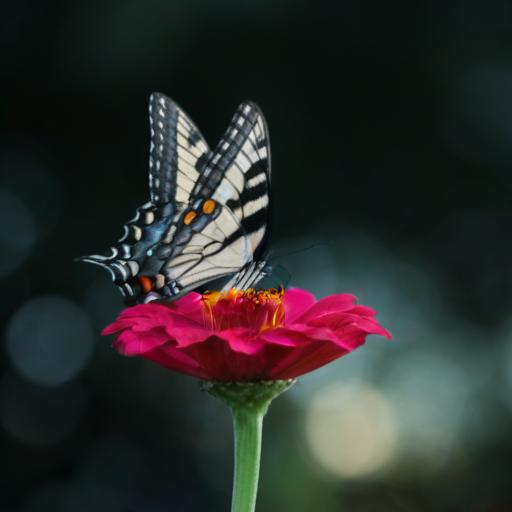} &
        \includegraphics[width=0.1\textwidth,height=0.1\textwidth]{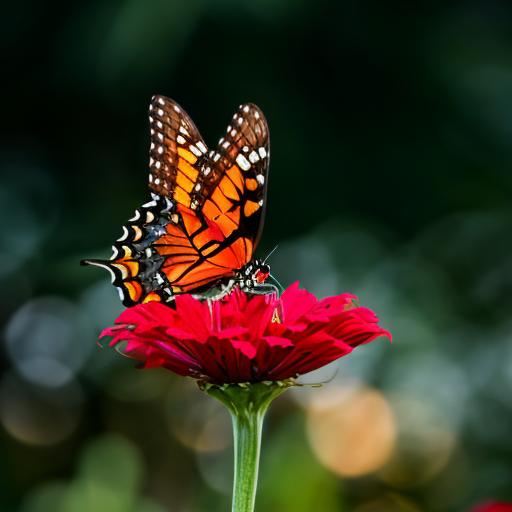} &
        \includegraphics[width=0.1\textwidth,height=0.1\textwidth]{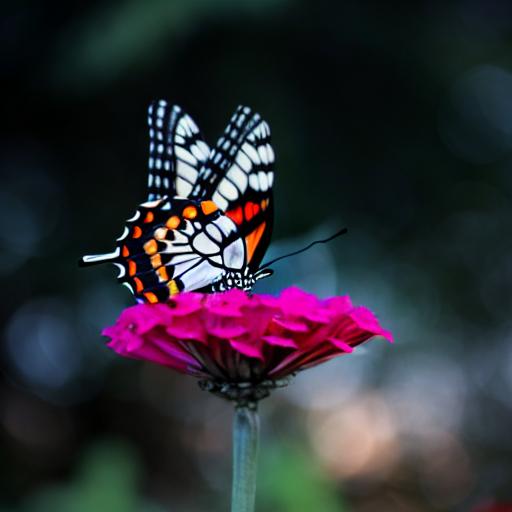} &
        \includegraphics[width=0.1\textwidth,height=0.1\textwidth]{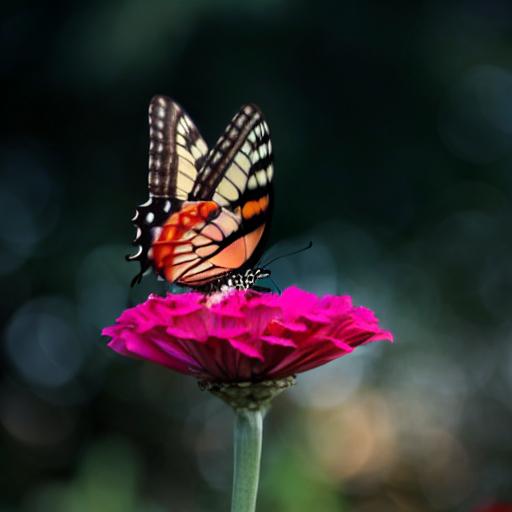} &
        \includegraphics[width=0.1\textwidth,height=0.1\textwidth]{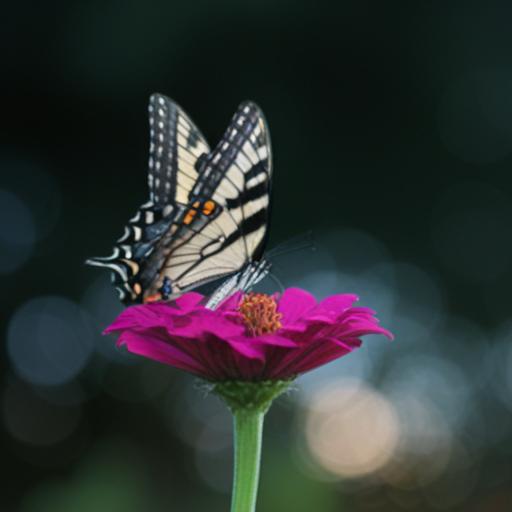} &
        \includegraphics[width=0.1\textwidth,height=0.1\textwidth]{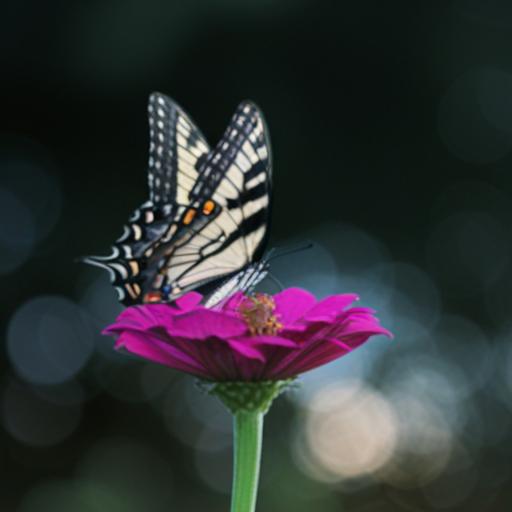} &
        \includegraphics[width=0.1\textwidth,height=0.1\textwidth]{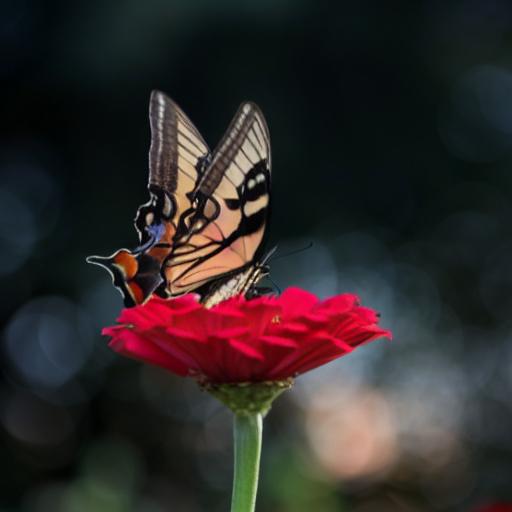} &
        \raisebox{0.04\textwidth}{\rotatebox[origin=t]{-90}{\scalebox{0.8}{\begin{tabular}{c@{}c@{}c@{}} ``Red Flower" \end{tabular}}}} \\ 

        \includegraphics[width=0.1\textwidth,height=0.1\textwidth]{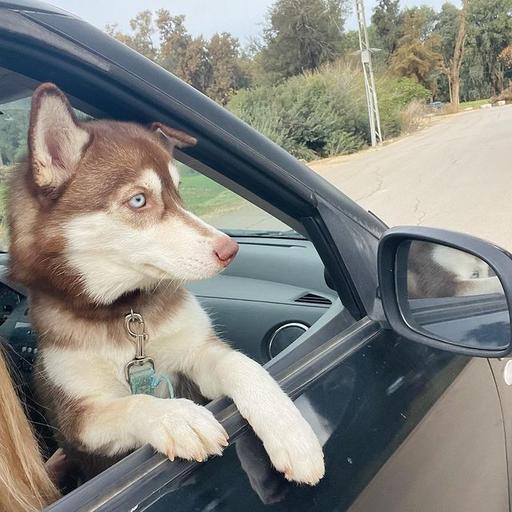} &
        \includegraphics[width=0.1\textwidth,height=0.1\textwidth]{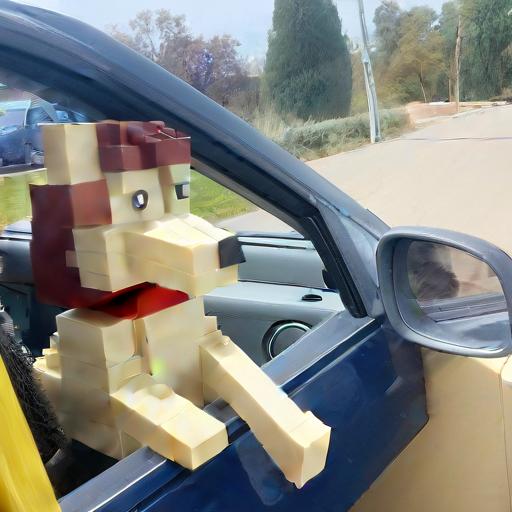} &
        \includegraphics[width=0.1\textwidth,height=0.1\textwidth]{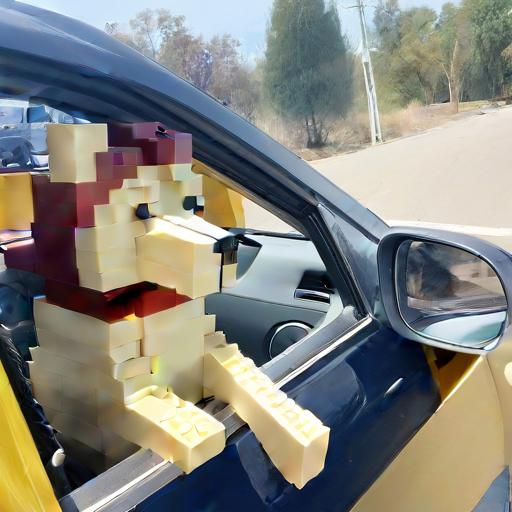} &
        \includegraphics[width=0.1\textwidth,height=0.1\textwidth]{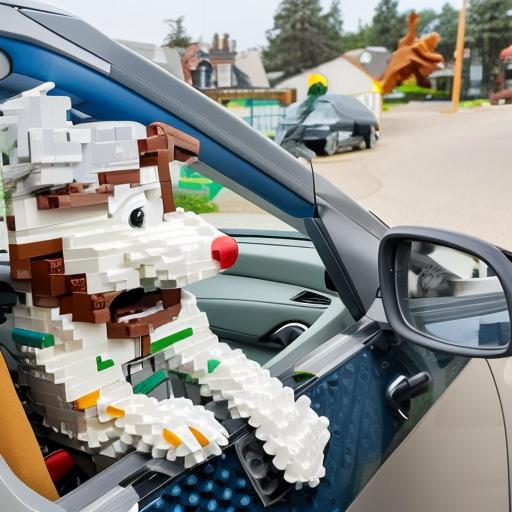} &
        \includegraphics[width=0.1\textwidth,height=0.1\textwidth]{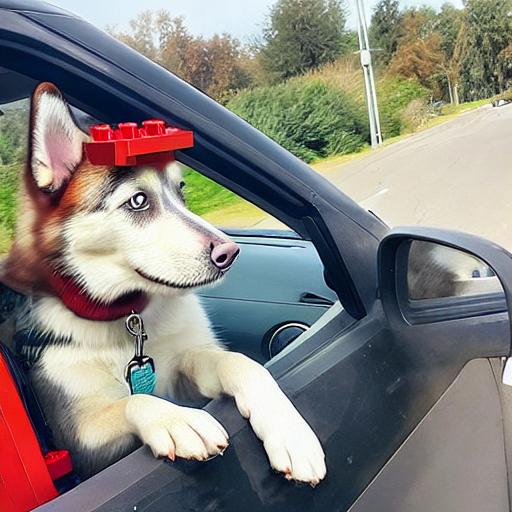} &
        \includegraphics[width=0.1\textwidth,height=0.1\textwidth]{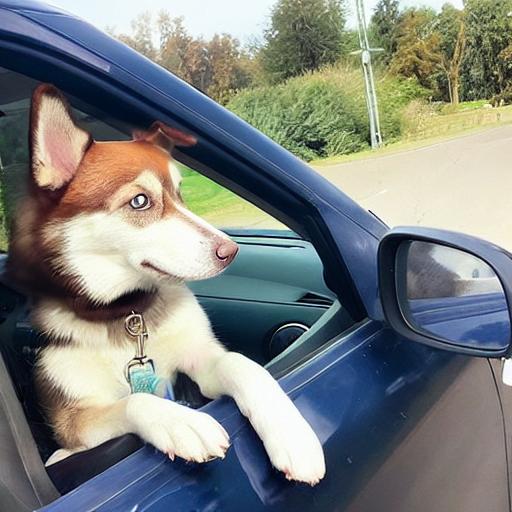} &
        \includegraphics[width=0.1\textwidth,height=0.1\textwidth]{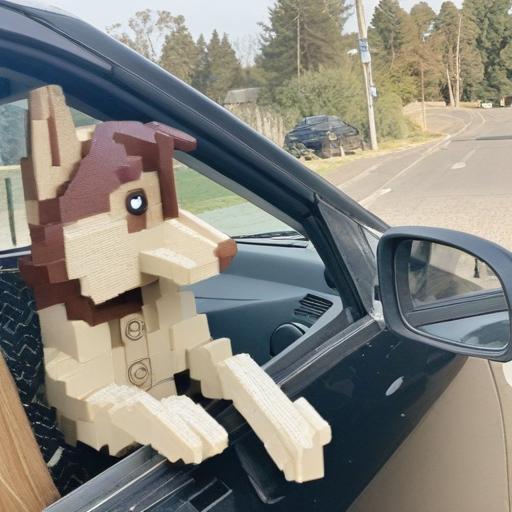} &
        \includegraphics[width=0.1\textwidth,height=0.1\textwidth]{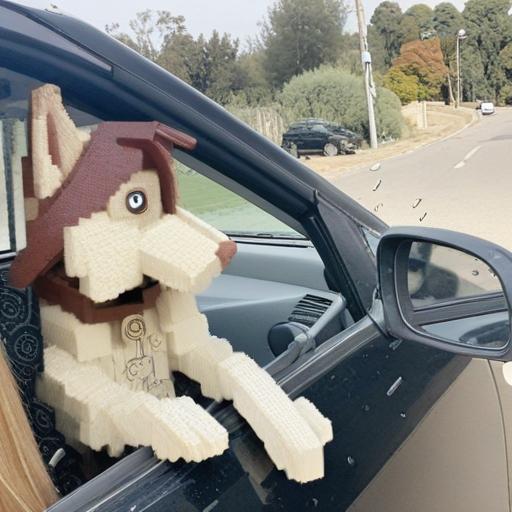} &
        \includegraphics[width=0.1\textwidth,height=0.1\textwidth]{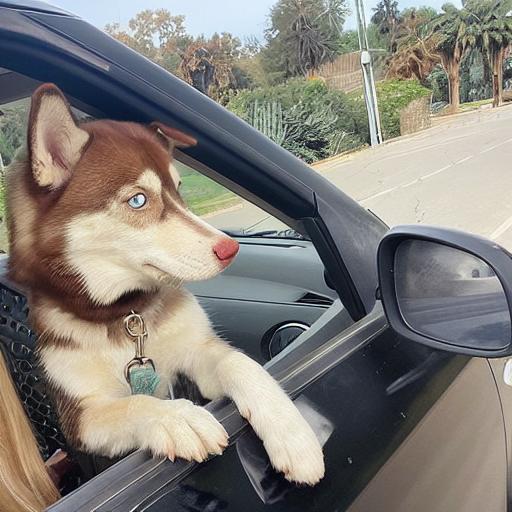} &
        \raisebox{0.03\textwidth}{\rotatebox[origin=t]{-90}{\scalebox{0.8}{\begin{tabular}{c@{}c@{}c@{}} ``Dog made \\ of lego" \end{tabular}}}} \\ 

        \includegraphics[width=0.1\textwidth,height=0.1\textwidth]{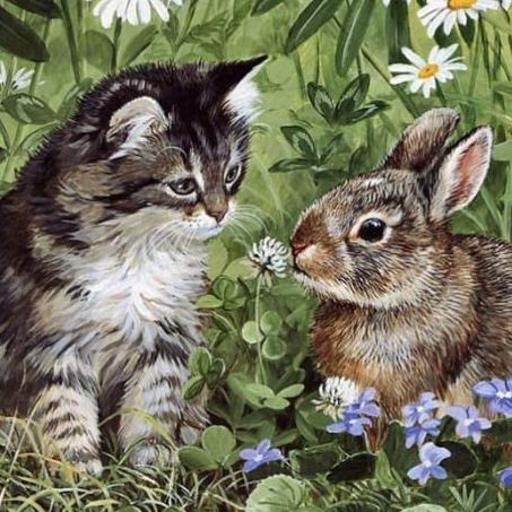} &
        \includegraphics[width=0.1\textwidth,height=0.1\textwidth]{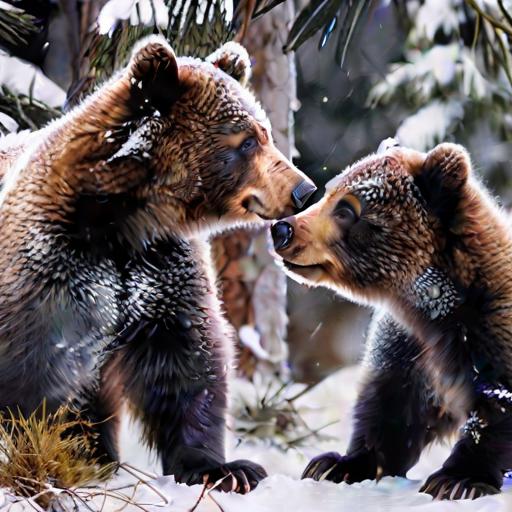} &
        \includegraphics[width=0.1\textwidth,height=0.1\textwidth]{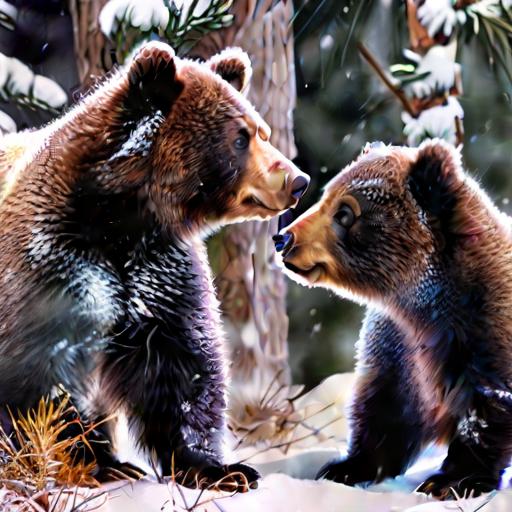} &
        \includegraphics[width=0.1\textwidth,height=0.1\textwidth]{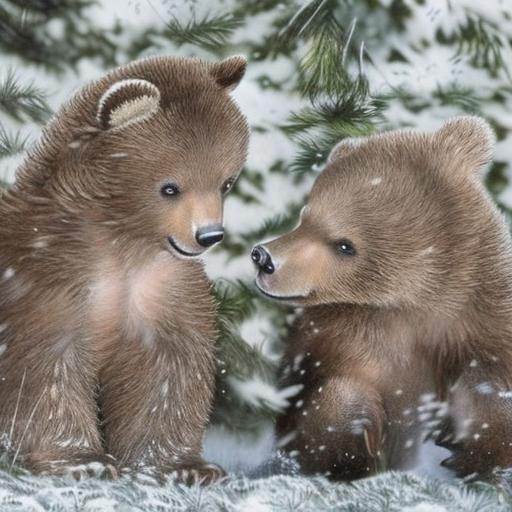} &
        \includegraphics[width=0.1\textwidth,height=0.1\textwidth]{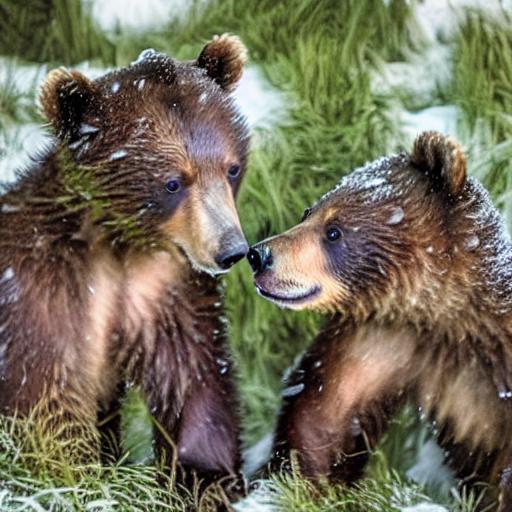} &
        \includegraphics[width=0.1\textwidth,height=0.1\textwidth]{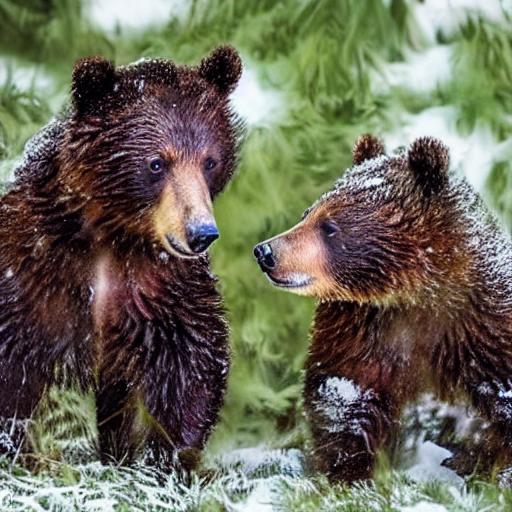} &
        \includegraphics[width=0.1\textwidth,height=0.1\textwidth]{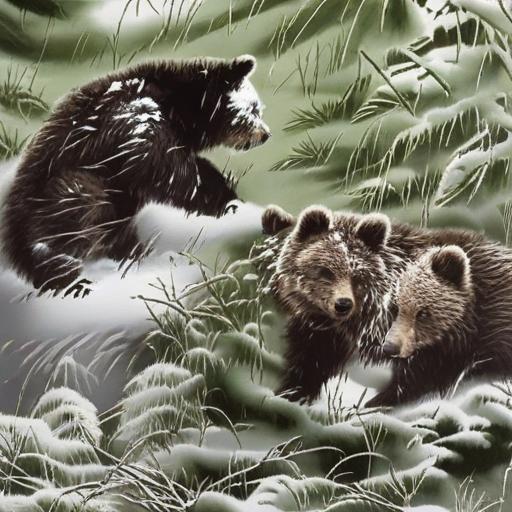} &
        \includegraphics[width=0.1\textwidth,height=0.1\textwidth]{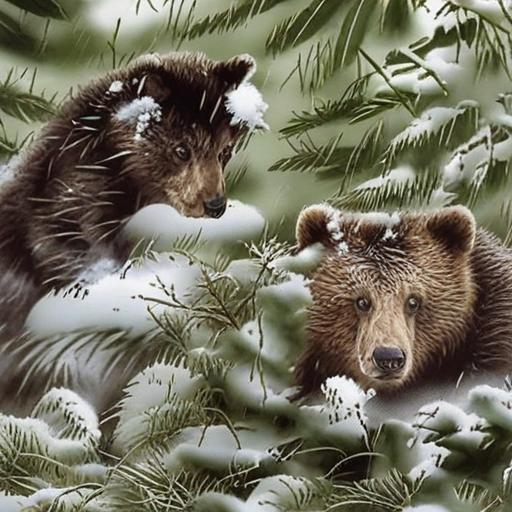} &
        \includegraphics[width=0.1\textwidth,height=0.1\textwidth]{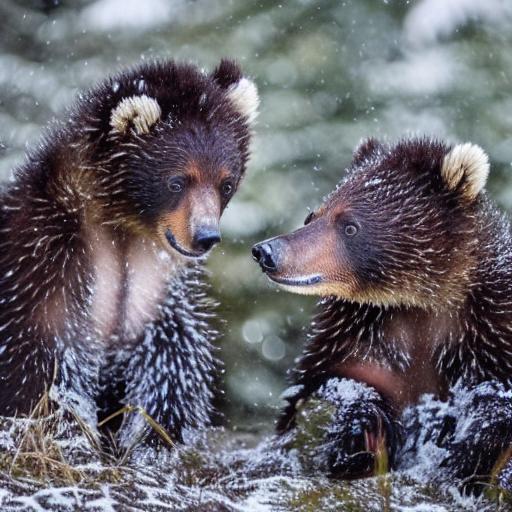} &
        \raisebox{0.03\textwidth}{\rotatebox[origin=t]{-90}{\scalebox{0.8}{\begin{tabular}{c@{}} ``Bear cubs \\ in the snow" \end{tabular}}}} \\ 

        \includegraphics[width=0.1\textwidth,height=0.1\textwidth]{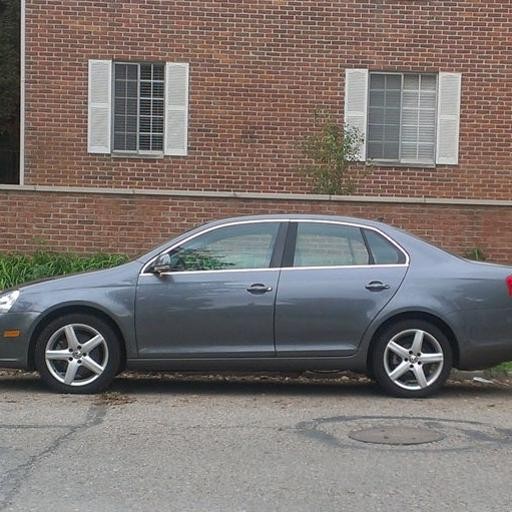} &
        \includegraphics[width=0.1\textwidth,height=0.1\textwidth]{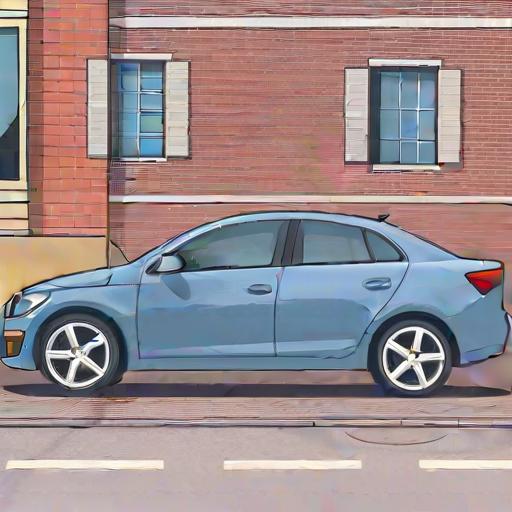} &
        \includegraphics[width=0.1\textwidth,height=0.1\textwidth]{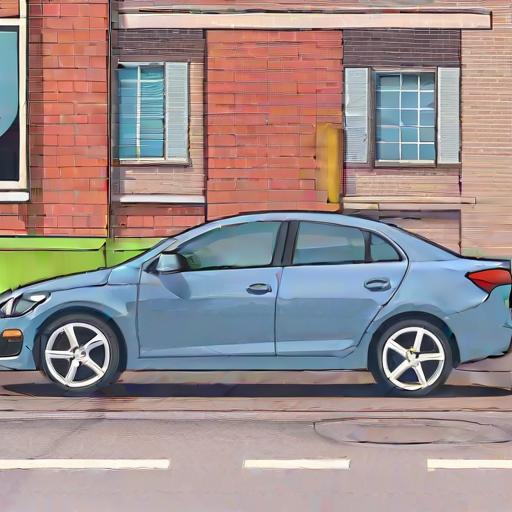} &
        \includegraphics[width=0.1\textwidth,height=0.1\textwidth]{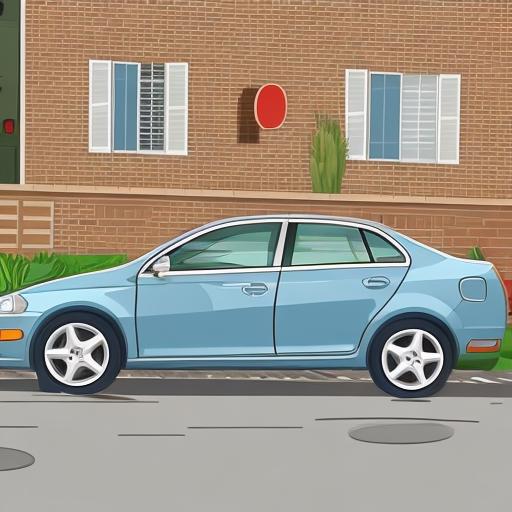} &
        \includegraphics[width=0.1\textwidth,height=0.1\textwidth]{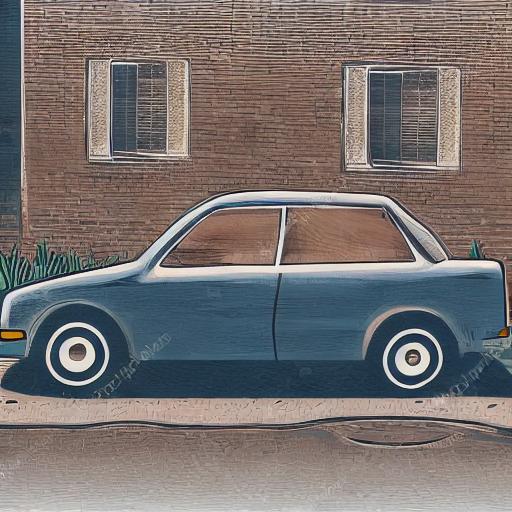} &
        \includegraphics[width=0.1\textwidth,height=0.1\textwidth]{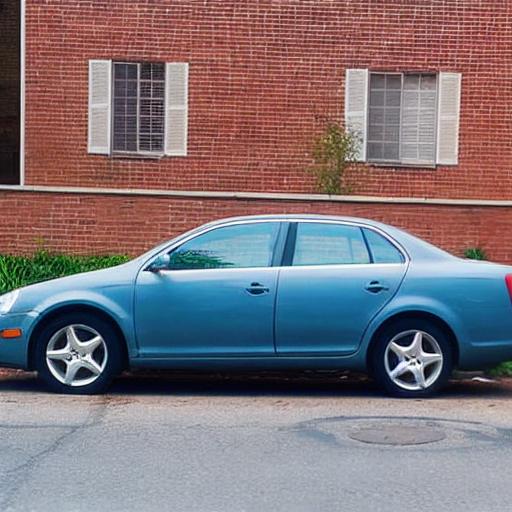} &
        \includegraphics[width=0.1\textwidth,height=0.1\textwidth]{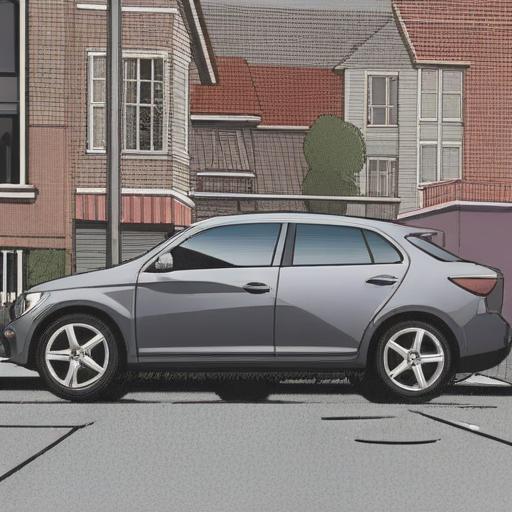} &
        \includegraphics[width=0.1\textwidth,height=0.1\textwidth]{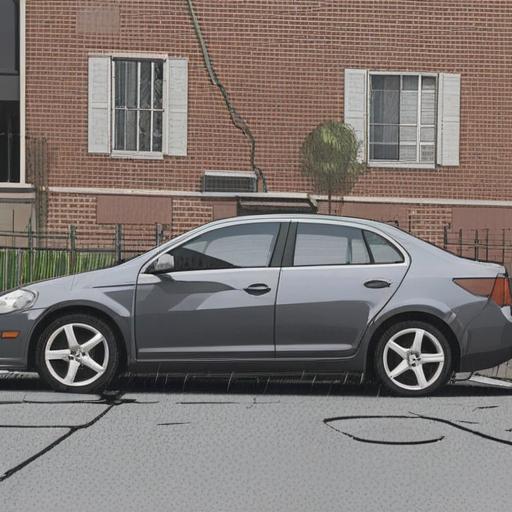} &
        \includegraphics[width=0.1\textwidth,height=0.1\textwidth]{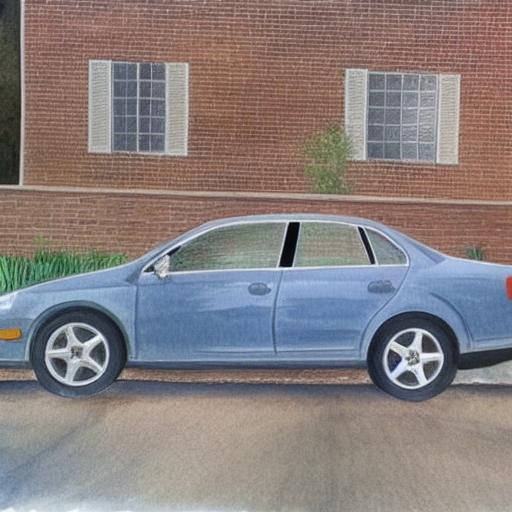} &
        \raisebox{0.04\textwidth}{\rotatebox[origin=t]{-90}{\scalebox{0.8}{\begin{tabular}{c@{}c@{}c@{}} ``Cartoon" \end{tabular}}}} \\
        
        \includegraphics[width=0.1\textwidth,height=0.1\textwidth]{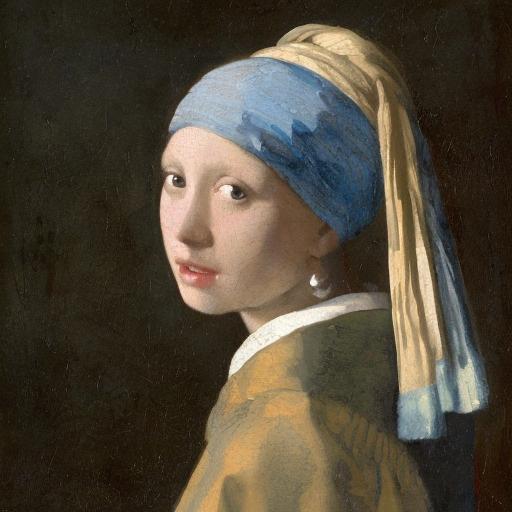} &
        \includegraphics[width=0.1\textwidth,height=0.1\textwidth]{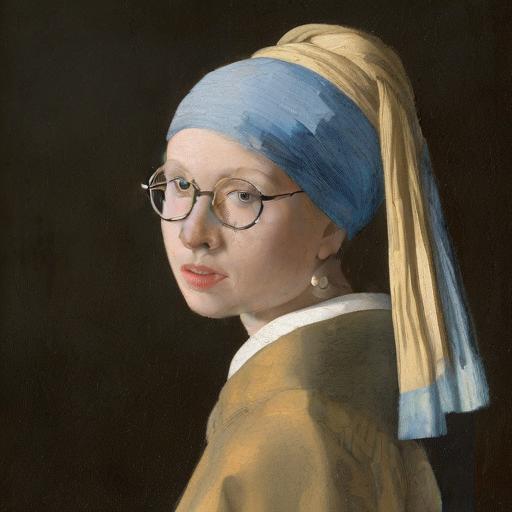} &
        \includegraphics[width=0.1\textwidth,height=0.1\textwidth]{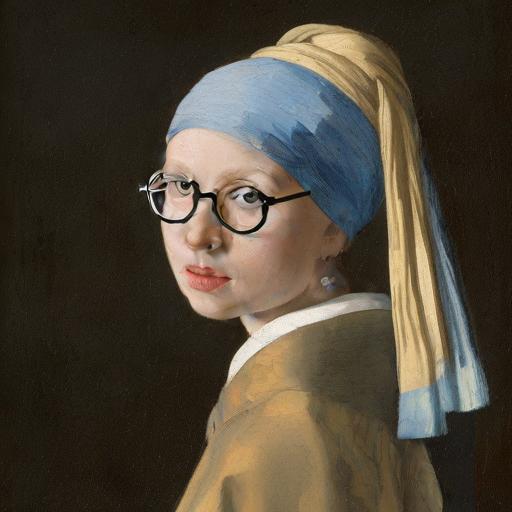} &
        \includegraphics[width=0.1\textwidth,height=0.1\textwidth]{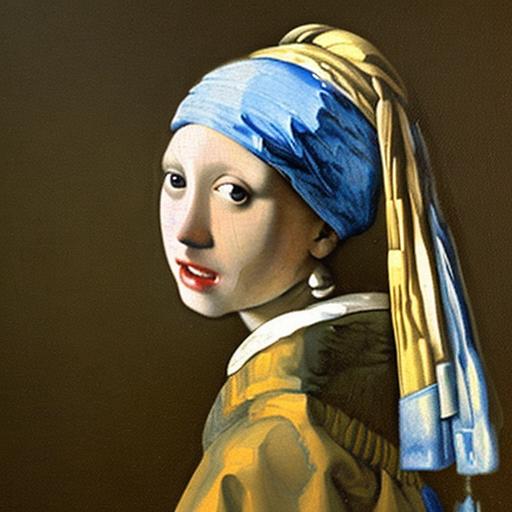} &
        \includegraphics[width=0.1\textwidth,height=0.1\textwidth]{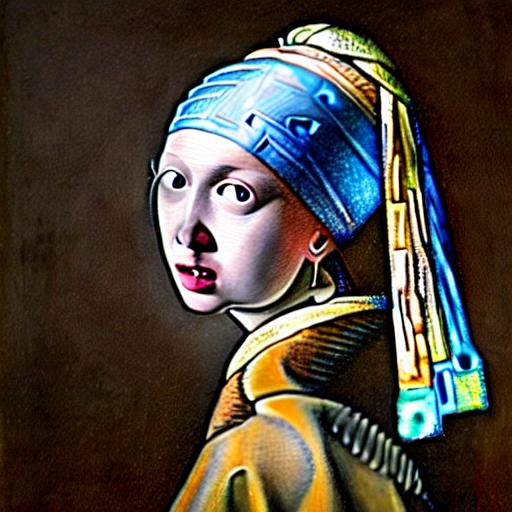} &
        \includegraphics[width=0.1\textwidth,height=0.1\textwidth]{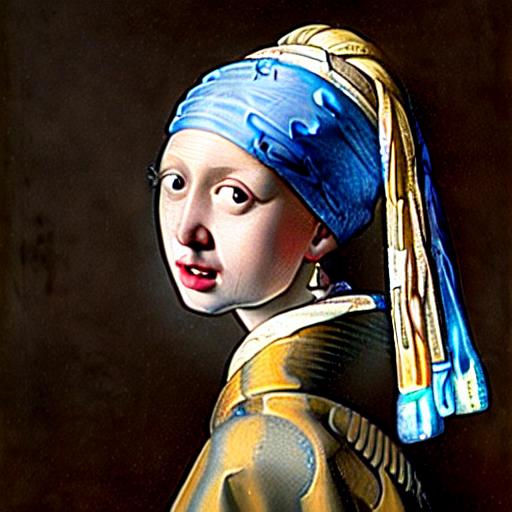} &
        \includegraphics[width=0.1\textwidth,height=0.1\textwidth]{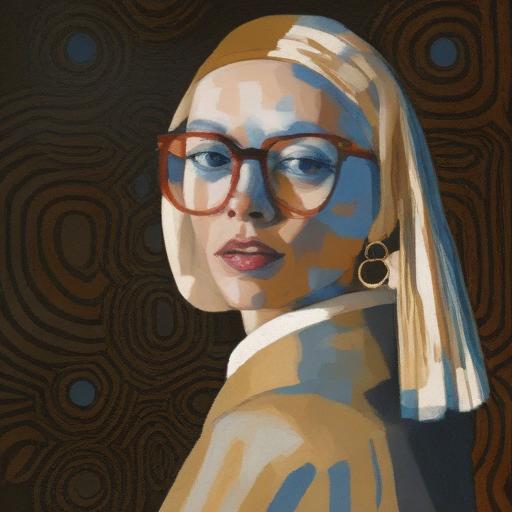} &
        \includegraphics[width=0.1\textwidth,height=0.1\textwidth]{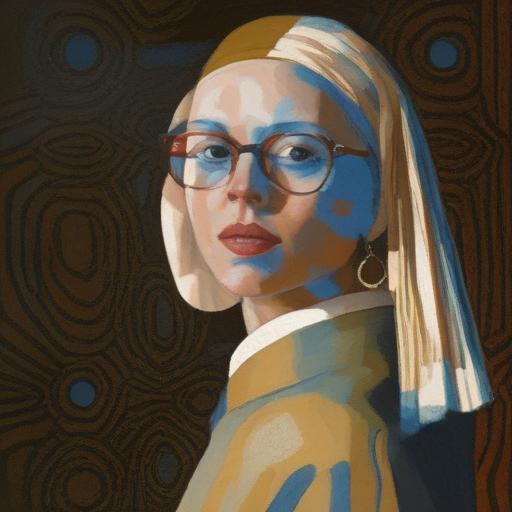} &
        \includegraphics[width=0.1\textwidth,height=0.1\textwidth]{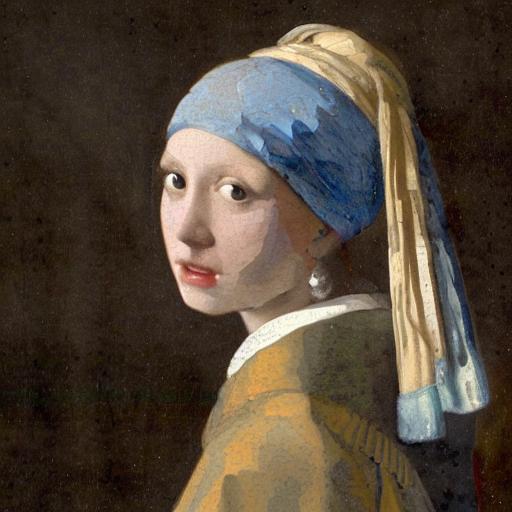} &
        \raisebox{0.04\textwidth}{\rotatebox[origin=t]{-90}{\scalebox{0.8}{\begin{tabular}{c@{}c@{}c@{}} ``With glasses" \end{tabular}}}} \\
    \end{tabular}
    }
    \caption{Comparisons against multi-step editing methods. Our results are on-par with existing baselines, while being $x5$-$x300$ faster.}
    \label{fig:ours_vs_many_steps}
\end{figure*}

Next, we compare our method against a series of baselines, including both many-step approaches as well as few-step sampling alternatives. In \cref{fig:ours_vs_many_steps} we compare our method to multi-step methods. We consider a large range of approaches, including both optimization based methods (Null-text, \cite{mokady2022null}), feature-preserving ones (PnP, \cite{tumanyan2022plug}), the baseline edit-friendly editing approach and the recently introduced ReNoise \cite{garibi2024renoise} which performs multi-step inversion ($\sim 40$ steps) but re-generates the image via a fast sampling method ($\sim 4$ steps). Our method achieves comparable or better results than most multi-step baselines, while being significantly faster.

\begin{figure*}
    \centering
    \setlength{\tabcolsep}{1.5pt}
    {\normalsize
    \begin{tabular}{c c c c c c c}
    
        {\small Input} & {\small Ours (4 steps)} & {\small Ours (3 steps)} & {\small SDEdit (0.5)} & {\small SDEdit (0.75)} & {\begin{tabular}{c} \small Edit-Friendly \\ \small (4 steps) \end{tabular}} \\

        \includegraphics[width=0.123\textwidth,height=0.123\textwidth]{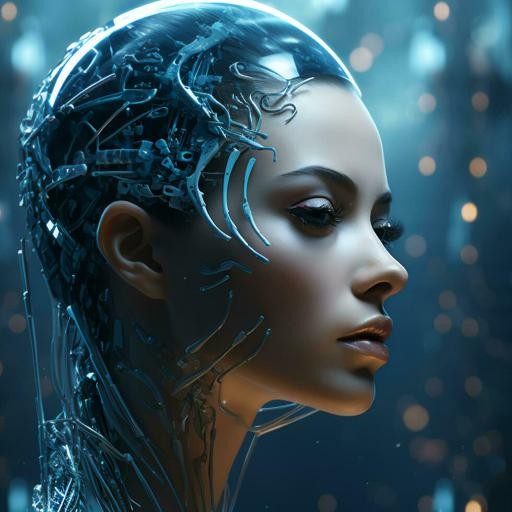} &
        \includegraphics[width=0.123\textwidth,height=0.123\textwidth]{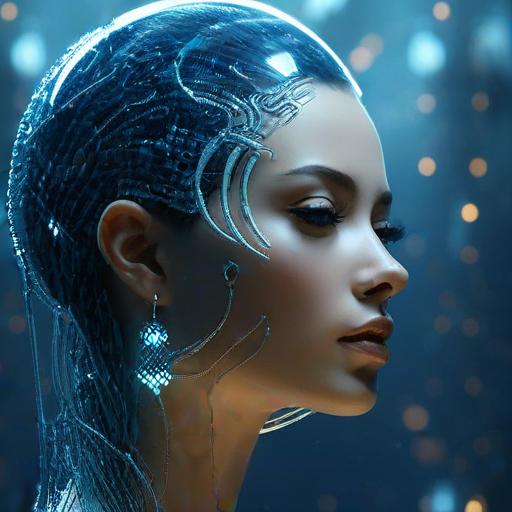} &
        \includegraphics[width=0.123\textwidth,height=0.123\textwidth]{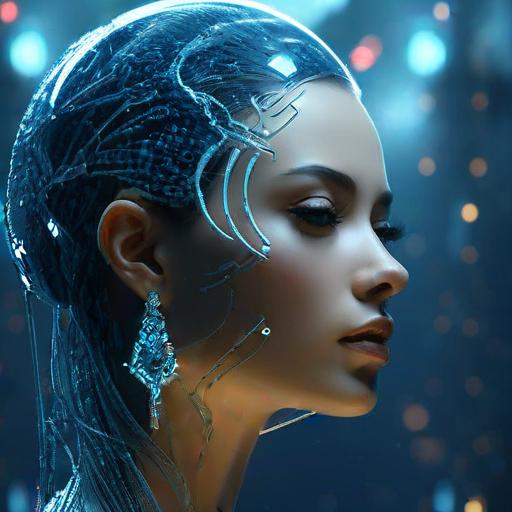} &
        \includegraphics[width=0.123\textwidth,height=0.123\textwidth]{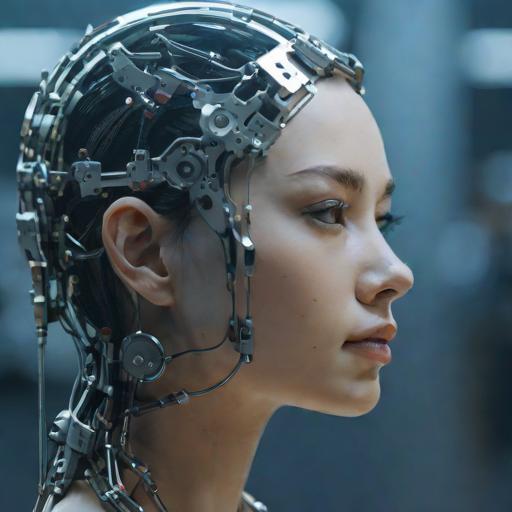} &
        \includegraphics[width=0.123\textwidth,height=0.123\textwidth]{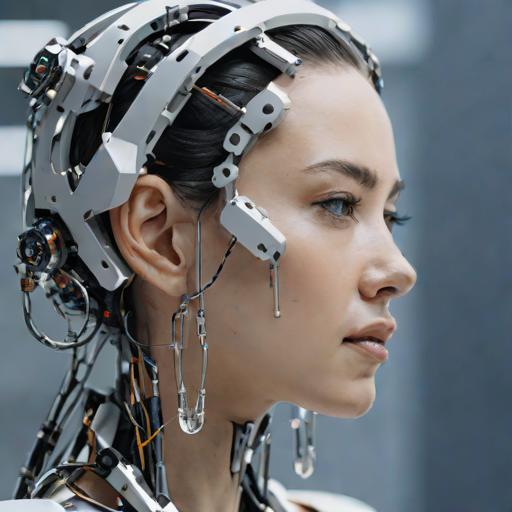} &
        \includegraphics[width=0.123\textwidth,height=0.123\textwidth]{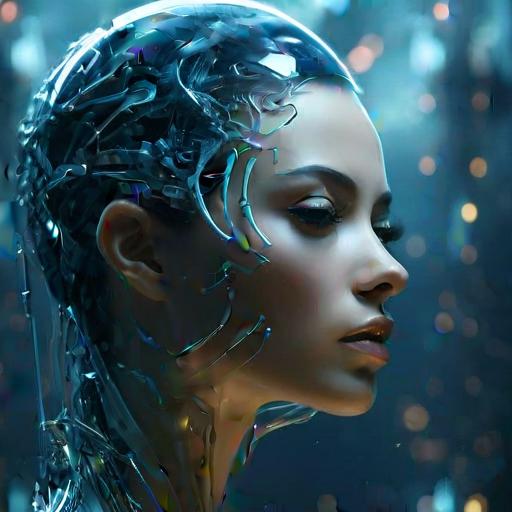} &
        \raisebox{0.05\textwidth}{\rotatebox[origin=t]{-90}{\scalebox{0.8}{\begin{tabular}{c@{}c@{}c@{}} ``With earrings" \end{tabular}}}} \\ 

        \includegraphics[width=0.123\textwidth,height=0.123\textwidth]{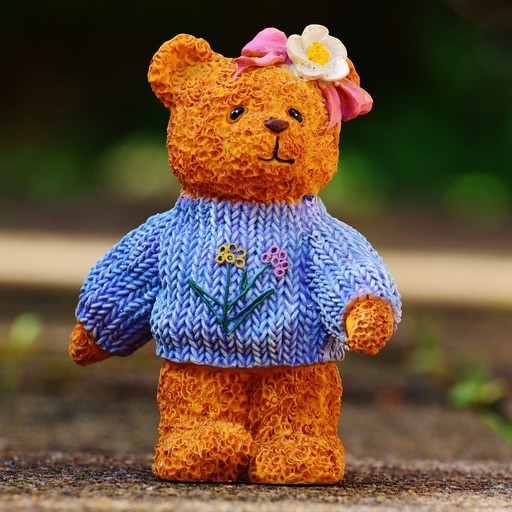} &
        \includegraphics[width=0.123\textwidth,height=0.123\textwidth]{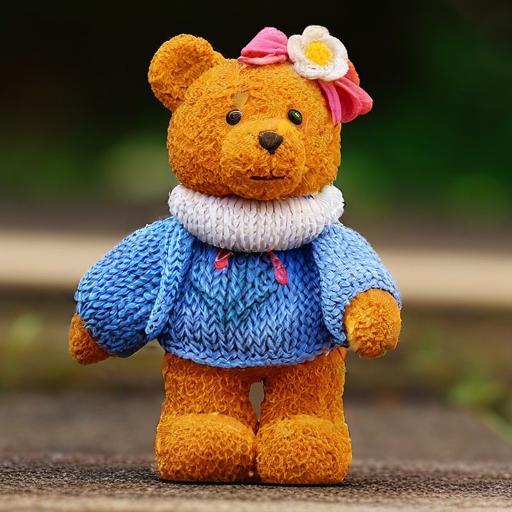} &
        \includegraphics[width=0.123\textwidth,height=0.123\textwidth]{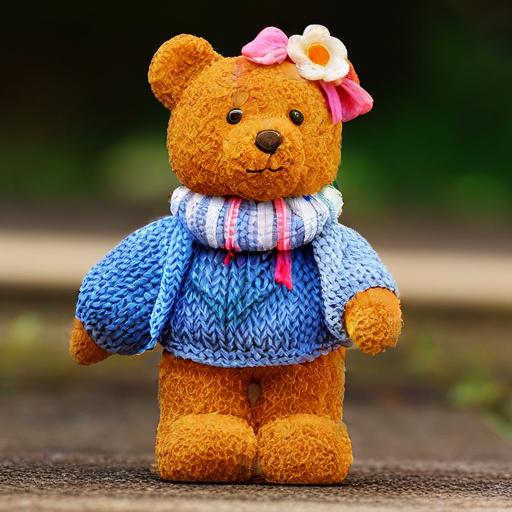} &
        \includegraphics[width=0.123\textwidth,height=0.123\textwidth]{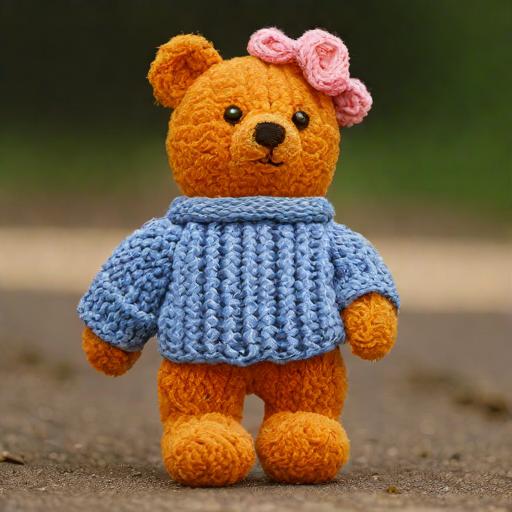} &
        \includegraphics[width=0.123\textwidth,height=0.123\textwidth]{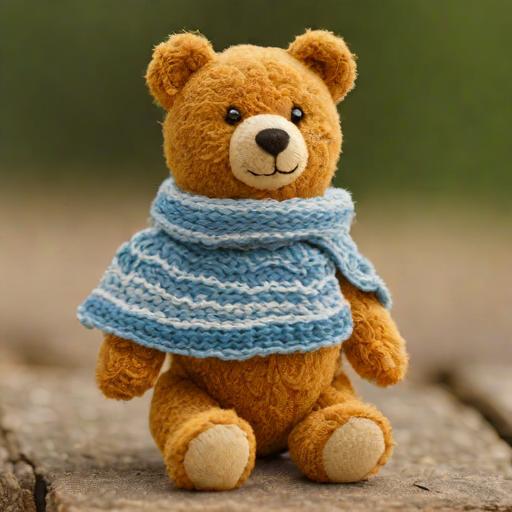} &
        \includegraphics[width=0.123\textwidth,height=0.123\textwidth]{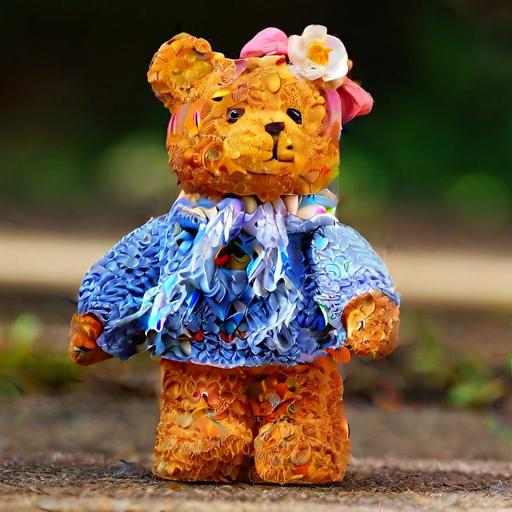} &
        \raisebox{0.05\textwidth}{\rotatebox[origin=t]{-90}{\scalebox{0.8}{\begin{tabular}{c@{}c@{}c@{}} ``Wearing a scarf" \end{tabular}}}} \\

        \includegraphics[width=0.123\textwidth,height=0.123\textwidth]{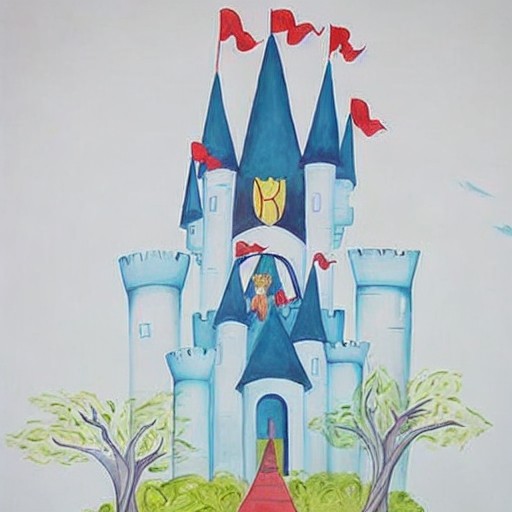} &
        \includegraphics[width=0.123\textwidth,height=0.123\textwidth]{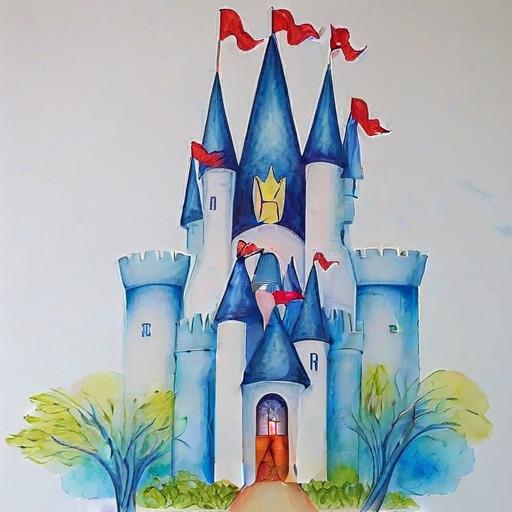} &
        \includegraphics[width=0.123\textwidth,height=0.123\textwidth]{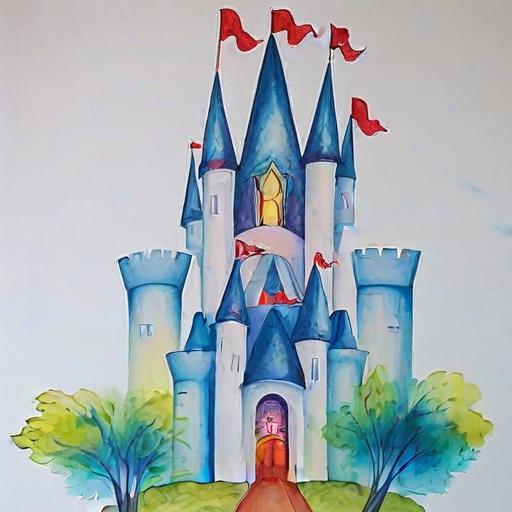} &
        \includegraphics[width=0.123\textwidth,height=0.123\textwidth]{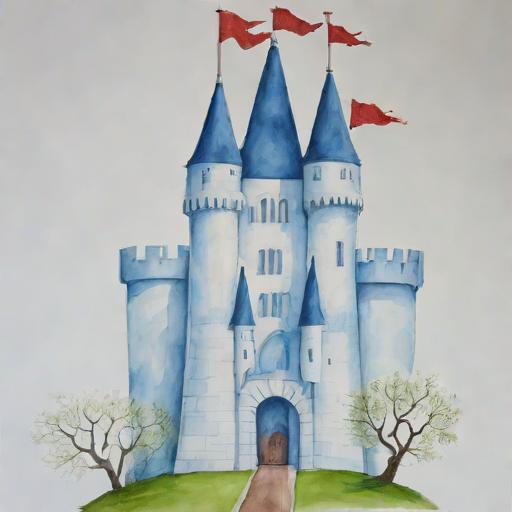} &
        \includegraphics[width=0.123\textwidth,height=0.123\textwidth]{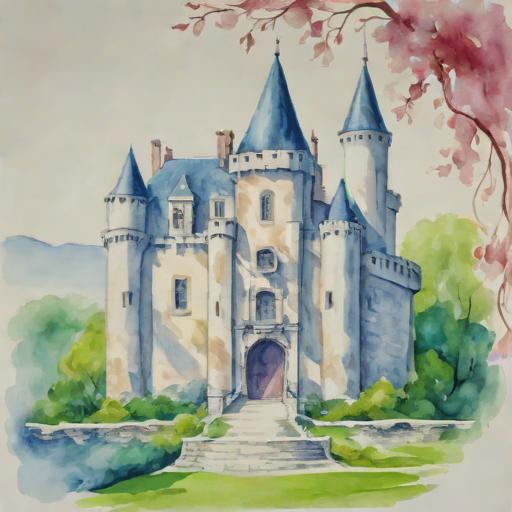} &
        \includegraphics[width=0.123\textwidth,height=0.123\textwidth]{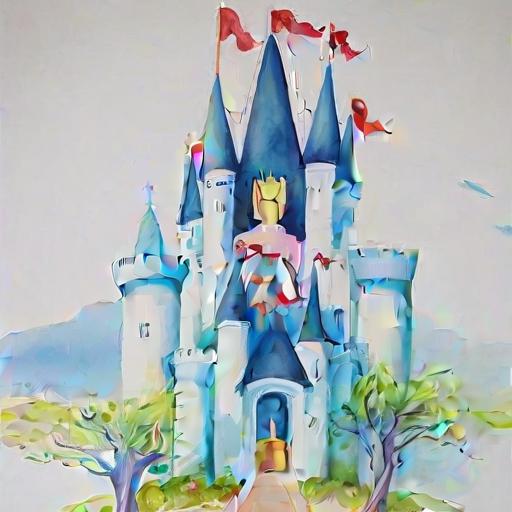} &
        \raisebox{0.05\textwidth}{\rotatebox[origin=t]{-90}{\scalebox{0.8}{\begin{tabular}{c@{}c@{}c@{}} ``Watercolor" \end{tabular}}}} \\

    \end{tabular}
    }
    \caption{Comparisons with few-step methods. Our method can better preserve the content of the original image, while applying meaningful edits.}\label{fig:ours_vs_few_steps}
\end{figure*}

In \cref{fig:ours_vs_few_steps} we compare our method to SDEdit~\cite{meng2022sdedit} and vanilla applications of edit-friendly inversion with SDXL-Turbo using $4$ steps. Our method can better maintain the details of the original image, while also achieving better prompt alignment. Notably, the edit-friendly approach creates significant artifacts. Please zoom in to better see the results.

For a quantitative evaluation, we employ $4$ commonly used metrics. The first two are the CLIP-space~\cite{radford2021learning} similarity between the input image and the editing result (CLIP-I), and the CLIP-space similarity between the target text and the editing result (CLIP-T). We further evaluate the CLIP-directional similarity~\cite{gal2021stylegan,brooks2022instructpix2pix}, \ie the cosine-similarity between the CLIP-space direction that connects the pre- and post-editing image, and the direction that points from a description of the original image to a description of the target edit (CLIP-Dir). Intuitively, this metric measures how well the image-space change aligns with the prompt difference, without changing content that is unrelated to the prompt. Finally, we also compare the LPIPS~\cite{zhang2018perceptual} score between the pre- and post-edited images.

The results are shown in \cref{tab:quant}. Notably, our method achieves favorable results on both prompt-alignment metrics, and indeed outperforms all other methods which do not also suffer from significant deviation from the original image. On the image preservation front, our approach does not outperform the state-of-the-art, but it achieves comparable results to many multi-step methods, despite being quicker by a factor of $\times5-\times500$.

\begin{figure}
    \centering
    \includegraphics[width=0.97\linewidth]{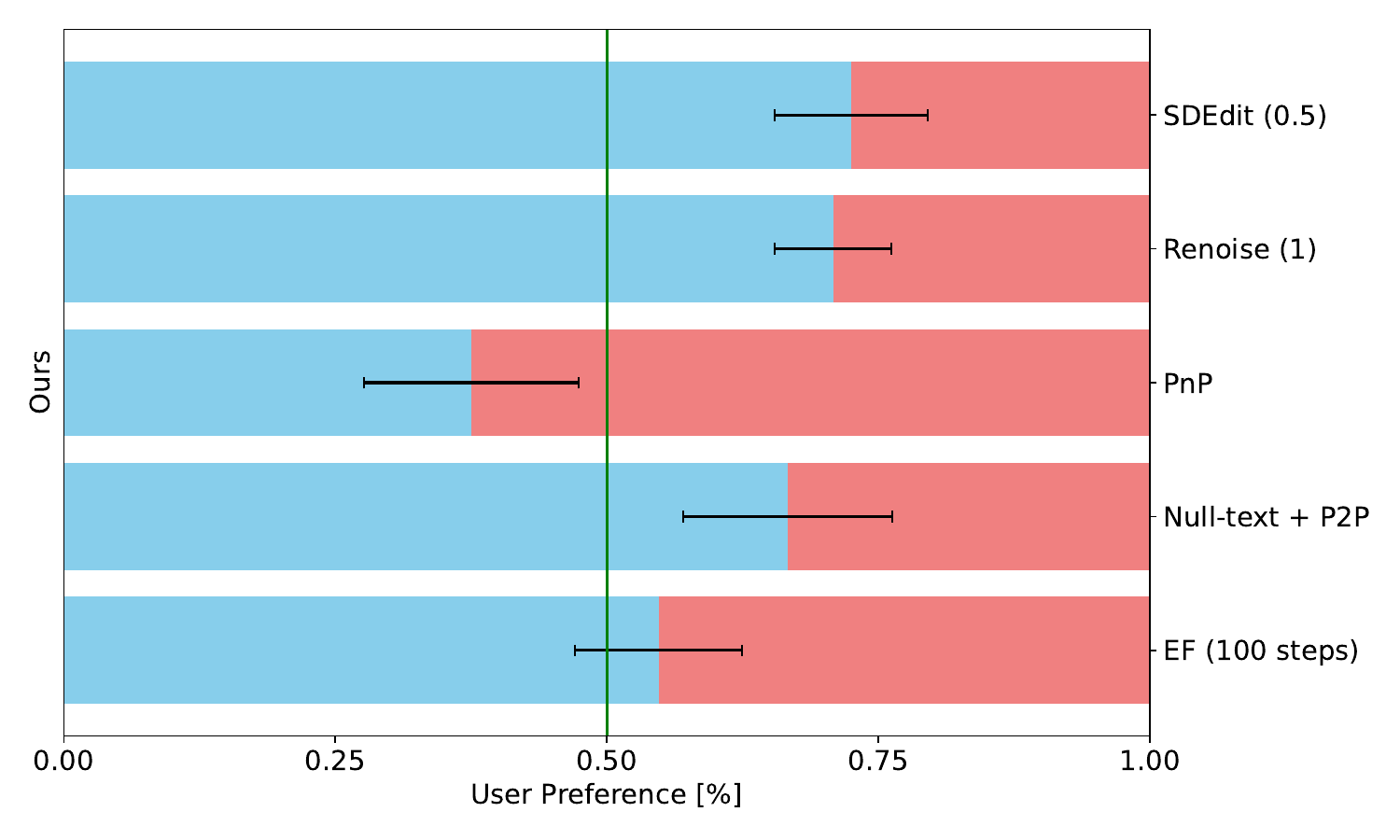}\vspace{-3pt}
    \caption{User study results. We show the \% of users who preferred each method when compared to ours. Error bars are the 68\% confidence interval.}
    \label{fig:user_study}\vspace{-5pt}
\end{figure}

To further validate our results, we also conduct a user preference study. We used a two-alternative forced choice setup, where each user was shown the source image, the target prompt, and two editing results including our method and one baseline. Users were asked to select the image that better aligns with the prompt, while preserving the content of the original image. We focused on the 5 most performant baselines, including both multi- and few-step approaches. In total, we collected 202 responses from 25 users. Results are shown in \cref{fig:user_study}. Overall, our approach is strongly preferred over competing few-step methods, while being competitive with multi-step results. Plug \& Play achieve higher user preference, but they are slower by a factor of $\times500$. All in all, these results demonstrate that our approach can successfully edit real images in as few as $3$ diffusion steps, without sacrificing quality.

\begin{table}
    \caption{Quantitative comparisons against text-based editing baselines. \textbf{Bold} indicates the best scoring method, \underline{underline} indicates the second best, and {\color{red} red} indicates the worst. EF denotes edit-friendly DDPM-inversion. Editing times include both inversion and generation, and are computed on a machine with a single A5000.}\label{tab:quant}\vspace{-3pt}
    \renewcommand{\arraystretch}{1.2}
    \addtolength{\tabcolsep}{-0.31em}
    \nprounddigits{3}
    \npdecimalsign{.}
    \begin{tabular}{c : c c c c c}
    & CLIP-T $\uparrow$ & CLIP-I $\uparrow$ & CLIP-Dir $\uparrow$ & LPIPS $\downarrow$ & Time $\downarrow$ \\
    \toprule
    EF (100 steps) & 0.276 &	0.760 & 0.173 & \underline{0.102} & 29.4s \\ 
    \begin{tabular}{c} EF + P2P \\[-0.5ex] (100 steps) \end{tabular} & 0.266 & \textbf{0.801} & 0.159 &	\textbf{0.090} & 43.2s \\ 
    Plug \& Play & 0.281 &	0.746 & \underline{0.197} & 0.116 & 202s \\ 
    Null-text + P2P & {\color{red} 0.263}	& 0.779	& 0.194	& \textbf{0.091} & 150s \\ 
    \midrule
    ReNoise (0.75) & 0.271 & \underline{0.759} & 0.182 & 0.110 & 1.75s \\ 
    ReNoise (1.0) & 0.275 & 0.727 & 0.187 & 0.123  & 2.30s \\
    \midrule
    Ours (4 steps) & \underline{0.291} & 0.745 & \textbf{0.216} & 0.118 & 0.412s \\
    Ours (3 steps) & \underline{0.291} & 0.748 & 0.211 & 0.118 & 0.321s \\
    EF (4 steps) & 0.269 & 0.756 & 0.120 & 0.105 & 0.576s \\
    EF (3 steps) & 0.266 & 0.737 & {\color{red} 0.114} & 0.115 & 0.429s \\
    SDEdit (0.75) & \textbf{0.301} &	{\color{red} 0.671} &	0.193 &	{\color{red} 0.191} & 0.065s \\ 
    SDEdit (0.5) & 0.281 & 0.751 &	0.163 &	0.149 & 0.098s \\
    \end{tabular}\vspace{-5pt}
\end{table}

\subsection{Ablation study}

\begin{figure}[!hb]
    \centering
    \setlength{\tabcolsep}{2pt}
    
    \begin{tabular}{c c c c}

        \small Input & 
        \begin{tabular}{@{}c@{}} \small Ours w/o \\[-0.5ex] \small guidance \end{tabular} & \begin{tabular}{@{}c@{}} \small Ours w/ \\[-0.5ex] \small last step shift \end{tabular} & \begin{tabular}{@{}c@{}} \small Ours w/ \\[-0.5ex] \small STD re-norm \end{tabular} \\

        \includegraphics[width=0.11\textwidth,height=0.11\textwidth]{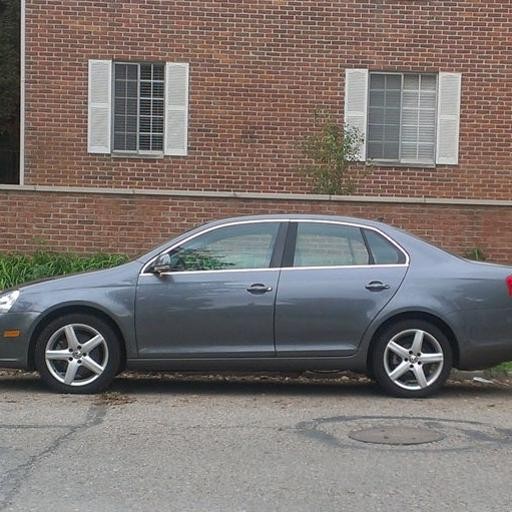} &
        \includegraphics[width=0.11\textwidth,height=0.11\textwidth]{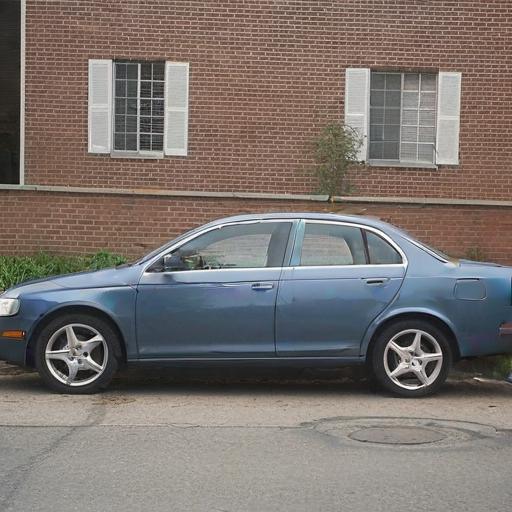} &
        \includegraphics[width=0.11\textwidth,height=0.11\textwidth]{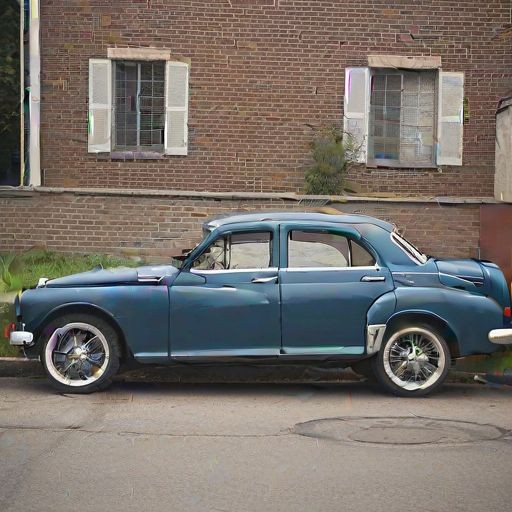} & \includegraphics[width=0.11\textwidth,height=0.11\textwidth]{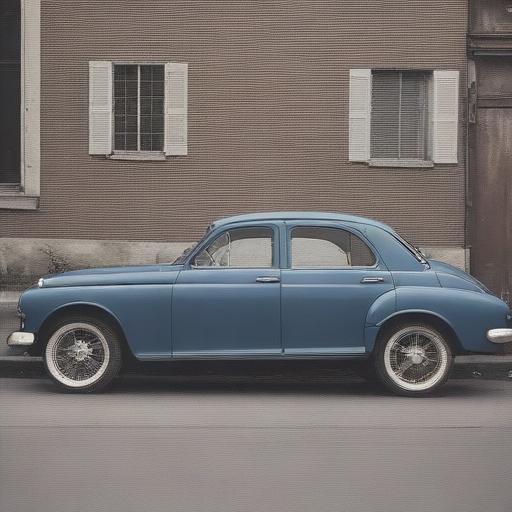} \\[2pt]

        \begin{tabular}{@{}c@{}} \small Ours w/o \\[-0.5ex] \small timestep shift \end{tabular} &
        \begin{tabular}{@{}c@{}} \small Ours w/o \\[-0.5ex] \small norm clipping \end{tabular} &

        \small Ours & \small Edit-friendly \\
        \includegraphics[width=0.11\textwidth,height=0.11\textwidth]{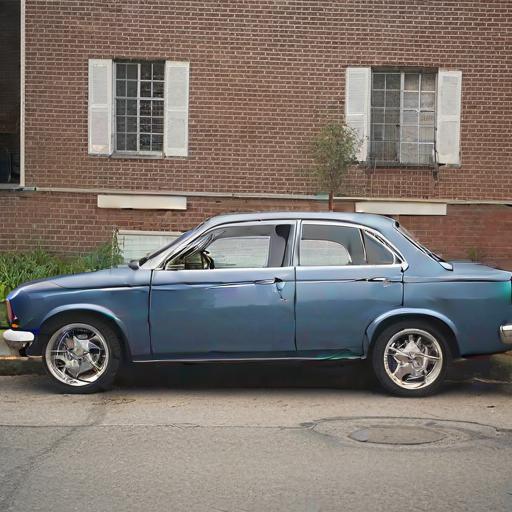} & 
        \includegraphics[width=0.11\textwidth,height=0.11\textwidth]{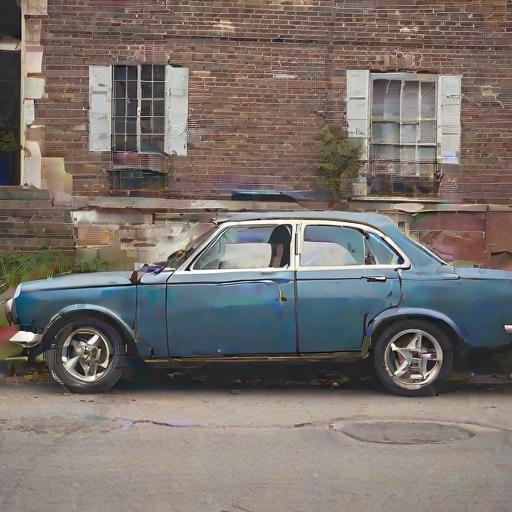} &
        \includegraphics[width=0.11\textwidth,height=0.11\textwidth]{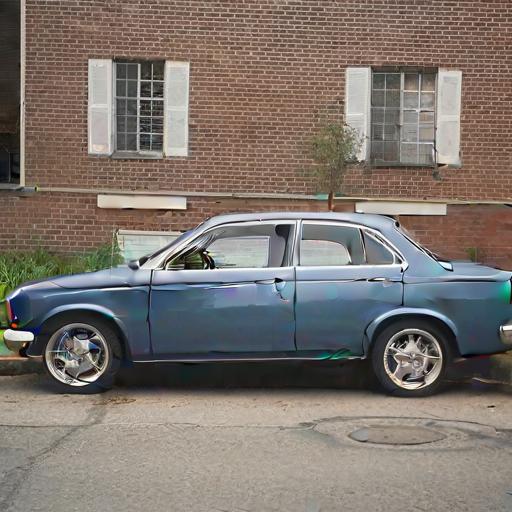} &
        \includegraphics[width=0.11\textwidth,height=0.11\textwidth]{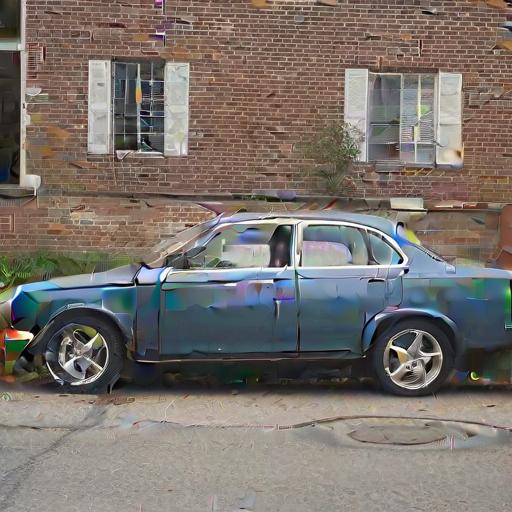} \\

        \multicolumn{3}{c}{\small ``Old car''} \\
        
    \end{tabular}
    \caption{Qualitative ablation results. 
    Removing the pseudo-guidance leads to diminished editing results. Removing the timestep shift or the norm clipping leads to increased visual artifacts, and alternative noise normalization schemes may lead to overly smoothed results. Removing both of our components and returning to vanilla ``edit-friendly'' leads to severe artifacts. Please zoom in to better view the artifacts.
}\label{fig:ablation_study}
\end{figure}

Next, we turn to an ablation study where we analyze the effect of the different components inherent in our approach. Specifically, we consider: (1) the effect of disabling the time-shift applied to the network predictions, (2) the effect of removing the noise-clipping at the final denoising step, (3) the effects of other noise normalization schemes, such as shifting only the last step or directly re-normalizing the noise statistics (4) the effect of disabling the pseudo-guidance, and (5) the performance of the naive edit-friendly DDPM inversion in this few-step setup. Visual results are provided in \cref{fig:ablation_study} while quantitative results are found in \cref{tab:quant_ablation}.

Removing the time-shift maintains the text-alignment, but harms the image-to-image similarity because of the introduction of visual artifacts. A similar effect can be observed to a lesser extent when removing the noise clipping or when shifting only the last step. Explicit re-normalization of noise maps leads to overly-smoothed results and loss of detail. Without pseudo-guidance, we observe significantly diminished prompt-alignment. Finally, with all methods disabled, the results show significantly diminished scores on all fronts, and have considerable visual artifacts.

\begin{table}
    \caption{Quantitative ablation study. \textbf{Bold} indicates the best scoring method, \underline{underline} indicates the second best. EF denotes edit-friendly DDPM-inversion.}\label{tab:quant_ablation}\vspace{-3pt}
    \addtolength{\tabcolsep}{-0.2em}
    \renewcommand{\arraystretch}{1.1}
    \centering
    \begin{tabular}{c : c c c c}
    & CLIP-T $\uparrow$ & CLIP-I $\uparrow$ & CLIP-Dir $\uparrow$ & LPIPS $\downarrow$ \\
    \toprule
    Ours & \underline{0.291} & \underline{0.745} & \textbf{0.216} & 0.118 \\
    w/o timestep shift & \underline{0.291} & 0.736 & 0.183 & 0.124 \\
    w/o clipping & 0.289 & 0.741 &	\underline{0.197} &	0.116 \\
    Shift last step & \textbf{0.296} & 0.728 & 0.189 & 0.128 \\
    w/ STD re-norm & 0.282 & 0.701 & 0.192 & 0.155 \\
    w/o guidance & 0.274 & \textbf{0.795} & 0.181 & \textbf{0.096} \\
    EF (4 steps) & 0.269 &0.756	& 0.120 & \underline{0.105} \\
    \end{tabular}\vspace{-1pt}
\end{table}

\section{Limitations}

\begin{figure}[!hb]
    \centering
    \large
    {
    \setlength{\tabcolsep}{2pt}
    
    \begin{tabular}{c c c c}
    
    \raisebox{0.06\textwidth}{\rotatebox[origin=t]{90}{\scalebox{1.0}{Input}}} &

    \includegraphics[width=0.135\textwidth,height=0.135\textwidth]{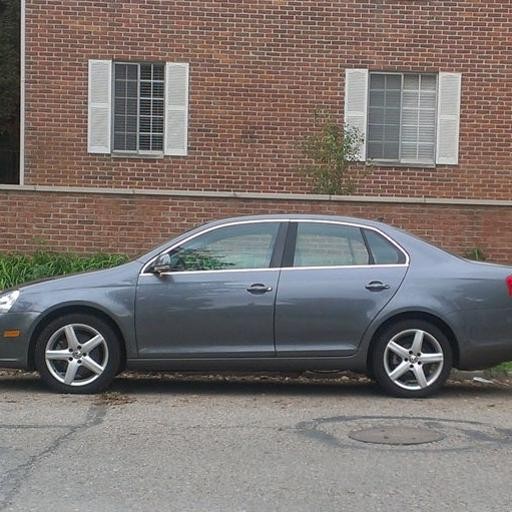} &
    \includegraphics[width=0.135\textwidth,height=0.135\textwidth]{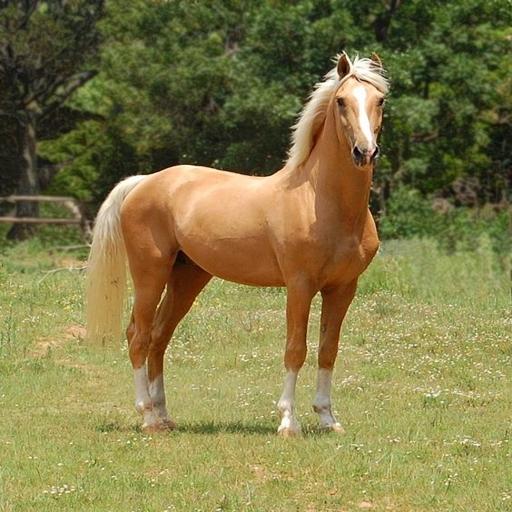} &
    \includegraphics[width=0.135\textwidth,height=0.135\textwidth]{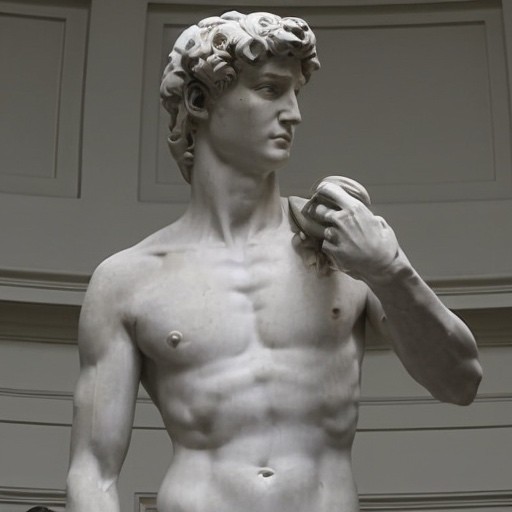} \\

    \raisebox{0.06\textwidth}{\rotatebox[origin=t]{90}{\scalebox{1.0}{Edited}}} & 
    \includegraphics[width=0.135\textwidth,height=0.135\textwidth]{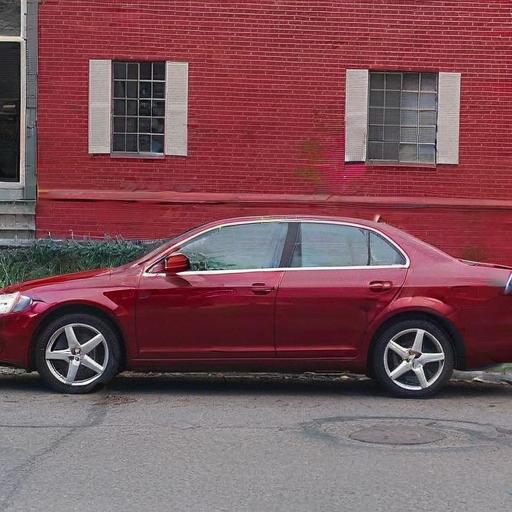} &          

    \includegraphics[width=0.135\textwidth,height=0.135\textwidth]{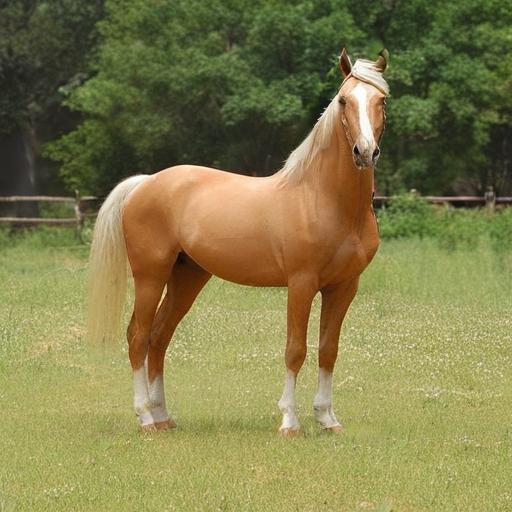} &
            
    \includegraphics[width=0.135\textwidth,height=0.135\textwidth]{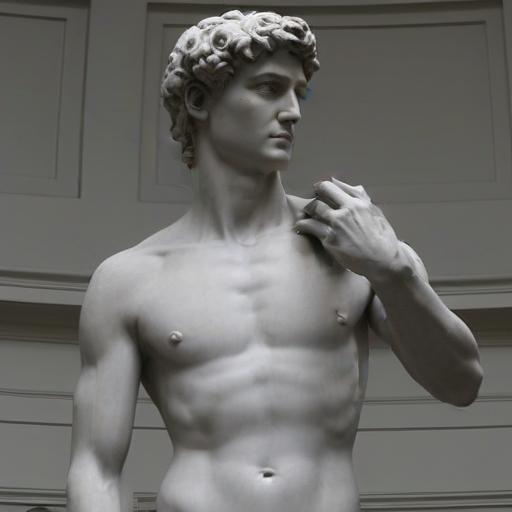} \\
    
    &
     ``Red car'' &          

     ``Wearing a hat'' &
            
     ``Crossing arms'' \\

    \end{tabular}
    }
    \caption{Method limitations. Our approach may display attribute leakage when editing an object, and may struggle to add novel objects or significantly modify poses.}\label{fig:limitations}
\end{figure}

Our method shares many of the limitations of prior text-based editing work. In particular, it can struggle with changing geometries through the prompt (\eg asking for a person to cross his arms). Similarly, it may struggle to introduce new objects into the scene (adding a hat to a horse). Another limitation can be found in prompt leakage to undesired areas of the scene (\eg turning a car red may lead to a shift in background color). In addition, it can struggle with stylization when converting real imagery to painting or vice versa. We show examples of limitations and failure cases in \cref{fig:limitations}.

\section{Conclusions}
We presented TurboEdit, a fast text-based editing method that leverages newly introduced fast-sampling methods to significantly reduce the time it takes to manipulate an image. Notably, our method achieves sub-second editing times, enabling an interactive editing experience.

Our approach is derived from a deeper analysis of the dynamics of existing noise-inversion approaches, leading to a series of technically simple fixes. Hence, we hope that our work can provide not only a useful application for creative workflows, but also shed a light on components and ideas that should be kept in mind when adapting other tools to the few-step regime.

In the future, we hope to investigate further improvements for editing flows in the few-step regime, such as a more principled alignment of noise schedules to the expected statistics, or exploring ways to improve geometric modifications and object insertion.

\subsection*{Acknowledgements}
This work was partially supported by ISF (grants 2492/20 and 3441/21). 

\begin{figure*}
    \centering
    \setlength{\tabcolsep}{1.5pt}
    {\normalsize
    \begin{tabular}{c c c c c c c}
    
        {\small Original Image} & \multicolumn{6}{c}{\small Editing results} \\

        \vspace{-4pt} \includegraphics[width=0.13\textwidth,height=0.13\textwidth]{figures/ours_only/09.jpg} &
        \includegraphics[width=0.13\textwidth,height=0.13\textwidth]{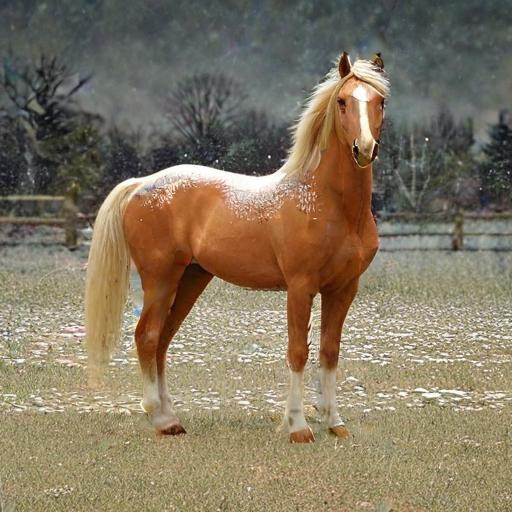} &
        \includegraphics[width=0.13\textwidth,height=0.13\textwidth]{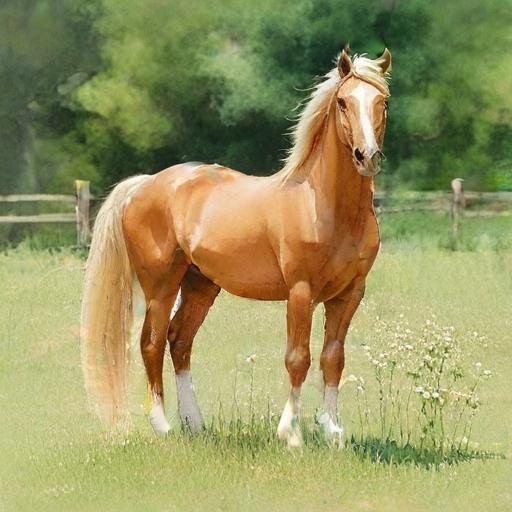} &
        \includegraphics[width=0.13\textwidth,height=0.13\textwidth]{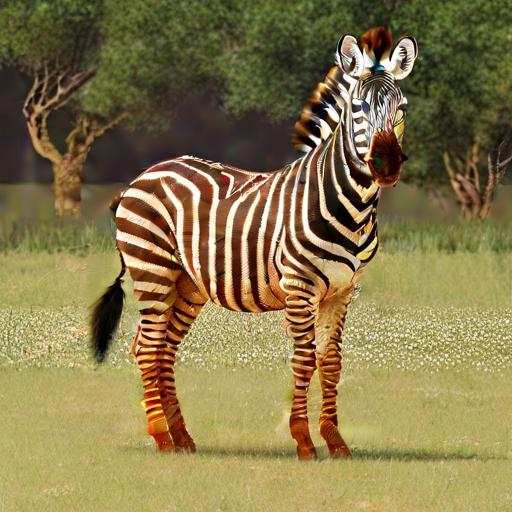} &
        \includegraphics[width=0.13\textwidth,height=0.13\textwidth]{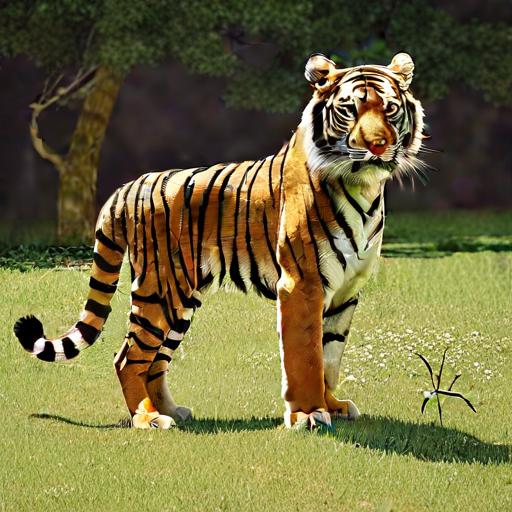} &
        \includegraphics[width=0.13\textwidth,height=0.13\textwidth]{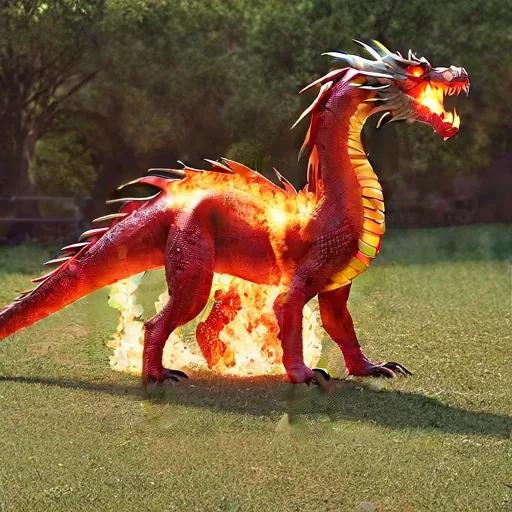} &
        \includegraphics[width=0.13\textwidth,height=0.13\textwidth]{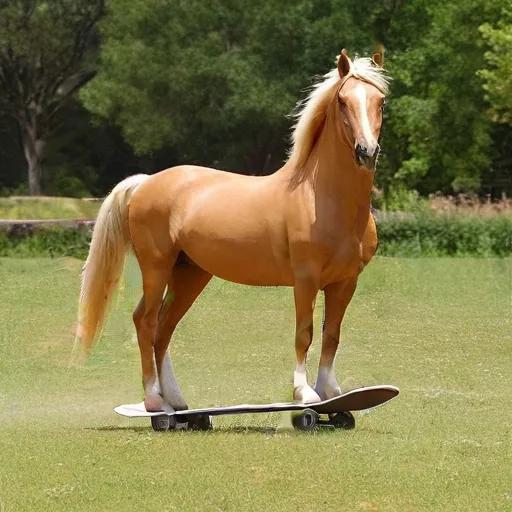} \\

        \vspace{2pt} & \footnotesize ``Snowing''  & \footnotesize ``Watercolor'' & \footnotesize ``Zebra''  & \footnotesize ``Tiger'' & \footnotesize ``Dragon'' & \footnotesize ``On a skateboard''\\

        \vspace{-4pt} \includegraphics[width=0.13\textwidth,height=0.13\textwidth]{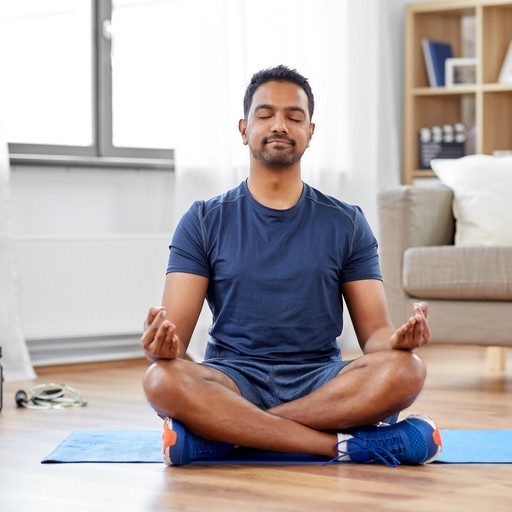} &
        \includegraphics[width=0.13\textwidth,height=0.13\textwidth]{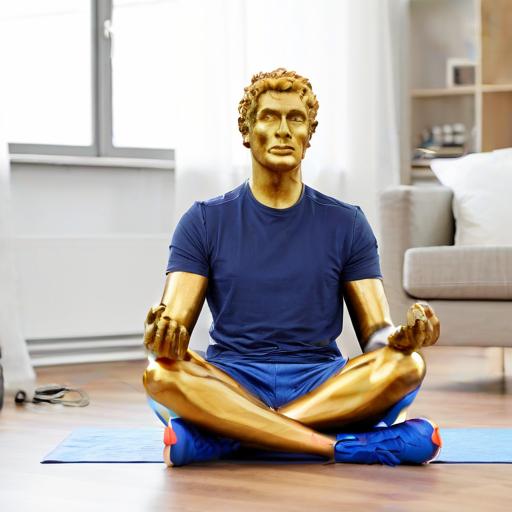} &
        \includegraphics[width=0.13\textwidth,height=0.13\textwidth]{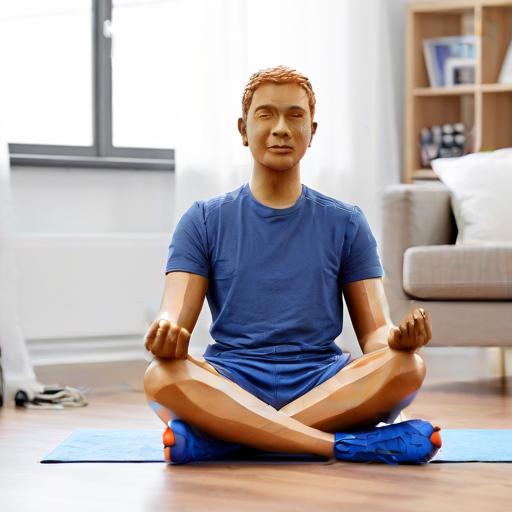} &
        \includegraphics[width=0.13\textwidth,height=0.13\textwidth]{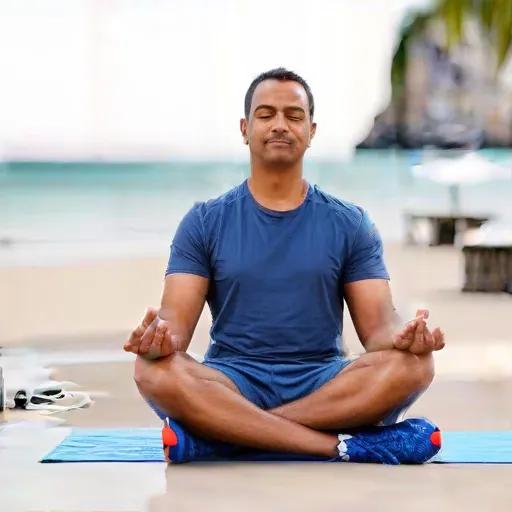} &
        \includegraphics[width=0.13\textwidth,height=0.13\textwidth]{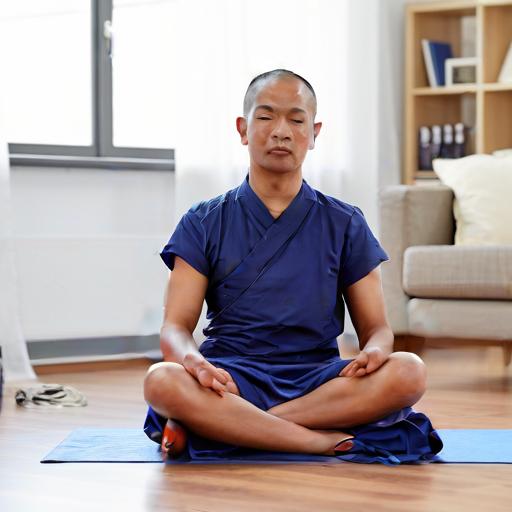} &
        \includegraphics[width=0.13\textwidth,height=0.13\textwidth]{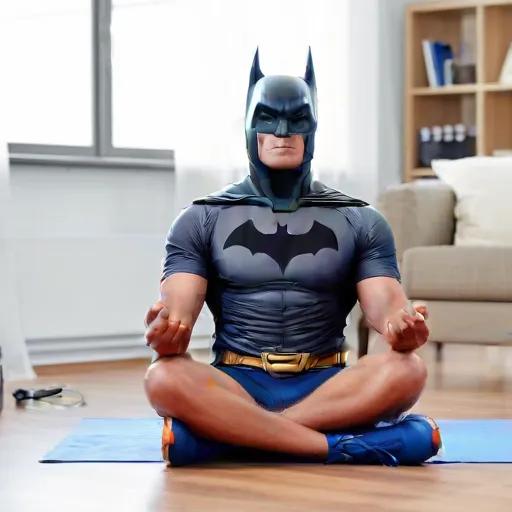} &
        \includegraphics[width=0.13\textwidth,height=0.13\textwidth]{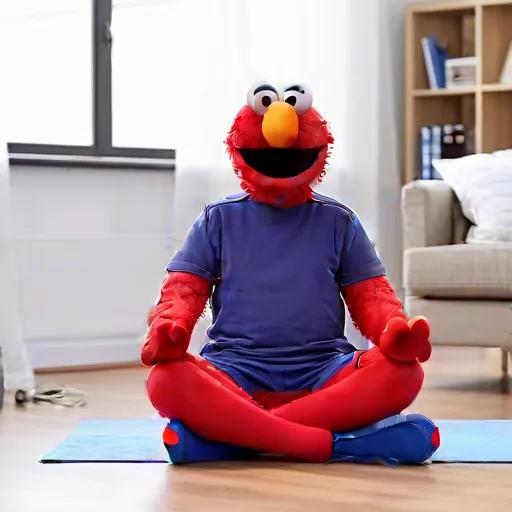} \\

        \vspace{2pt} & \footnotesize ``Golden sculpture''  & \footnotesize ``Wooden statue'' & \footnotesize ``At the beach'' & \footnotesize ``Monk'' & \footnotesize ``Batman'' & \footnotesize ``Elmo'' \\

        \vspace{-4pt} \includegraphics[width=0.13\textwidth,height=0.13\textwidth]{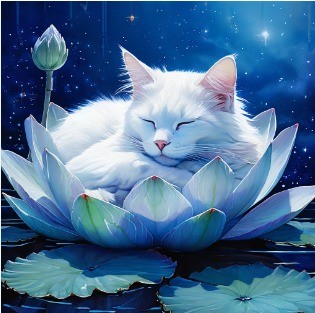} &
        \includegraphics[width=0.13\textwidth,height=0.13\textwidth]{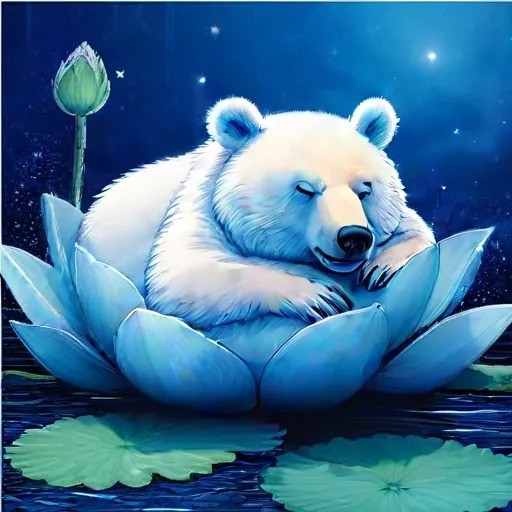} &
        \includegraphics[width=0.13\textwidth,height=0.13\textwidth]{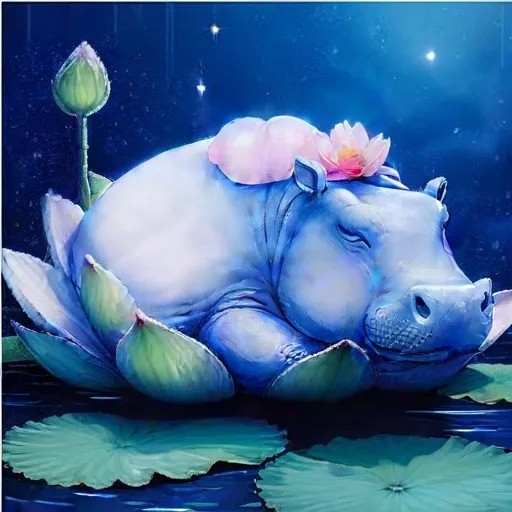} &
        \includegraphics[width=0.13\textwidth,height=0.13\textwidth]{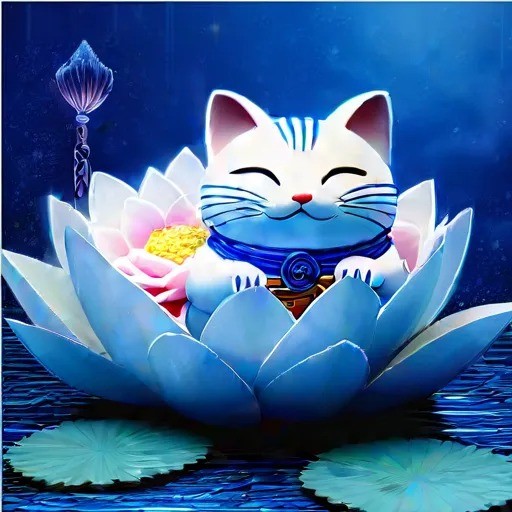} &
        \includegraphics[width=0.13\textwidth,height=0.13\textwidth]{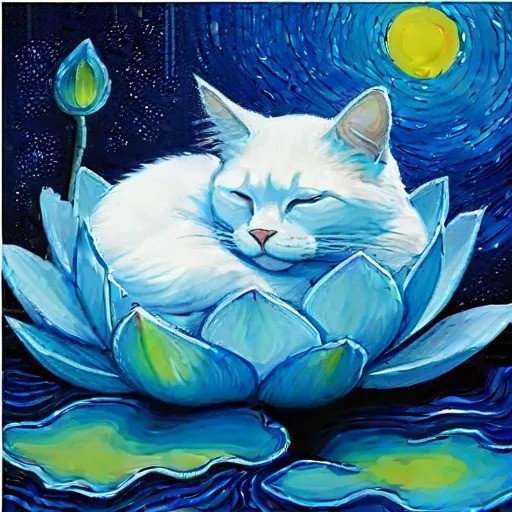} &
        \includegraphics[width=0.13\textwidth,height=0.13\textwidth]{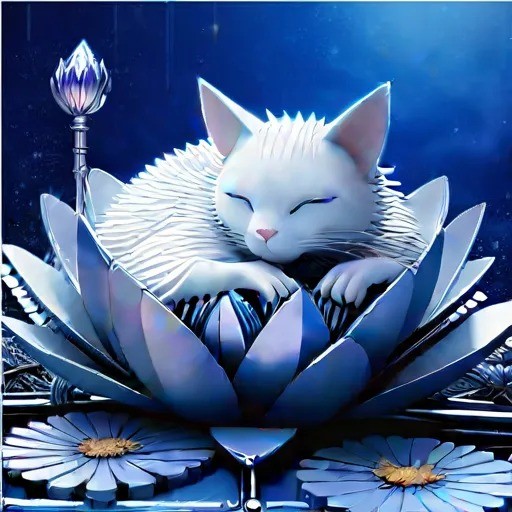} &
        \includegraphics[width=0.13\textwidth,height=0.13\textwidth]{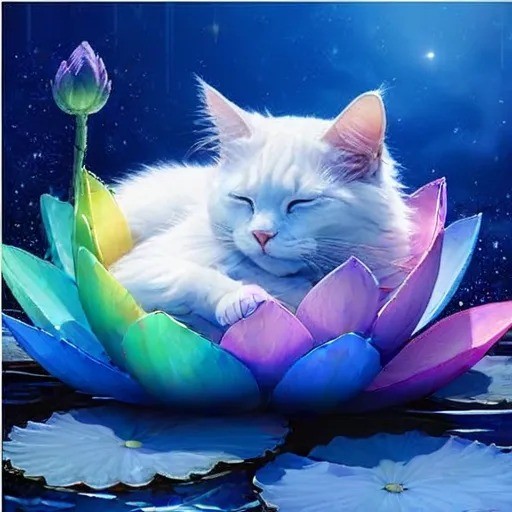} \\

        \vspace{2pt} & \footnotesize ``Bear'' & \footnotesize ``Hippo'' & \footnotesize ``Maneki Neko'' & \footnotesize ``Van Gogh'' & \footnotesize ``Made of metal'' & \footnotesize ``Colorful lotus'' \\

        \vspace{-4pt} \includegraphics[width=0.13\textwidth,height=0.13\textwidth]{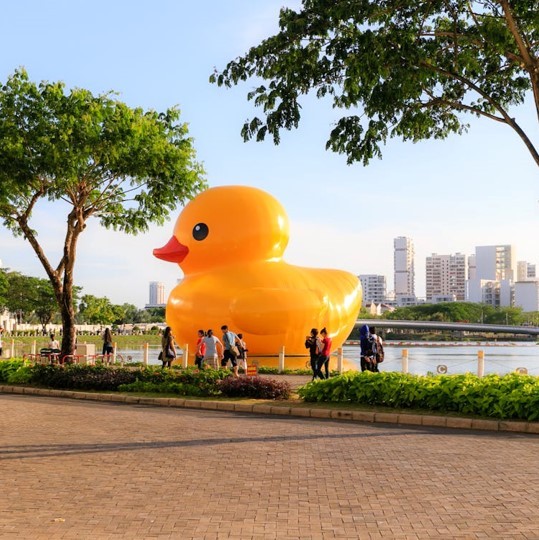} &
        \includegraphics[width=0.13\textwidth,height=0.13\textwidth]{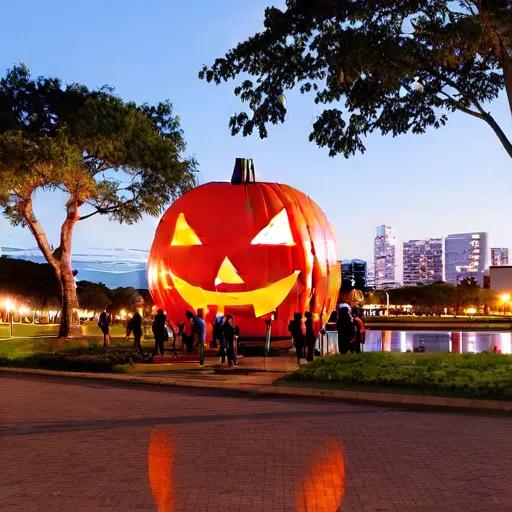} &
        \includegraphics[width=0.13\textwidth,height=0.13\textwidth]{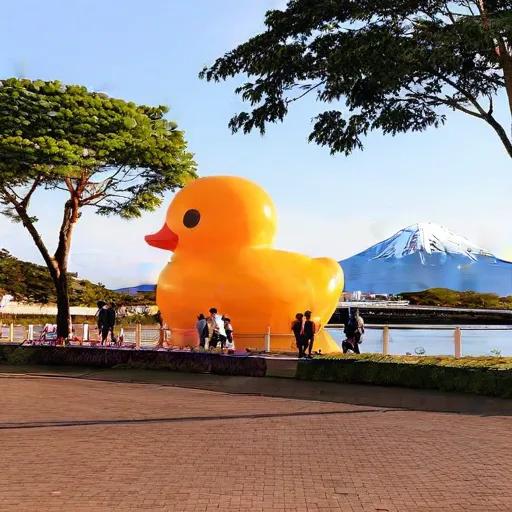} &
        \includegraphics[width=0.13\textwidth,height=0.13\textwidth]{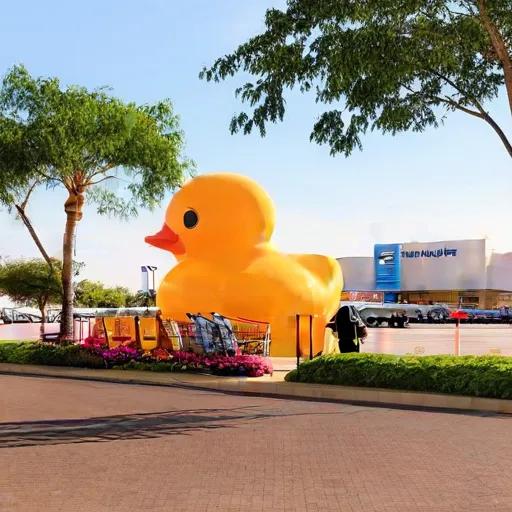} &
        \includegraphics[width=0.13\textwidth,height=0.13\textwidth]{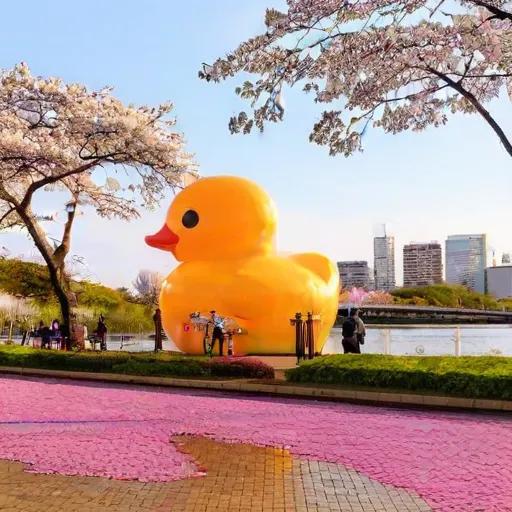} &
        \includegraphics[width=0.13\textwidth,height=0.13\textwidth]{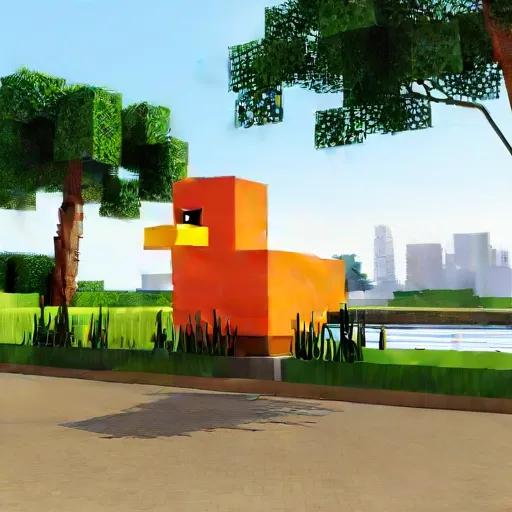} &
        \includegraphics[width=0.13\textwidth,height=0.13\textwidth]{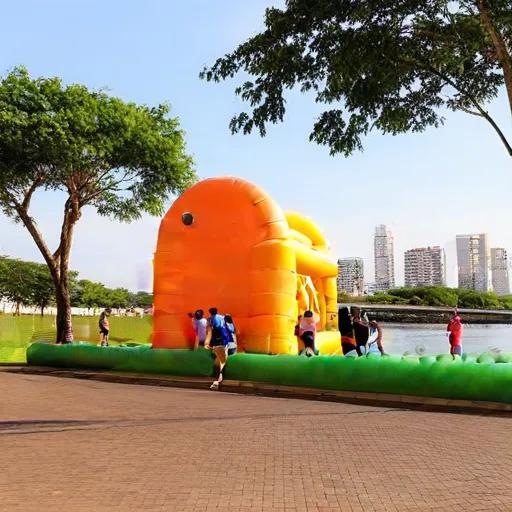} \\

        \vspace{2pt} & \footnotesize ``Halloween pumpkin'' & \footnotesize ``Near Fuji'' & \footnotesize ``Shopping at Walmart'' & \footnotesize ``Cherry blossom'' & \footnotesize ``Minecraft graphics'' & \footnotesize ``Bouncy castle'' \\

        \vspace{-4pt} \includegraphics[width=0.13\textwidth,height=0.13\textwidth]{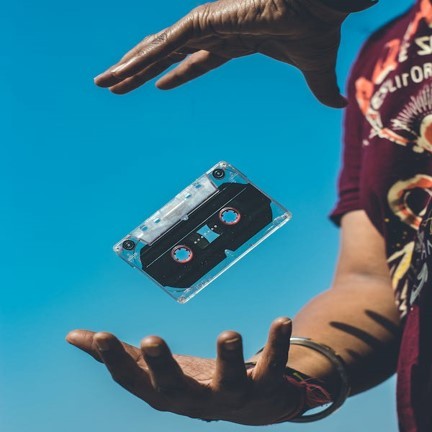} &
        \includegraphics[width=0.13\textwidth,height=0.13\textwidth]{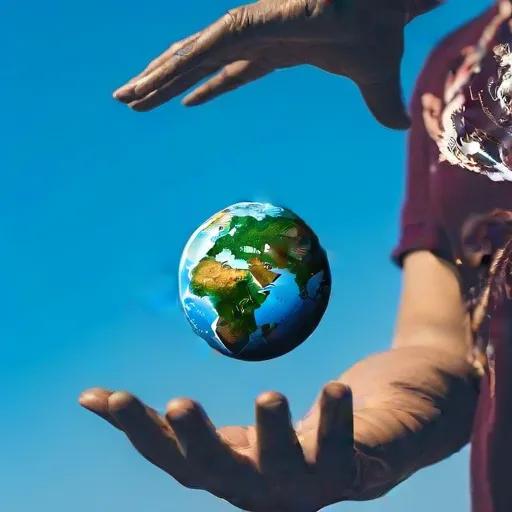} &
        \includegraphics[width=0.13\textwidth,height=0.13\textwidth]{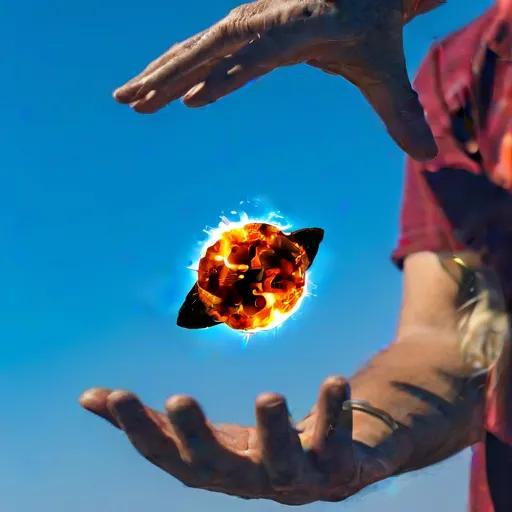} &
        \includegraphics[width=0.13\textwidth,height=0.13\textwidth]{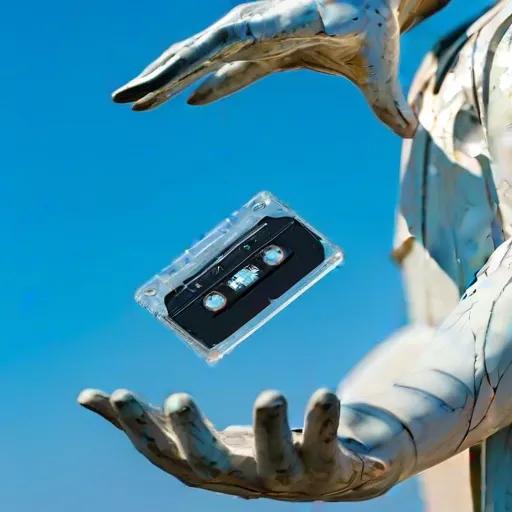} &
        \includegraphics[width=0.13\textwidth,height=0.13\textwidth]{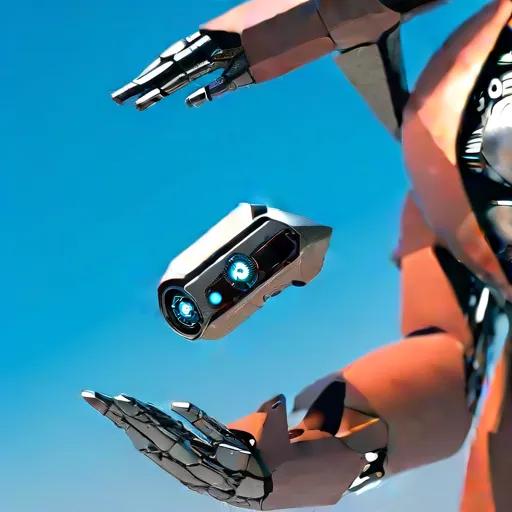} &
        \includegraphics[width=0.13\textwidth,height=0.13\textwidth]{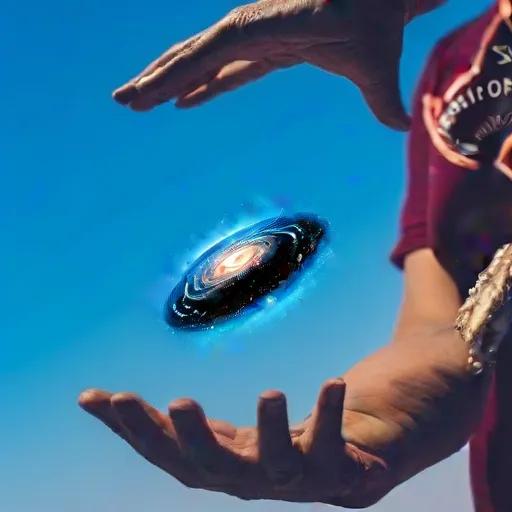} &
        \includegraphics[width=0.13\textwidth,height=0.13\textwidth]{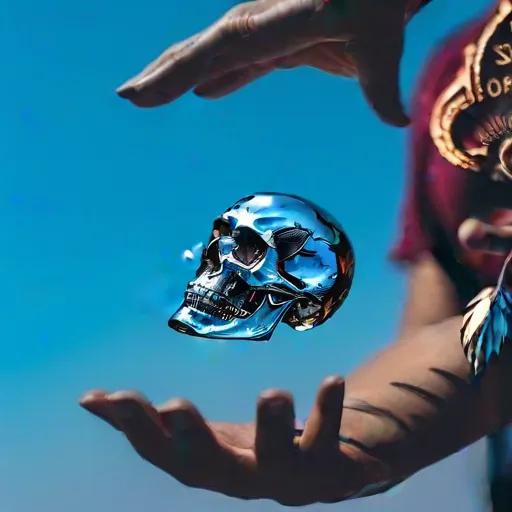} \\

        \vspace{2pt} & \footnotesize ``The Earth'' & \footnotesize ``Fireball'' & \footnotesize ``Marble sculpture'' & \footnotesize ``Futuristic robot'' & \footnotesize ``The galaxy'' & \footnotesize ``Crystal skull'' \\

        \vspace{-4pt} \includegraphics[width=0.13\textwidth,height=0.13\textwidth]{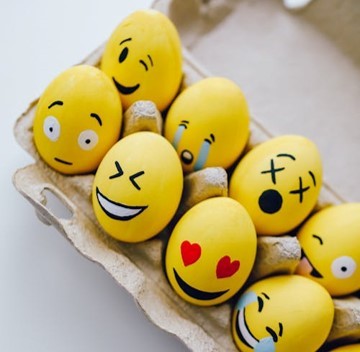} &
        \includegraphics[width=0.13\textwidth,height=0.13\textwidth]{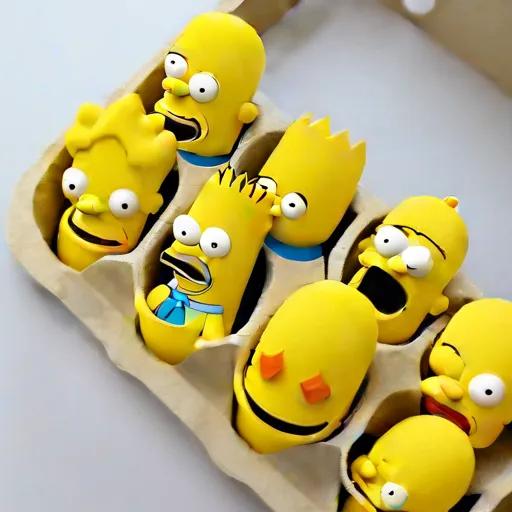} &
        \includegraphics[width=0.13\textwidth,height=0.13\textwidth]{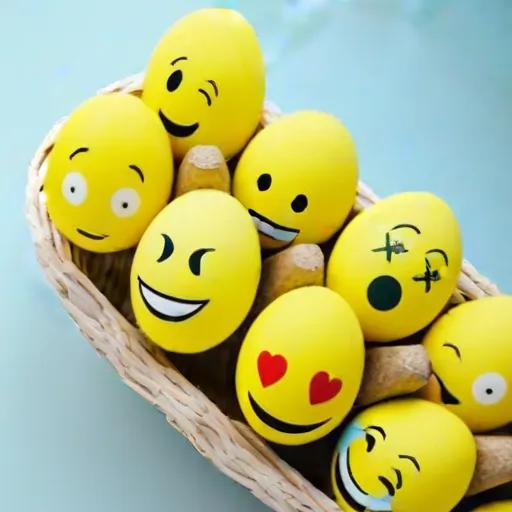} &
        \includegraphics[width=0.13\textwidth,height=0.13\textwidth]{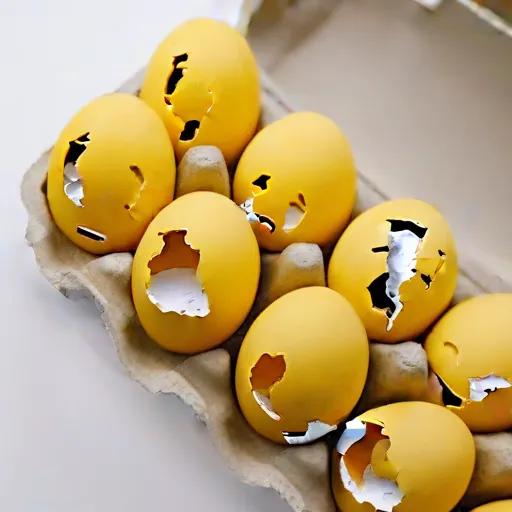} &
        \includegraphics[width=0.13\textwidth,height=0.13\textwidth]{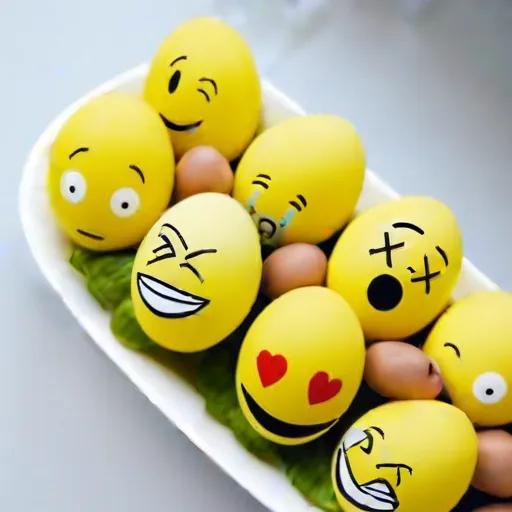} &
        \includegraphics[width=0.13\textwidth,height=0.13\textwidth]{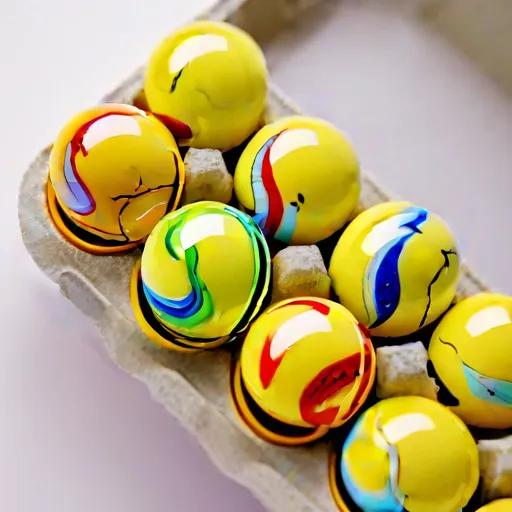} &
        \includegraphics[width=0.13\textwidth,height=0.13\textwidth]{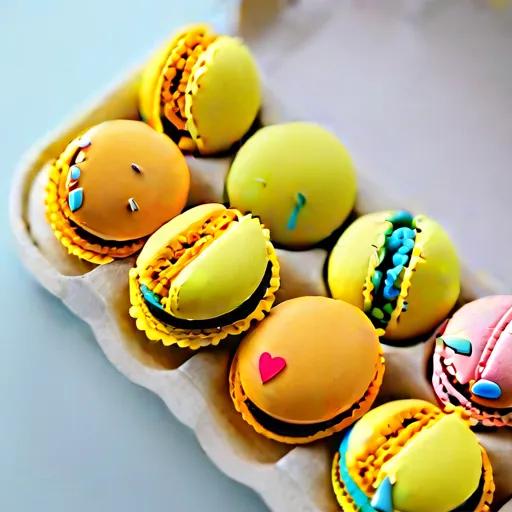} \\

        \vspace{2pt} & \footnotesize ``Simpsons characters'' & \footnotesize ``In a basket'' & \footnotesize ``Cracked'' & \footnotesize ``In a salad'' & \footnotesize ``Marbles'' & \footnotesize ``Macaroons'' \\

        \vspace{-4pt} \includegraphics[width=0.13\textwidth,height=0.13\textwidth]{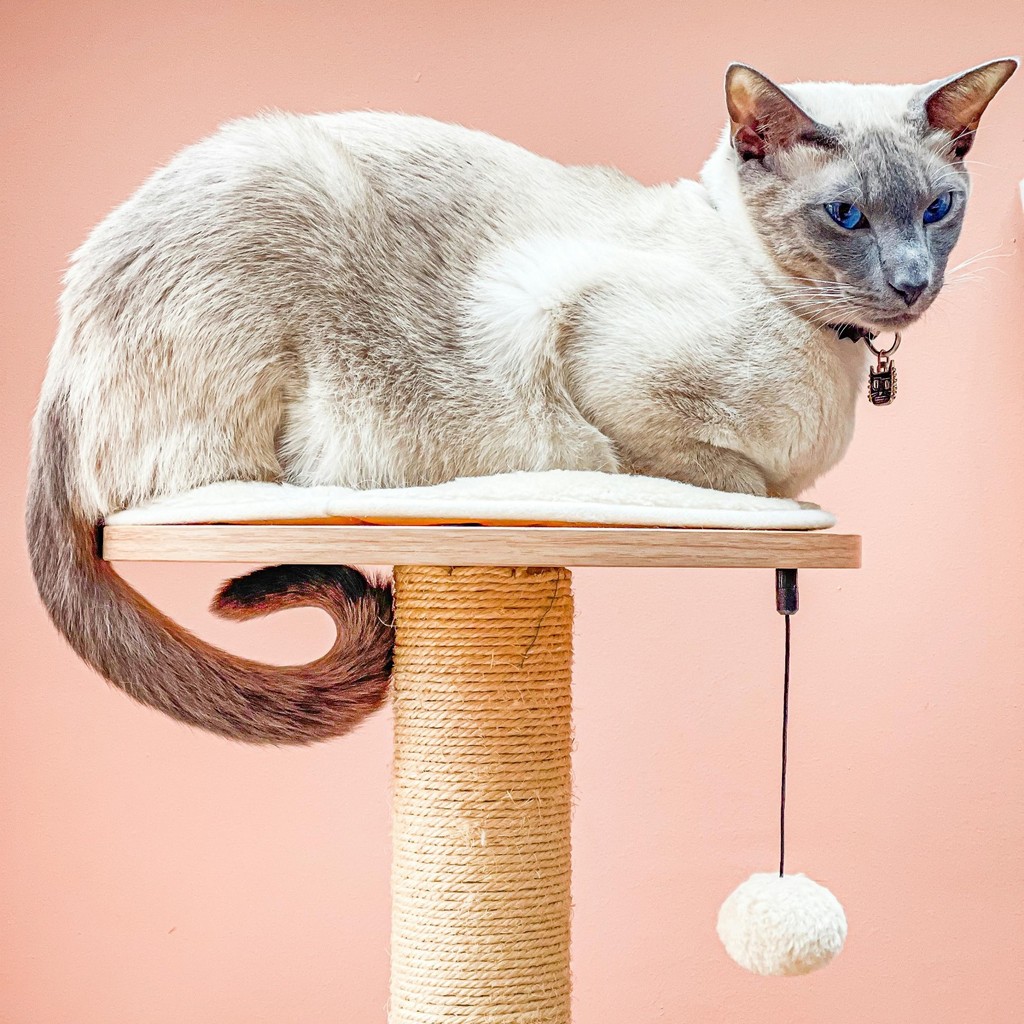} &
        \includegraphics[width=0.13\textwidth,height=0.13\textwidth]{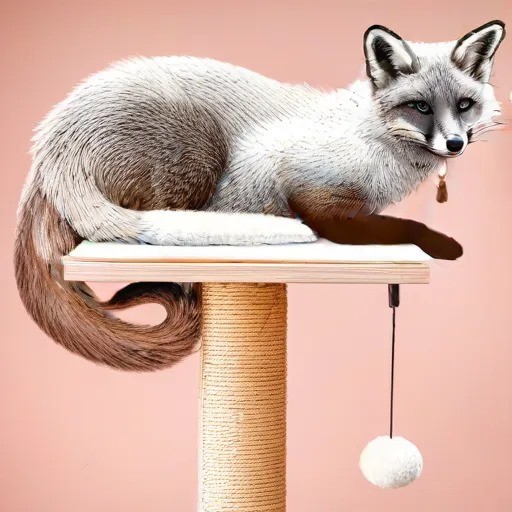} &
        \includegraphics[width=0.13\textwidth,height=0.13\textwidth]{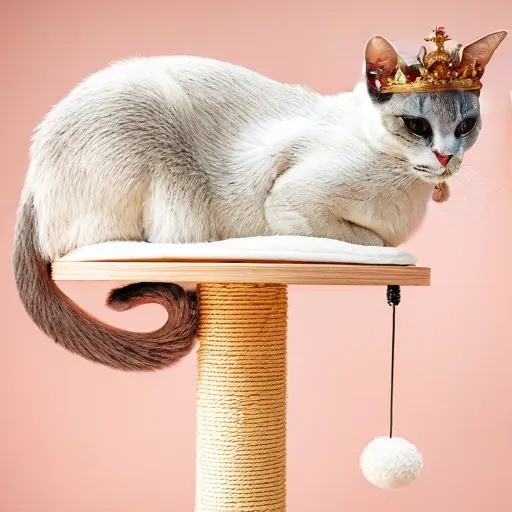} &
        \includegraphics[width=0.13\textwidth,height=0.13\textwidth]{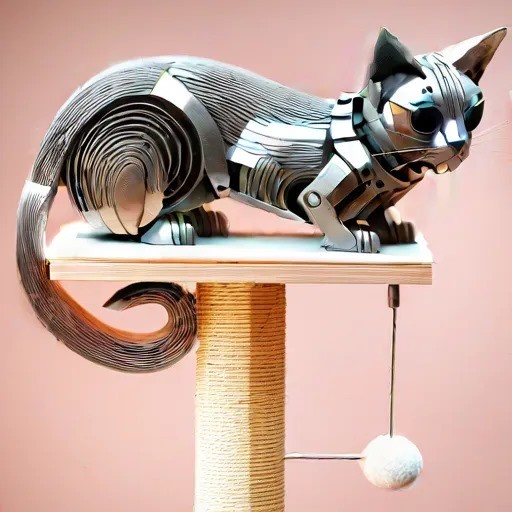} &
        \includegraphics[width=0.13\textwidth,height=0.13\textwidth]{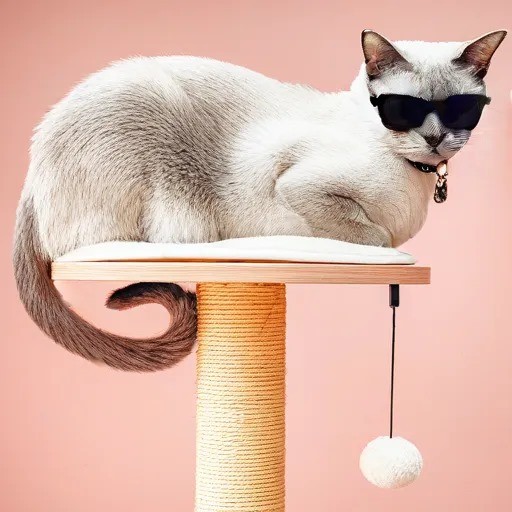} &
        \includegraphics[width=0.13\textwidth,height=0.13\textwidth]{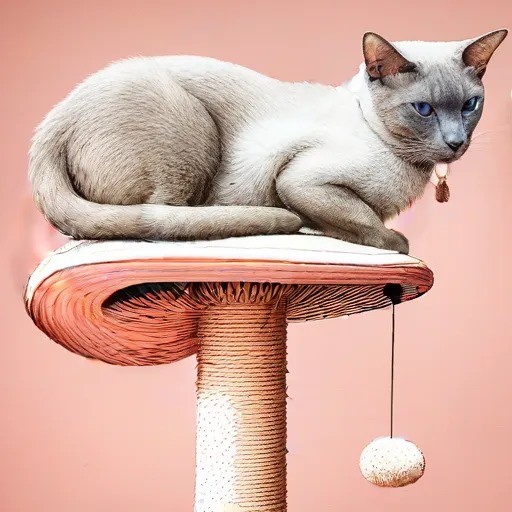} &
        \includegraphics[width=0.13\textwidth,height=0.13\textwidth]{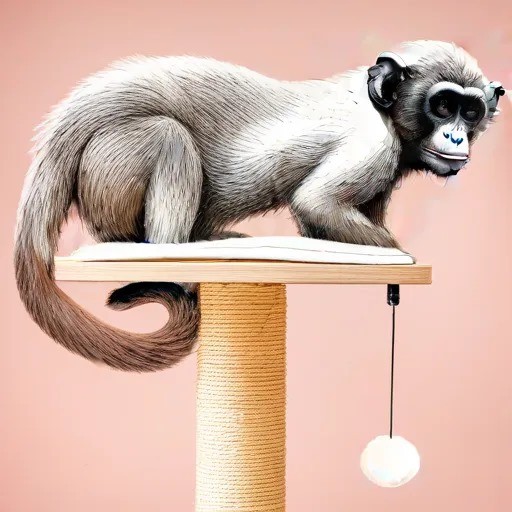} \\

        \vspace{2pt} & \footnotesize ``Fox'' & \footnotesize ``Wearing a crown'' & \footnotesize ``Robot'' & \footnotesize ``Wearing sunglasses'' & \footnotesize ``On a mushroom'' & \footnotesize ``Monkey'' \\

    \end{tabular}
    
    }
    \caption{Additional qualitative editing results of our method. All results use $4$ diffusion steps.}\label{fig:ours_only_extra}
\end{figure*}

\bibliographystyle{ACM-Reference-Format}
\bibliography{main}

\clearpage

\clearpage
\appendix

\begin{center} \huge
Supplementary Materials
\end{center}

\section{Connection between pseudo-guidance and Classifier-free Guidance}
\label{sec:leaner_guidance}
Let us now provide proof, that under the assumption that 
\begin{equation}\label{eq:approx_directions}
\mu_{t}{(\hat{x}_t, c)} - \mu_{t}{(x_t, c)} \approx \mu_{t}{(\hat{x}_t, \phi)} - \mu_{t}{(x_t, \phi)}
\end{equation}
Our proposed guidance is equivalent to using CFG but with fewer network evaluations. We consider the case where CFG is applied both during inversion and during generation. Let $\munet{x, \phi + \lambda_s \cdot (c - \phi)}$ be the notation for a U-net prediction followed by a scheduler step with CFG strength $\lambda_s$. Then, the following holds:
\begin{equation}
\munet{x, \phi + \lambda_s \cdot (c - \phi)} = \munet{x, \phi} + \lambda_s \cdot [\munet{x, c} - \munet{x, \phi}] ,
\end{equation}
\ie, one can first perform a scheduler step after summing the U-net predictions, or first perform two scheduler steps and then sum the results.
We can now write the inference equation as:
\begin{equation}
    \xhat{t-1} = x_{t-1} + [\munet{\xhat{t}, \phi + \lambda_s(\chat - \phi)} - \munet{x_t, \phi + \lambda_s(c - \phi)}] .
\end{equation}
Adding and subtracting the cross term $\mu_{t}{(\hat{x}_t, \phi + \lambda_s(c - \phi))}$, we get:
\begin{alignat}{2}
    \xhat{t-1} &= x_{t-1} &&+ [ \munet{\xhat{t}, \phi + \lambda_s(\chat - \phi)} \nonumber \\
            & &&- \mu_{t}{(\hat{x}_t, \phi + \lambda_s(c - \phi))} \nonumber \\
            & &&+ \mu_{t}{(\hat{x}_t, \phi + \lambda_s(c - \phi))} \nonumber \\
            & &&- \munet{x_t, \phi + \lambda_s(c - \phi)}] \nonumber \\
            &= x_{t-1} &&+ [\munet{\xhat{t}, \phi} + \lambda_s[\munet{\xhat{t}, \chat} - \munet{\xhat{t}, \phi}] \nonumber \\
            & &&- \munet{\xhat{t}, \phi} - \lambda_s[\munet{\xhat{t}, c} - \munet{\xhat{t}, \phi}] \nonumber \\ 
            & &&+ \munet{\xhat{t}, \phi} + \lambda_s[\munet{\xhat{t}, c} - \munet{\xhat{t}, \phi}] \nonumber \\ 
            & &&- \munet{x_t, \phi} - \lambda_s[\munet{x_t, c} - \munet{x_t, \phi}]] \nonumber \\
            &= x_{t-1} &&+ \munet{\xhat{t}, \phi} - \munet{x_t, \phi} + \lambda_s [\nonumber \\ 
            & && \munet{\xhat{t}, \chat} - \munet{\xhat{t}, \phi} \nonumber - \munet{\xhat{t}, c} + \munet{\xhat{t}, \phi} \nonumber \\ 
            & &&+ \munet{\xhat{t}, c} - \munet{\xhat{t}, \phi} \nonumber - \munet{x_t, c} + \munet{x_t, \phi} \nonumber] \\
            &= x_{t-1} &&+ \munet{\xhat{t}, \phi} - \munet{x_t, \phi} + \lambda_s [ \\ 
            & &&\munet{\xhat{t}, \chat} - \munet{\xhat{t}, c} \nonumber \\
            & &&+ (\munet{\xhat{t}, c} - \munet{x_t, c}) - (\munet{\xhat{t}, \phi} - \munet{x_t, \phi})] 
\end{alignat}
If $\mu_{t}{(\hat{x}_t, c)} - \mu_{t}{(x_t, c)} \approx \mu_{t}{(\hat{x}_t, \phi)} - \mu_{t}{(x_t, \phi)}$ 
then the term in line (15) is approximately zero and line (14) can be replaced: 
$\munet{\xhat{t}, \phi} - \munet{x_t, \phi} \rightarrow \munet{\xhat{t}, c} - \munet{x_t, c}$ 
and we get that DDPM inversion with CFG is equivalent to our proposed guidance. Empirically, we find that relation of \cref{eq:approx_directions} often holds for $3$ and $4$ steps with SDXL-Turbo, but we offer no guarantees for the general case.

\section{Further analysis on time shifts}

\begin{figure}[ht]
    \centering
    \includegraphics[width=0.97\linewidth,trim={0 0 0 2cm},clip]{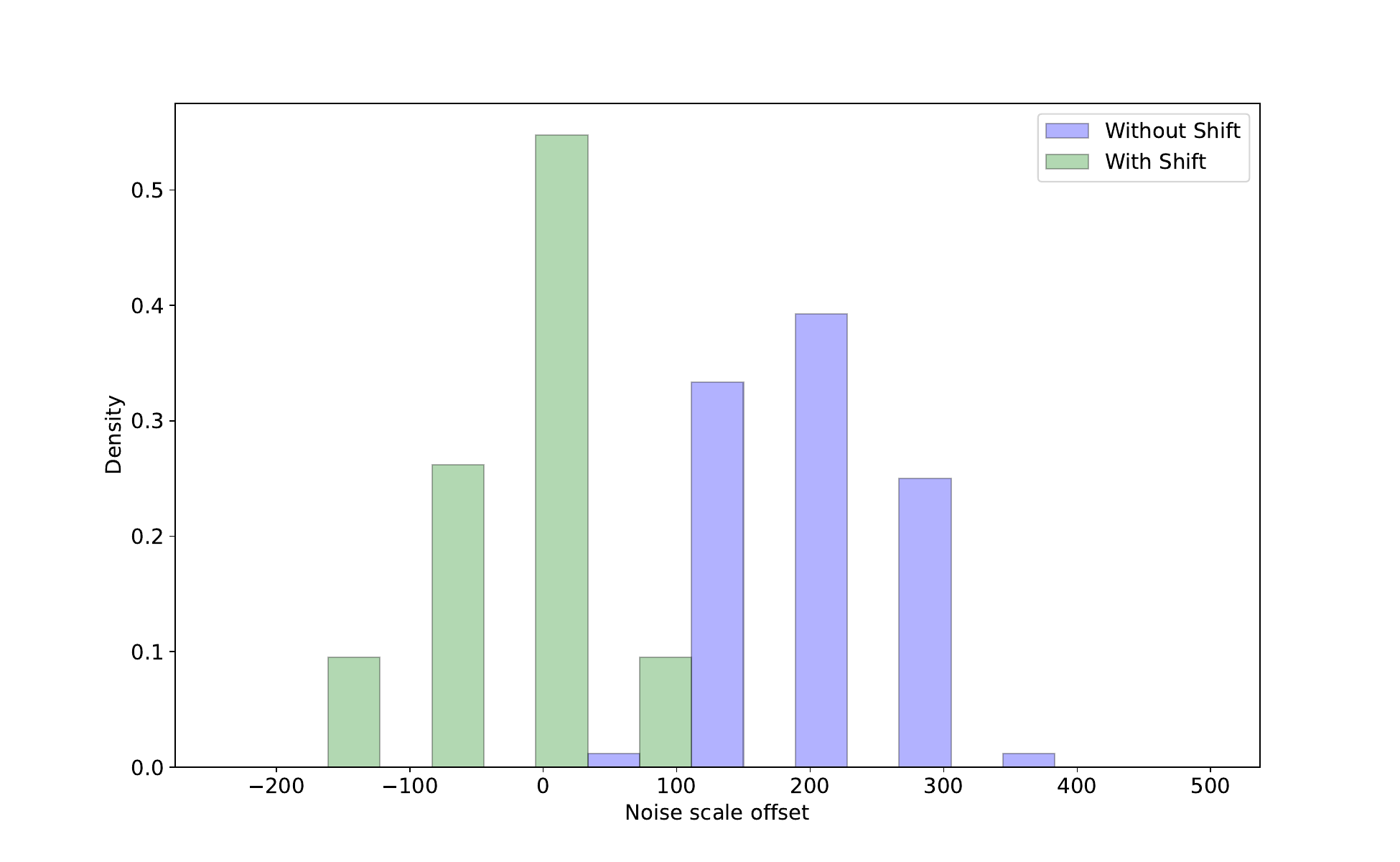}\vspace{-5pt}
    \caption{For each image in our evaluation set, and for each denoising time step, we calculate the time-offset between the inverted noises and the closest point on the true noise schedule that has the same level of noise. With vanilla edit-friendly DDPM inversion (blue), nearly all offsets are concentrated in the $[100, 300]$ step region. When inverting using a shifted schedule (green), the offsets are more concentrated around $0$.}\label{fig:offset_histogram}\vspace{-5pt}
\end{figure}

Here we provide further analysis on the behaviour of the inversion noise scales, and demonstrate the importance of applying the shifts throughout the entire generation process. In \cref{fig:offset_histogram}(blue) we plot the distribution of offsets between the DDPM-inversion corrections and the denoising step for which the true noise schedule is closest to this step. These shifts are approximately distributed around the $200$-step shift, which inspires our choice to correct for them through a $200$ step offset. We further plot the same distribution of offsets after applying our fix. Now, most predicted noises match the step in which they are used, and the overall shift is diminished.

\begin{figure}
    \centering
    \setlength{\tabcolsep}{4pt}
    \begin{tabular}{c c c c}
        
        & 
        \begin{tabular}{@{}c@{}} \small Final step \\[-0.5ex] \small timestep shift \end{tabular}
        & \begin{tabular}{@{}c@{}} \small Full \\[-0.5ex] \small timestep shift \end{tabular}
        \\
        & 

        \includegraphics[width=0.1\textwidth,height=0.1\textwidth]{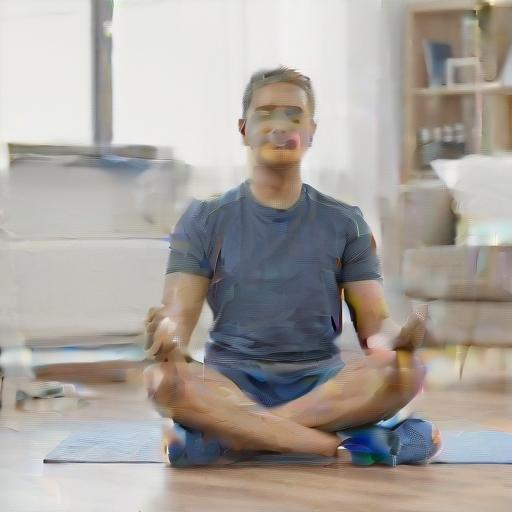} &
        \includegraphics[width=0.1\textwidth,height=0.1\textwidth]{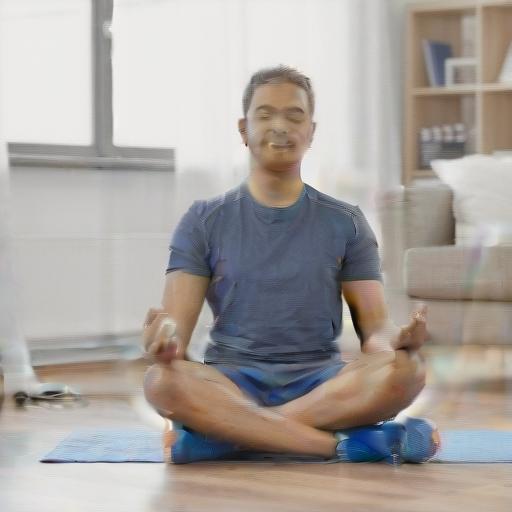} & 

        \raisebox{0.04\textwidth}{\rotatebox[origin=t]{-90}{\scalebox{0.9}{\begin{tabular}{c@{}c@{}c@{}} Iteration 1 \end{tabular}}}} \\ 

        &
        \includegraphics[width=0.1\textwidth,height=0.1\textwidth]{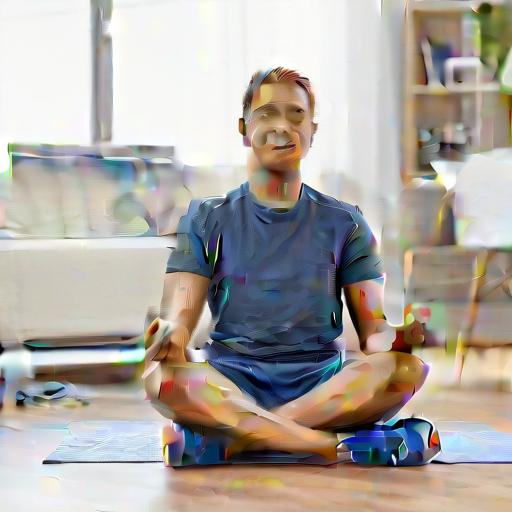} &
        \includegraphics[width=0.1\textwidth,height=0.1\textwidth]{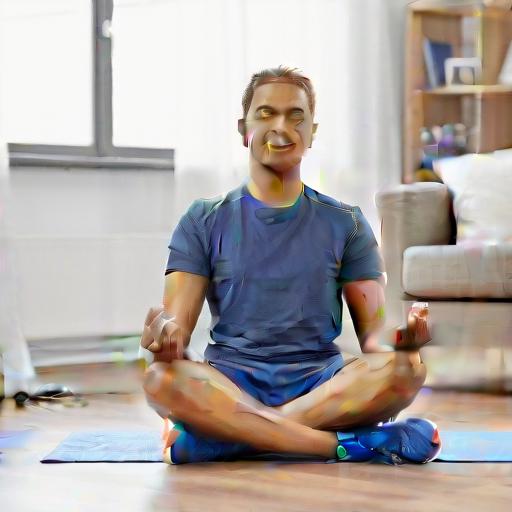} & 

        \raisebox{0.04\textwidth}{\rotatebox[origin=t]{-90}{\scalebox{0.9}{\begin{tabular}{c@{}c@{}c@{}} Iteration 2 \end{tabular}}}} \\ 
    
        &
        \includegraphics[width=0.1\textwidth,height=0.1\textwidth]{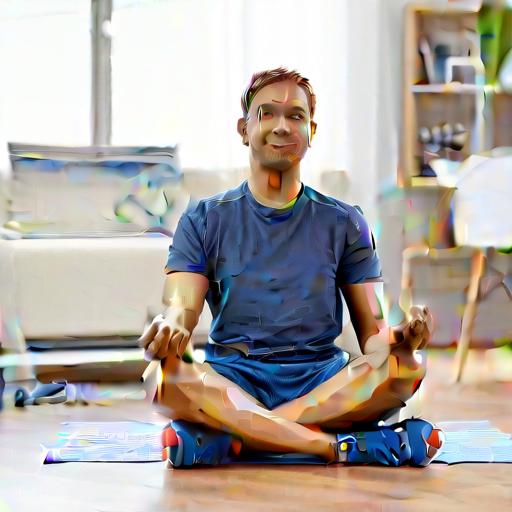} &
        \includegraphics[width=0.1\textwidth,height=0.1\textwidth]{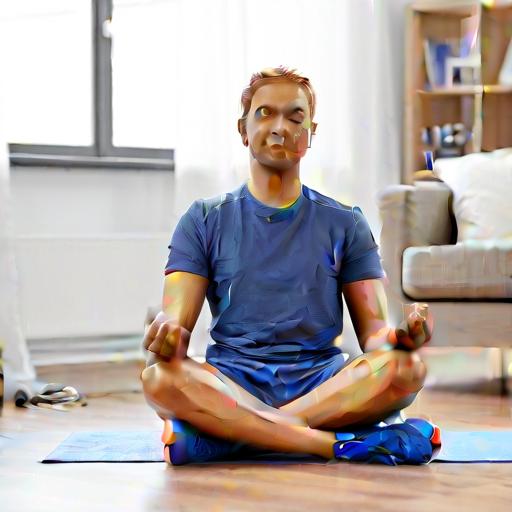} &
    
        \raisebox{0.04\textwidth}{\rotatebox[origin=t]{-90}{\scalebox{0.9}{\begin{tabular}{c@{}c@{}c@{}} Iteration 3 \end{tabular}}}} \\

        \includegraphics[width=0.1\textwidth,height=0.1\textwidth]{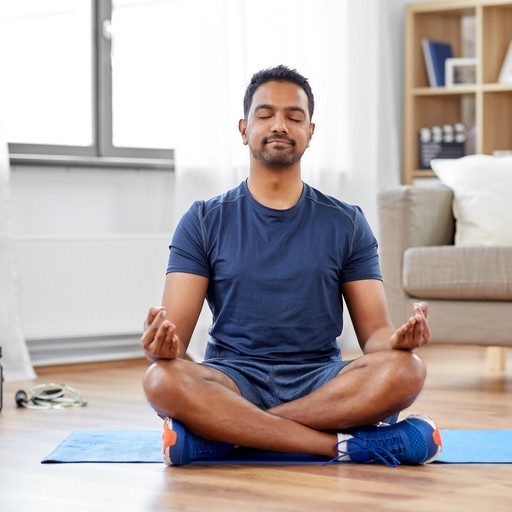} &

        \includegraphics[width=0.1\textwidth,height=0.1\textwidth]{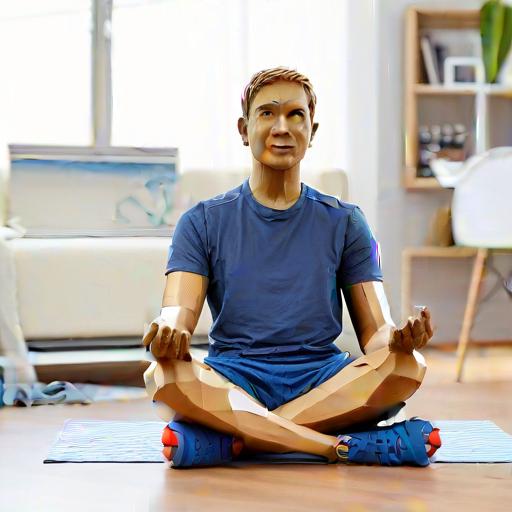} &
        \includegraphics[width=0.1\textwidth,height=0.1\textwidth]{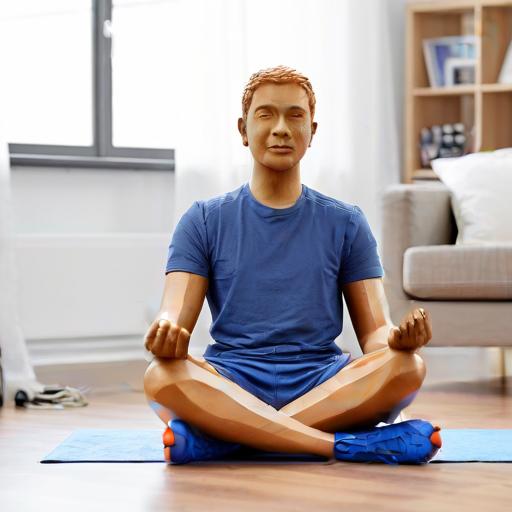} &
    
        \raisebox{0.04\textwidth}{\rotatebox[origin=t]{-90}{\scalebox{0.9}{\begin{tabular}{c@{}c@{}c@{}} Iteration 4 \end{tabular}}}} \\
        
        \small Input Image & \multicolumn{2}{c}{ \small "wooden statue"} & \\
    
    \end{tabular}
    \caption{Comparison of our full timestep shift approach with an extra shifted timestep at the final step.
    We show the edited image at each iteration of the edit process before adding the noise.
    The extra shifted step has a "cleaning" effect that removes some of the artifacts, 
    yet the image is already too damaged and dissimilar to the original image.
    Our full timestep shift approach keeps the image visually appealing at each step.}\label{fig:compare_all_timesteps}
\end{figure}

In \cref{fig:compare_all_timesteps} we ask whether it is important to shift the schedule throughout the entire generative process, or if it is sufficient to shift only the last step. The intuition behind this experiment is that applying only a final-step shift can lead to increased smoothing of the image (due to the removal of extra noise), and possibly account for the artifact removal on its own. However, we find experimentally that this is not the case. As can be seen, the unshifted steps accumulate strong artifacts, and the final step fails to overcome them. Moreover, we observe that this approach leads to worse preservation of background objects. All in all, we find that shifting the entire schedule is beneficial.

\section{Edit-Friendly and Delta-Denoising equivalence}
\label{sec:ef_dds_equivalence}
Here, we provide proof that Delta Denoising Score (DDS) and Edit-Friendly DDPM Inversion (EF) are equivalent, under an appropriate choice of learning rate and timestep sampling for the DDS method.

First, recall the DDPM sampler's definition of $\mu_{t}{(x_t, c)}$:

\begin{equation}
    \mu_{t}{(x_t, c)} = \frac{1}{\sqrt{\alpha_t}}\left(\mathbf{x}_t-\frac{1-\alpha_t}{\sqrt{1-\bar{\alpha}_t}} \boldsymbol{\epsilon}_t^\phi\left(\mathbf{x}_t, c\right)\right) ,
\end{equation}
and the noising equation:
\begin{equation}
    x_t = \sqrt{\bar\alpha_t} x_0 + \sqrt{1-\bar\alpha_t}\, \tilde{\epsilon}_t ,
\end{equation}
where $\alpha_t$ is derived from the diffusion noising schedule and $\bar\alpha_t = \prod_{s=1}^t \alpha_s$ and $\boldsymbol{\epsilon}_t^\phi$ denotes the diffusion model's output.

For simplicity, we denote $a_t = \frac{1}{\sqrt{\alpha_t}}$, $b_t = \frac{1-\alpha_t}{\sqrt{1-\bar{\alpha}_t}}$, $c_t = \sqrt{\bar\alpha_t}$ and $d_t = \sqrt{1-\bar\alpha_t}$, and re-write these equations as:
\begin{equation}\label{eq:mu_proof}
    \mu_{t}{(x_t, c)} = a_t\left({x}_t-b_t{\epsilon}_t^\phi\left({x}_t, c\right)\right) ,
\end{equation}
\begin{equation}\label{eq:noise_proof}
    x_t = c_t x_0 + d_t \tilde{\epsilon}_t
\end{equation}

Next, recall that DDS operates through an iterative optimization scheme. Let $x$ be the original image, $\hat{x}^i$ the optimized image at iteration $i$, and let $x_t$ and $\hat{x}^i_t$ be their respective representations after noising to time $t$ via \cref{eq:noise_proof}. The DDS gradient is then:
\begin{equation}
\nabla_{\hat{x}} \mathcal{L}_{\text{DDS}} = \epsilon_{t}^{\phi}\left(\hat{x}^i_t, \hat c\right) - \epsilon_{t}^{\phi} \left(x_t, c\right) ,
\label{Eq:sds_dir}
\end{equation}
where $c$ and $\hat{c}$ are the prompts describing the original image and the edit target respectively and $t$ is sampled randomly for each optimization iteration.
Since this gradient is applied directly to $\hat{x}^i$, we thus have:
\begin{equation}\label{eq:sds_step}
    \hat{x}^{i + 1} = \hat{x}^{i} - \gamma ({\epsilon}_t^\phi(\hat{x}^i_t, \hat{c}) - {\epsilon}_t^\phi({x}_t, {c}))
\end{equation}

Next, we consider DDS under the following two conditions: (1) Rather than sampling diffusion timesteps randomly, we do so sequentially following a standard denoising schedule (\ie we start with $t=T$ and move towards $t=0$). We will thus use $\hat{x}^{t}_0$ to denote the un-noised, DDS-optimized image at the iteration corresponding to timestep $t$, and $\hat{x}^{t}_t$ to denote its noised version at this timestep.
(2) We select a time-dependent learning rate given by:

\begin{equation}
\label{eq:lr}
\gamma_t = \frac{1 - \alpha_t}{\sqrt{\bar{\alpha}_t} \sqrt{1-\bar{\alpha}_t}} = \frac{b_t}{c_t}
\end{equation}

Plugging these conditions into \cref{eq:sds_step} we have:

\begin{align*}
    \hat{x}^{t-1}_0 &= \hat{x}^{t}_0 - \frac{b_t}{c_t} ({\epsilon}_t^\phi(\hat{x}^t_t, \hat{c}) - {\epsilon}_t^\phi({x}_t, {c})) \\
    &= = \hat{x}^{t}_0 - \frac{b_t}{c_t} ({\epsilon}_t^\phi(\hat{x}^t_t, \hat{c}) - {\epsilon}_t^\phi({x}_t, {c})) + \frac{1}{c_t} x_t - \frac{1}{c_t} x_t ,
\end{align*}
and by repeatedly applying \cref{eq:mu_proof,eq:noise_proof} we have:
\begin{align}\
     \hat{x}^{t-1}_0 &=  \hat{x}^{t}_0 - \frac{b_t}{c_t} ({\epsilon}_t^\phi(\hat{x}^{t}_t, \hat{c}) - {\epsilon}_t^\phi({x}_t, {c})) + \frac{1}{c_t} x_t - \frac{1}{c_t} x_t \nonumber \\ \nonumber
                  &= \hat{x}^{t}_0 + \frac{b_t}{c_t} {\epsilon}_t^\phi({x}_t, {c}) - \frac{1}{c_t} x_t - \frac{b_t}{c_t}{\epsilon}_t^\phi(\hat{x}^{t}_t, \hat{c}) + x_0 + \frac{d_t}{c_t} \tilde{\epsilon}_t \\ \nonumber
                  &= \frac{1}{c_t} (c_t \hat{x}^{t}_0 + d_t \tilde{\epsilon}_t ) - \frac{1}{c_t} (x_t - b_t {\epsilon}_t^\phi({x}_t, {c})) - \frac{b_t}{c_t}{\epsilon}_t^\phi(\hat{x}_t, \hat{c}) + x_0 \\ \nonumber
                  &= \frac{1}{c_t} \hat{x}_t - \frac{1}{c_t a_t} \mu(x_t, c) - \frac{b_t}{c_t}{\epsilon}_t^\phi(\hat{x}_t, \hat{c}) + x_0 \\ \nonumber
                  &= x_0 + \frac{1}{c_t} (\hat{x}_t  - b_t {\epsilon}_t^\phi(\hat{x}_t, \hat{c})) - \frac{1}{c_t a_t} \mu(x_t, c) \\ \label{eq:unnoised_dds}
                  &= x_0 + \frac{1}{c_t a_t} (\mu(\hat{x}_t, \hat{c}) - \mu(x_t, c)) .
\end{align}

From the definitions of $c_t$, $a_t$ and $\bar\alpha_t$ we further have:
\begin{equation}
    \frac{1}{c_t a_t} = \left(\sqrt{\prod_{s=1}^t \alpha_s} \frac{1}{\sqrt{{\alpha}_t}}\right)^{-1} = \left(\sqrt{\prod_{s=1}^{t-1} \alpha_s}\right)^{-1} = \frac{1}{c_{t-1}}
\end{equation}

Adding noise (\cref{eq:noise_proof}) to \cref{eq:unnoised_dds} and plugging in this constant relation gives:
\begin{align*}
    \hat{x}_{t-1} &= c_{t-1} (x_0 + \frac{1}{c_{t-1}} (\mu(\hat{x}_t, \hat{c})) -  \mu(x_t, c)) + d_{t-1} \tilde{\epsilon}_{t-1} \\
    &= \left(c_{t-1} x_0 + d_{t-1} \tilde{\epsilon}_{t-1}\right) + \left(\mu(\hat{x}_t, \hat{c}) -  \mu(x_t, c)\right)  \\
                  &= x_{t-1} + \left(\mu(\hat{x}_t, \hat{c}) -  \mu(x_t, c)\right)
\end{align*}
which is exactly equal to the edit friendly step (\cref{eq:edit_friendly_inference}).

\begin{figure}
    {
    \setlength{\tabcolsep}{2pt}
    
    \begin{tabular}{c c c c}

    Input & DDS & Edit Friendly & \\
    \includegraphics[width=0.135\textwidth,height=0.135\textwidth]{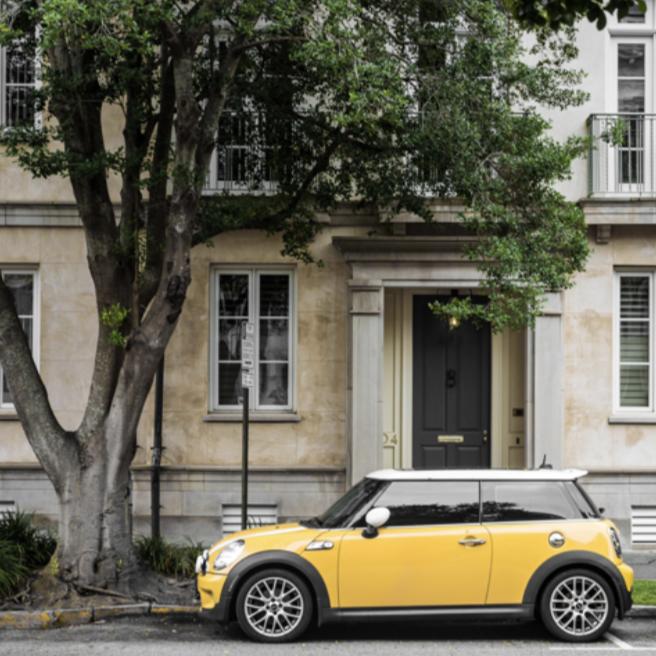} &
    \includegraphics[width=0.135\textwidth,height=0.135\textwidth]{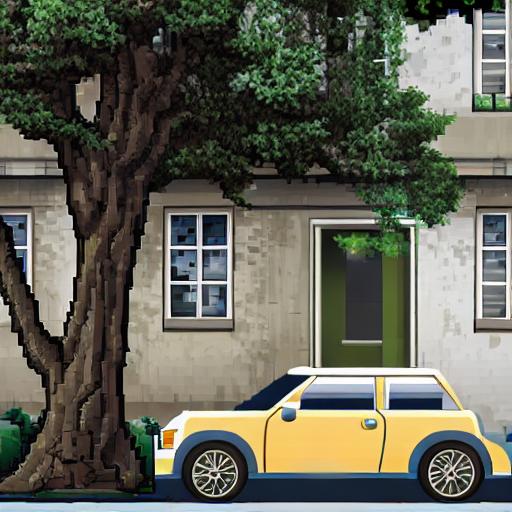} &
    \includegraphics[width=0.135\textwidth,height=0.135\textwidth]{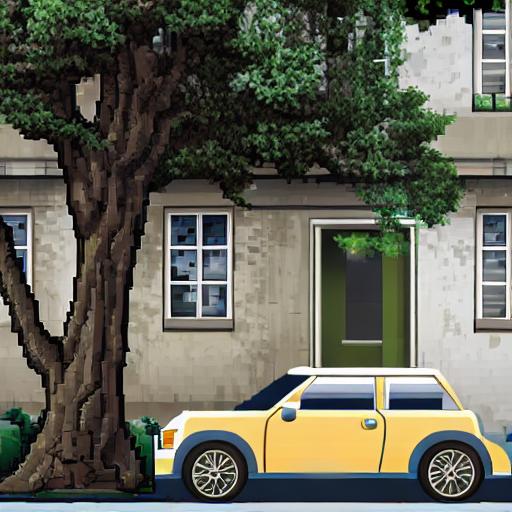} &

    \raisebox{0.06\textwidth}{\rotatebox[origin=t]{-90}{\scalebox{1.0}{Pixel Art}}} \\

    \includegraphics[width=0.135\textwidth,height=0.135\textwidth]{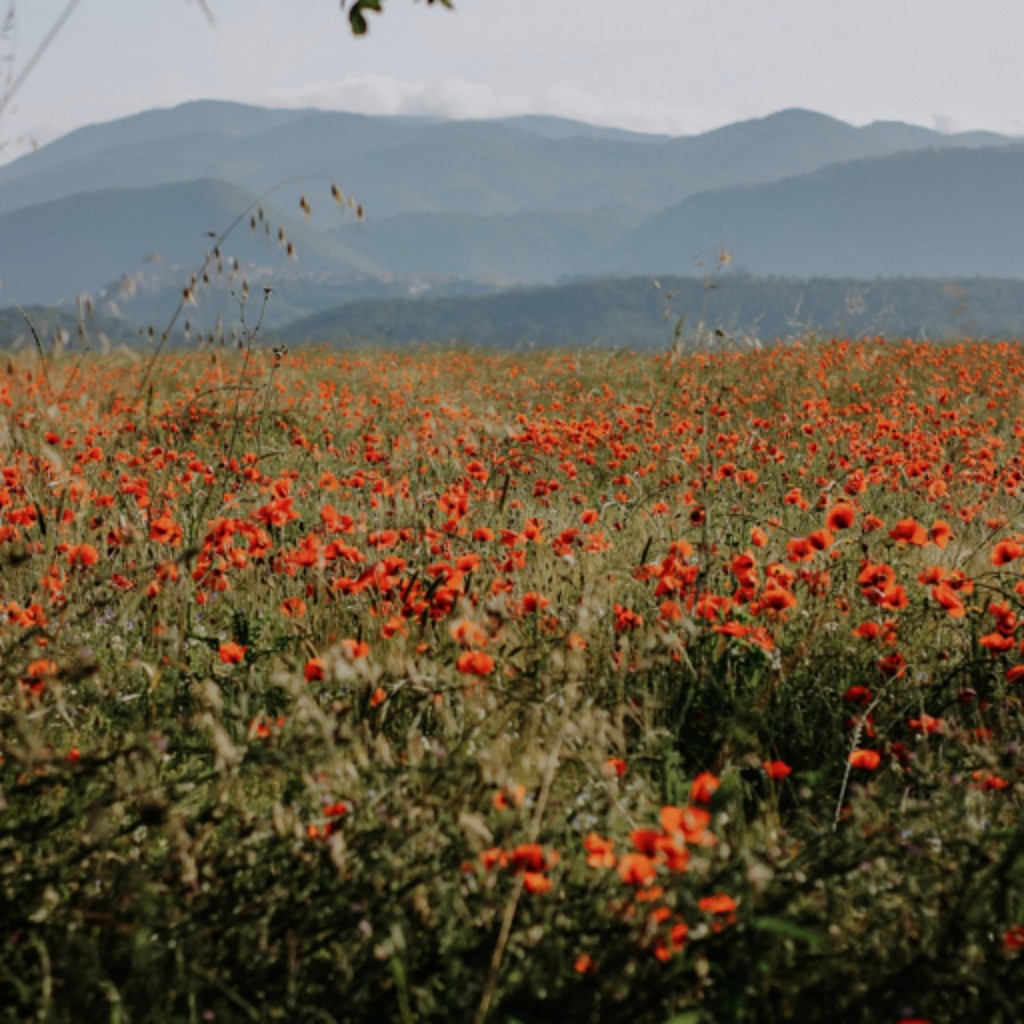} &
    \includegraphics[width=0.135\textwidth,height=0.135\textwidth]{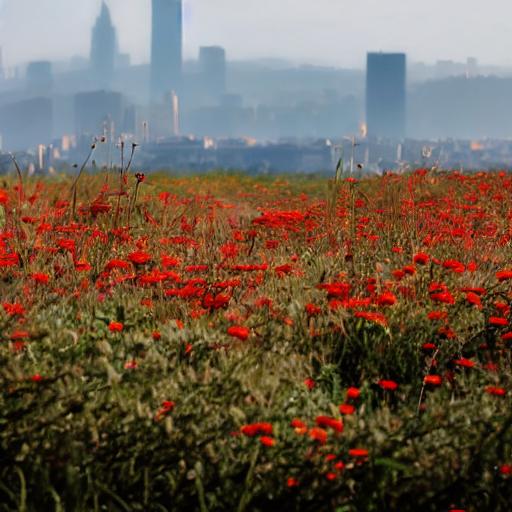} &
    \includegraphics[width=0.135\textwidth,height=0.135\textwidth]{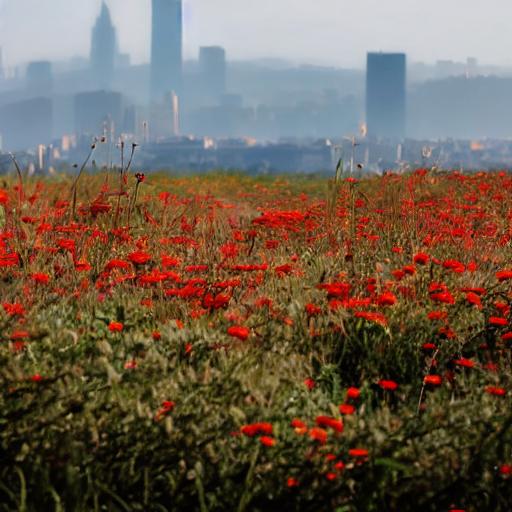} &

    \raisebox{0.06\textwidth}{
        \rotatebox[origin=t]{-90}{
            \scalebox{1.0} City
        }
    }  \\

    \includegraphics[width=0.135\textwidth,height=0.135\textwidth]{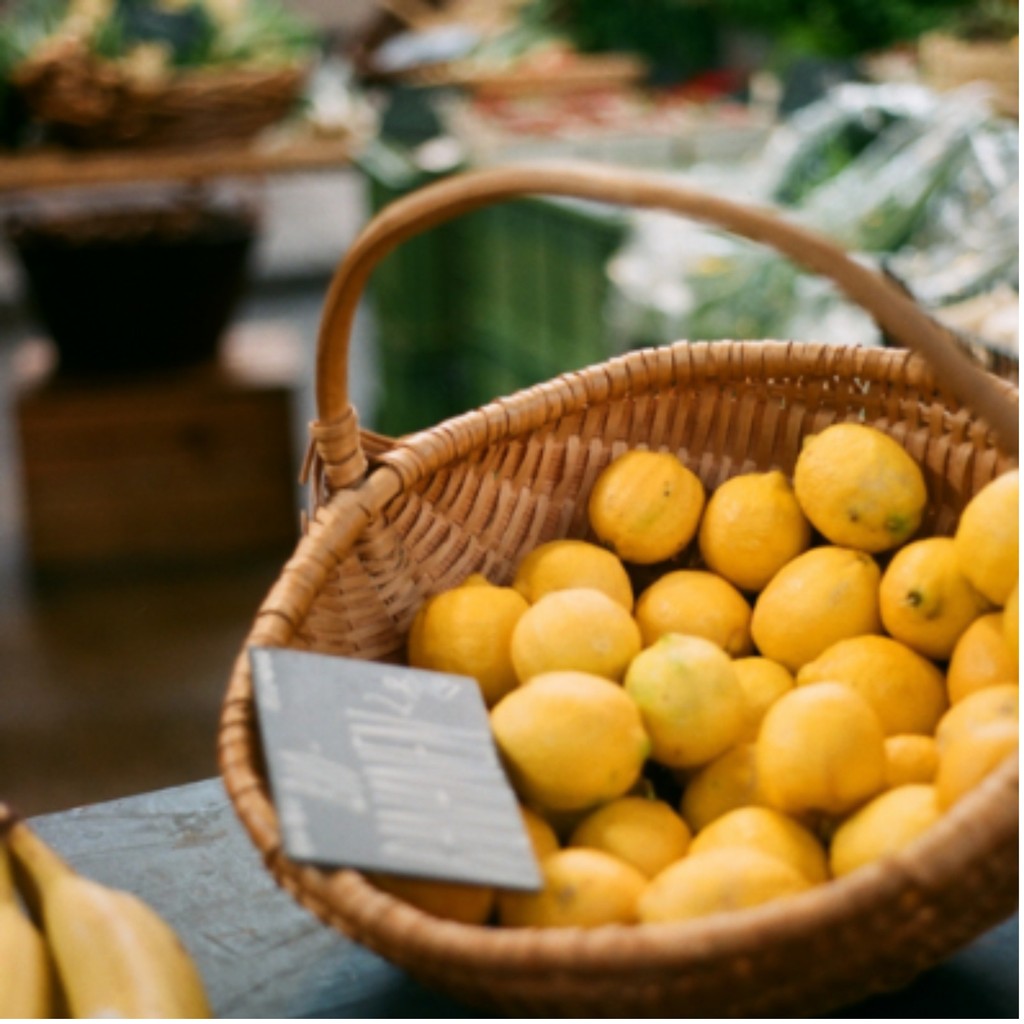} &
    \includegraphics[width=0.135\textwidth,height=0.135\textwidth]{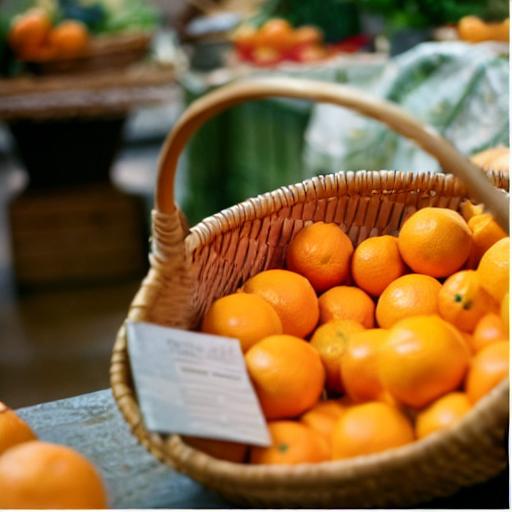} &
    \includegraphics[width=0.135\textwidth,height=0.135\textwidth]{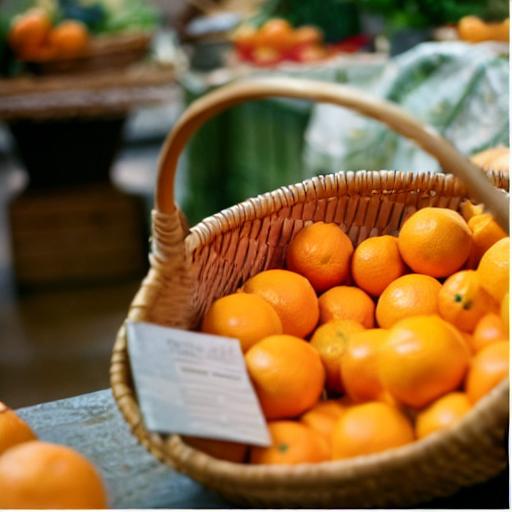} &

    \raisebox{0.06\textwidth}{
        \rotatebox[origin=t]{-90}{
            \scalebox{1.0} Oranges
        }
    }  \\

    \includegraphics[width=0.135\textwidth,height=0.135\textwidth]{figures/dds_vs_ef/input/30.jpeg} &
    \includegraphics[width=0.135\textwidth,height=0.135\textwidth]{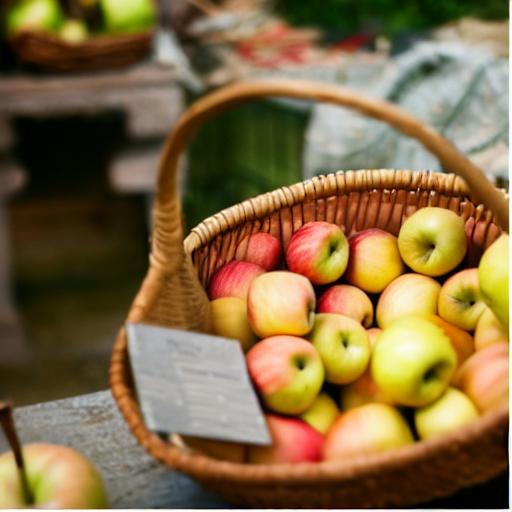} &
    \includegraphics[width=0.135\textwidth,height=0.135\textwidth]{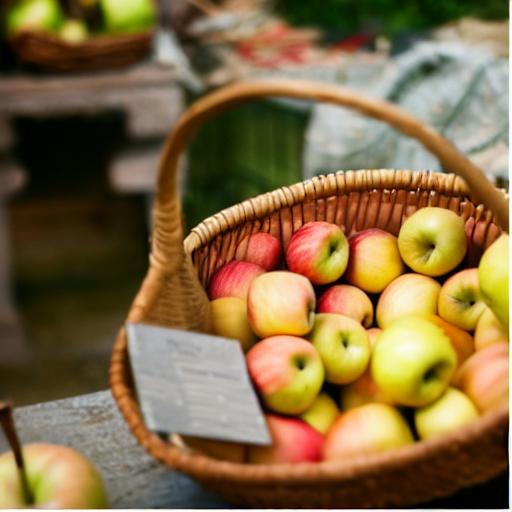} &

    \raisebox{0.06\textwidth}{
        \rotatebox[origin=t]{-90}{
            \scalebox{1.0} Apples
        }
    }  \\

    \includegraphics[width=0.135\textwidth,height=0.135\textwidth]{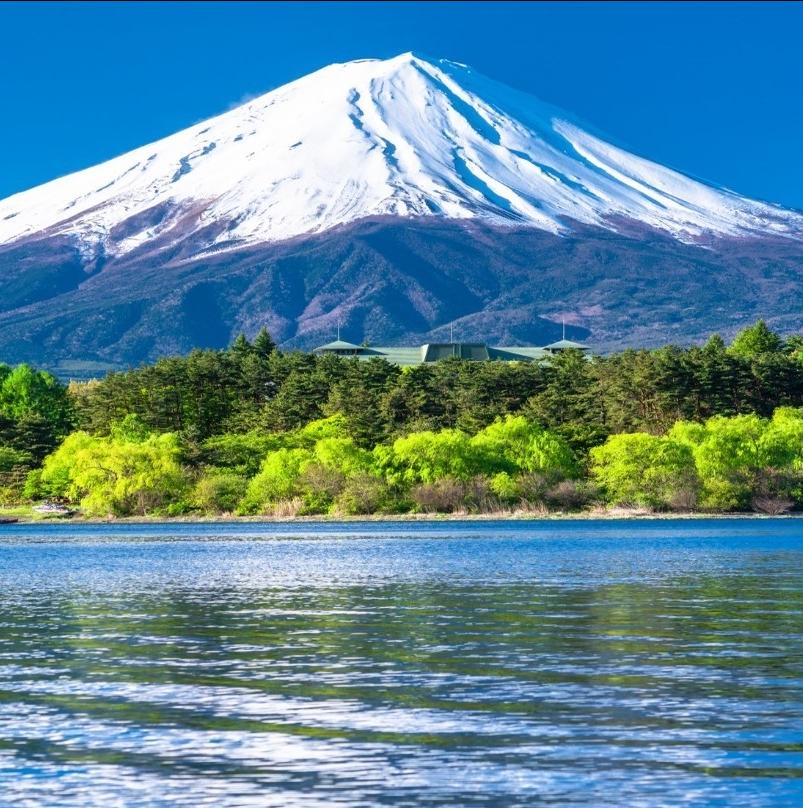} &
    \includegraphics[width=0.135\textwidth,height=0.135\textwidth]{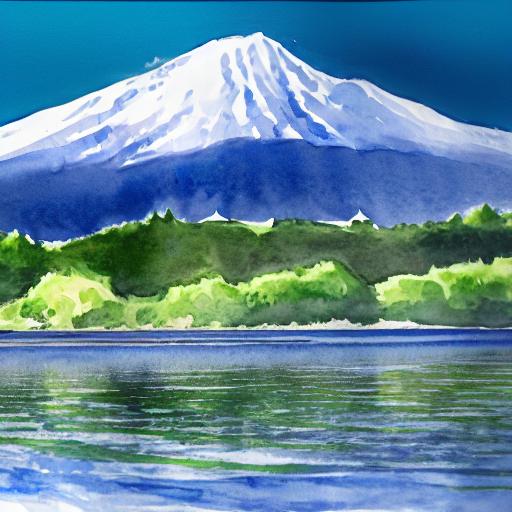} &
    \includegraphics[width=0.135\textwidth,height=0.135\textwidth]{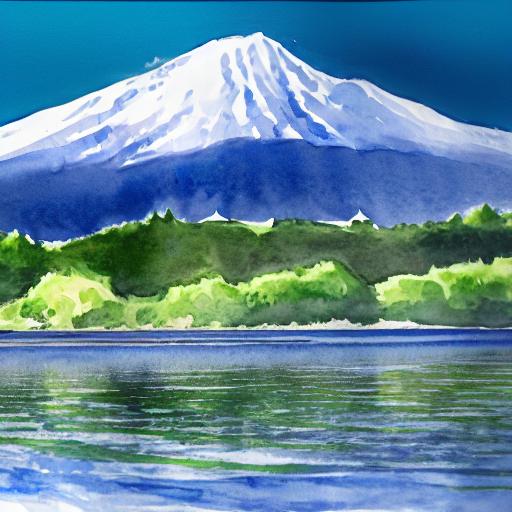} &

    \raisebox{0.06\textwidth}{
        \rotatebox[origin=t]{-90}{
            \scalebox{1.0} Watercolor
        }
    }  \\

    \end{tabular}
    }
    \caption{We show empirically that DDS with sequential timestep sampling and the learning rate of \cref{eq:lr} is equivalent to edit-friendly DDPM inversion. We run both methods using the same number of steps and guidance.}\label{fig:dds_vs_ef}
\end{figure}

This surprising result confirms that DDS and Edit-friendly share not only a similar correction term, but indeed under the specific choice of consecutive timesteps samples and a specific learning rate, the two methods converge to the exact same update rule. We further verify this empirically by editing a set of images using both methods, where DDS uses the modified schedule and learning rate. The results are shown in \cref{fig:dds_vs_ef}. As can be observed, the results are visually identical, further confirming our theoretical observation.

Finally, we note that a similar analysis can be applied to methods that combine DDS and EF (e.g., Posterior Distillation Sampling \cite{koo2024posterior}). Indeed, through an appropriate choice of learning rates and sampling steps, such methods can also be made to follow \cref{eq:edit_friendly_inference} for the specific case of image editing.

\clearpage

\end{document}